\documentclass{article}

\usepackage{amssymb}
\usepackage{centernot}
\usepackage{pifont}
\usepackage{tcolorbox}
\usepackage{rotating}
\newcommand{\cmark}{\ding{51}} 
\newcommand{\xmark}{\ding{55}} 
\tcbuselibrary{breakable}
\usepackage[preprint]{neurips_2026}

\usepackage[utf8]{inputenc} 
\usepackage[T1]{fontenc}    
\usepackage{hyperref}       
\usepackage{url}            
\usepackage{booktabs}       
\usepackage{amsfonts}       
\usepackage{amssymb}        
\usepackage{amsmath}        
\usepackage{nicefrac}       
\usepackage{microtype}      
\usepackage{xcolor}         
\usepackage{graphicx}
\usepackage{subcaption}
\usepackage{svg}
\usepackage{algorithm,algorithmic}
\usepackage{amsthm}

\usepackage{multirow}
\usepackage{tikz}
\usepackage{xcolor}
\usepackage{calc}
\usepackage{fancyvrb}
\usepackage{pifont}

\definecolor{wandbgold}{HTML}{FFBE00}

\theoremstyle{definition}

\theoremstyle{remark}

\newcommand{\projectbadge}[4][white]{%
  \tcbox[on line,
    colback=#1,
    colframe=#2,
    boxsep=1pt, left=4pt, right=4pt, top=2pt, bottom=2pt,
    boxrule=0.6pt,
    arc=3pt,
    fontupper=\small\bfseries
  ]{%
    {\hypersetup{hidelinks}\href{#3}{#4}}%
  }%
}

\definecolor{creamfill}{RGB}{255,249,235}   
\definecolor{creamstroke}{RGB}{210,180,140} 

\title{Efficient Continual Learning in Language Models via Thalamically Routed Cortical Columns}

\author{%
  Afshin Khadangi \\
  SnT, University of Luxembourg; \texttt{afshin.khadanki@uni.lu} \\
}

\begin{document}

\maketitle

\noindent\makebox[\linewidth]{%
  \projectbadge[wandbgold!20]{wandbgold}{https://api.wandb.ai/links/afshin-khadangi-university-of-luxembourg/fzk2ecqo}{W\&B Logs}%
  \hfill
  \projectbadge[creamfill]{creamstroke}{https://trc2lm.github.io}{TRC$^{2}$ Website}%
  \hfill
  \projectbadge[teal!10]{teal!50!black}{https://afshin.xyz}{Author Website}
}

\begin{abstract}
Large language models deployed in the wild must adapt to evolving data, user behavior, and task mixtures without erasing previously acquired capabilities. In practice, this remains difficult: sequential updates induce catastrophic forgetting, while many stabilization methods rely on external procedures that are costly, brittle, or difficult to scale. We present \textbf{TRC$^{2}$} (Thalamically Routed Cortical Columns), a decoder-only architecture that makes continual learning a property of the backbone itself. TRC$^{2}$ combines stacked cortical columns with a thalamic modulatory pathway for selective inter-column communication and a hippocampal pathway for event-selective retrieval, delayed surprise-based writing, and replay-driven consolidation. This design localizes fast plasticity while preserving a slower stable computation pathway. We further introduce a causal memory-update scheme and an online replay controller that adjusts consolidation strength from measured forgetting. Across a task-sequential language-modeling stream over C4, WikiText-103, and GSM8K, TRC$^{2}$ consistently improves task-boundary modeling quality and substantially reduces cumulative forgetting relative to Transformer, Mamba, MoE, DeepSeekMoE and continual learning baselines trained under the same pipeline. Ablations show that the thalamic and hippocampal components are central to the retention gains, while the full model remains competitive in throughput and training cost.
\end{abstract}

\section{Introduction}
\label{sec:introduction}

Large language models are increasingly expected to function as persistent systems rather than static artifacts. In deployment, they face evolving corpora, shifting user preferences, and changing task mixtures, often without clean task boundaries and under constraints that preclude full retraining or unrestricted replay. This creates the core tension of continual learning: the model must remain plastic enough to absorb new information while stable enough to preserve prior competence. Existing solutions only partly resolve this tension. Retraining is expensive, and lightweight updates such as adapters or low-rank tuning still accumulate interference over long streams. More broadly, this remains a form of catastrophic forgetting in sequential learning \cite{mccloskey1989catastrophic,kirkpatrick2017overcoming}.

Replay has remained one of the most reliable ways to reduce forgetting. From early rehearsal methods \cite{robins1995catastrophic} to episodic-memory approaches \cite{lopezpaz2017gradient,chaudhry2019tiny,rolnick2019experience}, revisiting selected past experience consistently improves retention under non-stationary training. Yet replay does not by itself solve the architectural problem. In most language model pipelines, memory is added as an external procedure rather than built into the model's computation, so adaptation is still expressed largely through global parameter updates.

This gap has become more visible as sequence models diversify. State-space and hybrid backbones improve efficiency and long-context handling \cite{lahoti2026mamba3,jambateam2025jamba}, while structured gating improves stability and context extrapolation \cite{qiu2025gatedattention}. At the same time, sparse routing remains fragile under distribution shift in mixture based and state-space settings \cite{therien2025continualmoe,zhan2025routingmamba}. Related work on test-time learning and model merging reaches a similar conclusion from another angle: deployment streams contain useful adaptation signals, but current methods usually exploit them through procedures layered on top of a mostly unchanged backbone \cite{hu2025testtime,bertolissi2025ttmm,qiu2025mingle}. The result is a system in which plasticity is useful but poorly localized.

\begin{figure}[t]
    \centering
    \includegraphics[width=\textwidth]{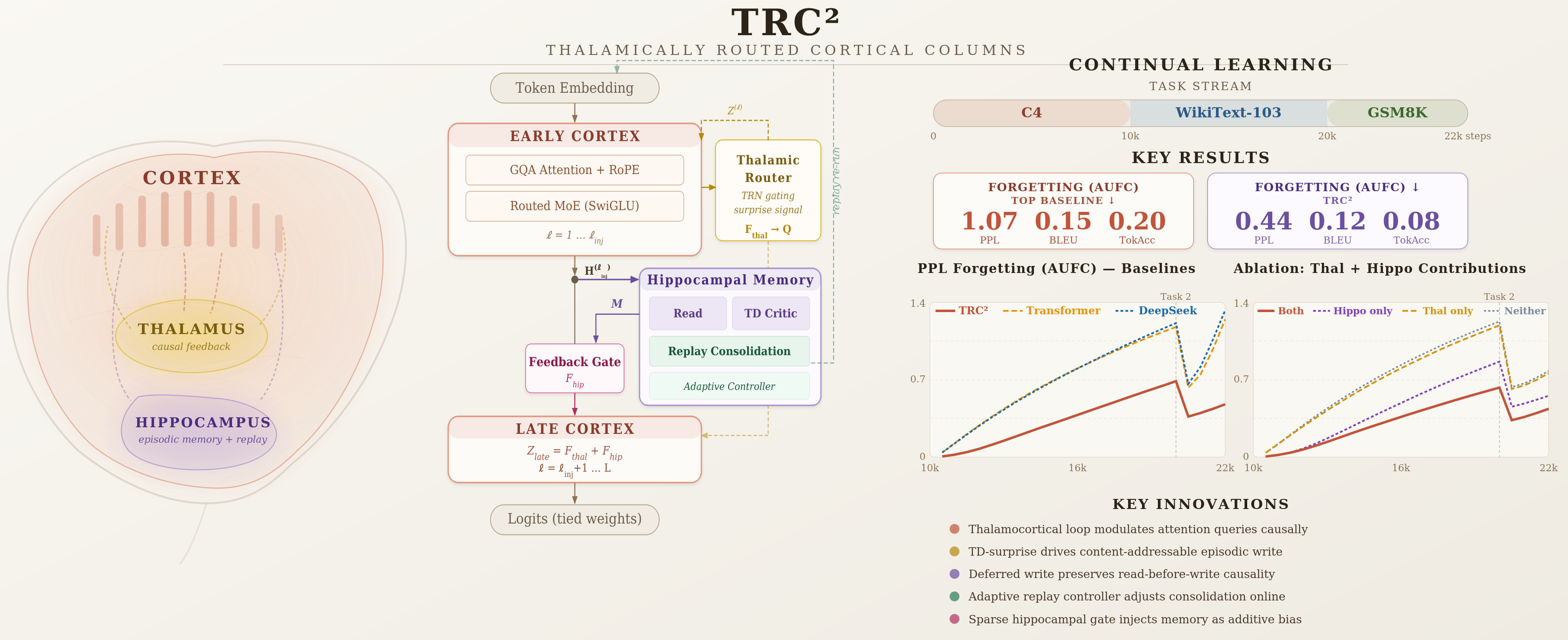}
    \caption{Overview of the \textsc{TRC}$^2$ architecture and its main empirical trends.}
    \label{fig:trc2_main}
\end{figure}

We therefore argue that continual learning should be expressed at the architectural level. We introduce \textbf{TRC}$^{2}$ (Figure \ref{fig:trc2_main}), a decoder-only backbone that combines stacked cortical columns with two global systems: a \emph{thalamic router}, which turns intermediate cortical state into selective modulatory feedback for later columns, and a \emph{hippocampal memory}, which supports event-selective retrieval, delayed surprise based writing, and replay driven consolidation. This design localizes fast adaptation while preserving a slower stable pathway. Unlike methods that constrain update subspaces but still rely on a largely static backbone and external regularization \cite{biswas2026ella}, \textsc{TRC}$^2$ makes interference control part of the forward computation itself.

We evaluate \textsc{TRC}$^2$ in a task-sequential language modeling stream over C4, WikiText-103, and GSM8K against Transformer, Mamba, MoE, and DeepSeekMoE baselines trained under the same pipeline. Across this setting, \textsc{TRC}$^2$ improves task boundary modeling quality and substantially reduces cumulative forgetting. Ablations show that these gains are driven by the thalamic and hippocampal pathways rather than scale alone.

Our contributions are as follows.
\begin{itemize}
    \item We present \textsc{TRC}$^2$, a decoder-only architecture for continual language modeling that integrates thalamic modulation and hippocampal episodic memory into the backbone.
    \item We introduce a causal memory update and consolidation scheme with retrospective hippocampal writes, replay from past stored chunks only, and online replay control driven by measured forgetting.
    \item We provide a controlled empirical study against Transformer, Mamba, MoE, DeepSeekMoE, and other continual learning baselines under a shared task stream and optimization pipeline, together with ablations that isolate the role of thalamic routing and hippocampal memory in quality, retention, and efficiency.
\end{itemize}

\section{Related Work}
\label{sec:related-work}

Continual learning for large language models now includes continual pre-training, domain adaptation, instruction updates, and long horizon knowledge maintenance. Recent surveys distinguish methods that modify the model from methods that rely on external augmentation, and also emphasize persistent evaluation difficulties at scale \cite{zheng2024lifelongllm_survey,shi2024clllm_survey}. Our work falls in the first category: we ask whether the backbone itself can be structured for streaming distribution shift.

A major line of work constrains adaptation to small parameter subspaces. DoRA improves low-rank adaptation by separating magnitude and direction \cite{liu2024dora}, while related methods study how such updates compose over time through gated LoRA mixtures \cite{wu2024mole} and lifelong mixtures with routing constraints \cite{wang-li-2024-lemoe}. These results show that both update geometry and update selection matter. \textsc{TRC}$^2$ differs by embedding selective routing and memory in the backbone rather than in an external low-rank path.

Replay and memory remain among the most effective tools for reducing forgetting. In language models, however, replay is often treated as an auxiliary procedure rather than an architectural component. Our approach is closest to replay based continual learning, but differs in two respects: memory access is event-selective and causal, and replay strength is adjusted online from measured forgetting. This places \textsc{TRC}$^2$ between purely parametric adaptation and purely external rehearsal.

Selective computation provides a second precedent. DeepSeekMoE studies expert specialization and shared experts \cite{dai-etal-2024-deepseekmoe}; LLaMA-MoE shows that dense decoders can be converted into sparse expert systems through continued pre-training \cite{zhu-etal-2024-llama}; sparse expertization can also be made highly parameter-efficient for instruction adaptation \cite{zadouri2024extremepeftmoe}; and router design itself often becomes the limiting factor in sparse systems \cite{zhang2025mor}. \textsc{TRC}$^2$ shares this emphasis on selectivity, but uses routing for modulatory control and plasticity management rather than only for scaling feed-forward capacity.

Efficient sequence backbones further widen the design space. Mamba establishes selective state-space computation as a competitive backbone \cite{gu2024mamba}; BlackMamba combines state-space dynamics with sparse experts \cite{anthony2024blackmamba}; RWKV-type models offer another recurrent alternative \cite{peng2024eaglefinch}; and FlashAttention-3 highlights the importance of kernel efficiency in practical performance \cite{shah2024flashattention3}. Our design is also informed by neuroscience, using the thalamic and hippocampal pathways as architectural guidance rather than literal biological models. This perspective is consistent with work on hippocampal-cortical consolidation, predictive reward representations, and structured biological control signals \cite{lee2023neocortical,yaghoubi2026predictive,hansen2022mapping,liu2025phase,song2024inferring}.

\section{Method}
\label{sec:method}

We study \textsc{TRC}$^2$, a decoder-only language model that maps an input sequence
\[
x_{1:T}\in\{1,\dots,V\}^T
\]
to autoregressive logits over a vocabulary of size $V$. For a batch of size $B$, the hidden state at layer $\ell$ is
\[
H^{(\ell)}\in\mathbb{R}^{B\times T\times d},
\]
where $d$ is the model width and $L$ is the number of cortical columns. The input embedding is
\begin{equation}
H^{(0)}_{b,t}=E[x_{b,t}],
\end{equation}
with rotary position encoding inside attention and RMS normalization throughout.

\textsc{TRC}$^2$ has three interacting parts: a stack of cortical columns for causal sequence modeling, a thalamic router that produces modulatory inter-column feedback, and a hippocampal memory that supports event-selective retrieval, delayed writing, and replay-based consolidation. Let
\begin{equation}
\ell_{\mathrm{inj}}=\max\!\left(1,\left\lfloor \frac{2L}{3}\right\rfloor\right)
\end{equation}
be the hippocampal injection point. Layers $1,\dots,\ell_{\mathrm{inj}}$ form the early cortex. Layers $\ell_{\mathrm{inj}}+1,\dots,L$ form the late cortex and receive both thalamic feedback from the previous column and a fixed hippocampal feedback term computed once from the early-cortex state.

\subsection{Cortical columns and thalamic routing}

Each cortical column maps a sequence state $H$ and thalamic signal $Z$ to
\begin{equation}
\begin{aligned}
U&=\mathrm{RMSNorm}(H),\quad
A=\mathrm{Attn}(U,Z),\quad
\widetilde{H}=H+\mathrm{Drop}(A),\quad
V=\mathrm{RMSNorm}(\widetilde{H}),\\
M&=\mathrm{MoE}(V),\quad
H^{+}=\widetilde{H}+\mathrm{Drop}(M),\quad
C=W_{\mathrm{L5}}H^{+},
\end{aligned}
\end{equation}
where $C$ is the layer-5 projection sent to the thalamic router.

In attention, grouped-query projections are formed from the normalized state, but the thalamic signal modulates only the queries:
\begin{equation}
Q = UW_Q + ZW_Q^{(\mathrm{thal})}, \qquad K = UW_K, \qquad V = UW_V.
\end{equation}
After rotary encoding and repetition of key-value heads, the attention output is
\begin{equation}
Y_{\mathrm{attn}}=
\mathrm{softmax}\!\left(
\frac{QK^\top}{\sqrt{d_h}} + M_{\mathrm{causal}}
\right)V,
\end{equation}
followed by the usual output projection. The feed-forward stage is a top-$k_E$ routed SwiGLU mixture of experts with an auxiliary load-balancing penalty,
\begin{equation}
\mathcal{L}_{\mathrm{lb}}=\sum_{\ell=1}^{L}\mathcal{L}_{\mathrm{lb}}^{(\ell)}.
\end{equation}
All routing details and exact load terms are deferred to the appendix.

The thalamic router converts $C$ into a causal modulatory signal for the next column. It first compresses the cortical state,
\begin{equation}
Z_0=\mathrm{RMSNorm}(CW_c),
\end{equation}
then combines a local pathway with a diffuse pathway driven by the strictly past mean $\mu^{<t}$:
\begin{align}
s_{b,t} &= \frac{1}{r}\left\|Z_{0,b,t,:}-\mu^{<t}_{b,t}\right\|_2^2, \\
Z_1 &= \mathrm{SiLU}(Z_0W_{\mathrm{loc}})
+ \beta_{\mathrm{diff}}\,g_{\mathrm{state}}\odot \mathrm{SiLU}(\mu^{<t}W_{\mathrm{diff}}), \\
g_{\mathrm{state}} &= \sigma(Z_0w_{\mathrm{state}}+b_{\mathrm{state}}+\alpha_s s).
\end{align}
A TRN-like competition stage then suppresses diffuse over activation, and the result is mapped back to model width:
\begin{equation}
F_{\mathrm{thal}}=
\left(\mathrm{Compete}(Z_1)W_{\mathrm{back}}\right)\odot \sigma(g_{\mathrm{mod}}).
\end{equation}
This signal is added to the query stream of the next cortical column.

\subsection{Hippocampal memory, feedback, and replay}

At layer $\ell_{\mathrm{inj}}$, the early-cortex state queries a content addressable episodic memory with keys $K_{\mathrm{mem}}$ and values $V_{\mathrm{mem}}$. Queries are
\begin{equation}
Q_{\mathrm{hip}}=H^{(\ell_{\mathrm{inj}})}W_Q^{(\mathrm{hip})}.
\end{equation}
For efficiency, retrieval uses only the most recent
\begin{equation}
n_{\mathrm{read}}=\min(n_{\mathrm{mem}},S_{\max})
\end{equation}
stored entries, keeps the exact top-$k_H$ matches, and returns
\begin{equation}
R_{b,t,:}=\sum_{i\in\mathcal{N}_{b,t}}\alpha_{b,t,i}V_{\mathrm{mem},i,:},
\qquad
M_{b,t,:}=\left(R_{b,t,:}W_O^{(\mathrm{hip})}\right)\odot \sigma(g_{\mathrm{hip}}).
\end{equation}

The hippocampus also computes intrinsic learning signals from a detached early-cortex state
\[
X=\mathrm{stopgrad}\!\left(H^{(\ell_{\mathrm{inj}})}\right).
\]
Fast and slow predictor heads define a learning progress reward,
\begin{equation}
r_t=\max\!\left(0,
\left\langle \bar{P}_t,\bar{X}_{t+1}\right\rangle
-
\left\langle \bar{P}^{\,\mathrm{slow}}_t,\bar{X}_{t+1}\right\rangle
\right),
\end{equation}
which is used together with fast and slow value heads to form a critic loss $\mathcal{L}_{\mathrm{td}}$ and a surprise score from the slow temporal difference residual. The raw predictor loss is denoted $\mathcal{L}_{\mathrm{pred}}^{\mathrm{raw}}$. Full predictor and critic definitions are deferred to the appendix.

Memory writes are strictly delayed. During the forward pass, the model stores pending tuples
\[
(X,s_{1:T}),
\]
but does not modify memory. After backpropagation, it selects up to $k_W$ high-surprise candidates per sequence, filters them with an adaptive threshold, and writes only the surviving states into a circular buffer:
\begin{equation}
k_{b,t}=W_K^{(\mathrm{write})}X_{b,t,:},
\qquad
v_{b,t}=W_V^{(\mathrm{write})}X_{b,t,:}.
\end{equation}
This preserves read-before-write causality within each forward pass.

The hippocampal readout is converted into a fixed feedback signal for all late columns:
\begin{align}
G &= \sigma\!\left([\mathrm{stopgrad}(H^{(\ell_{\mathrm{inj}})});M]W_{\mathrm{gate}} + b_{\mathrm{gate}}\right), \\
F_{\mathrm{hip}} &= \sigma(a_{\mathrm{hip}})\left((G\odot M)W_{\mathrm{hip}\rightarrow\mathrm{thal}}\right),
\end{align}
with optional channel sparsification in $G$. Thus late layers receive
\begin{equation}
Z^{(\ell-1)}_{\mathrm{late}} = F_{\mathrm{thal}}^{(\ell-1)} + F_{\mathrm{hip}}.
\end{equation}

The hippocampus also maintains two replay stores over raw token chunks: a recent ring buffer and a long-term reservoir. Replay samples are drawn before the current batch is inserted, so replay never uses the same batch being optimized. Given sampled replay chunks $x^{\mathrm{rep}}$, the replay loss is standard next-token supervision,
\begin{equation}
\mathcal{L}_{\mathrm{rep}}
=
-\frac{1}{B_R(L_R-1)}
\sum_{b=1}^{B_R}\sum_{t=1}^{L_R-1}
\log p_{\theta}\!\left(x^{\mathrm{rep}}_{b,t+1}\mid x^{\mathrm{rep}}_{b,1:t}\right).
\end{equation}

Replay strength is adjusted online by a controller that monitors forgetting on previously seen tasks in log-perplexity space. Let $\bar{f}(s)$ denote the mean forgetting signal and $P_{\mathrm{sel}}(s)$ the reference perplexity used for normalization. The controller forms
\begin{equation}
g(s)=\frac{\bar{f}(s)}{\max\!\left(1,\left|\log P_{\mathrm{sel}}(s)\right|\right)},
\end{equation}
tracks an exponential moving average of $g(s)$, and updates the replay coefficient $\lambda_{\mathrm{rep}}$, replay batch size $B_R$, and long-term replay fraction $\rho_{\mathrm{long}}$ through a clipped proportional integral rule. Exact update equations are given in the appendix.

\subsection{Full model and objective}

The early cortex runs first with recurrent thalamic feedback between adjacent columns. At layer $\ell_{\mathrm{inj}}$, the model reads from hippocampal memory and computes a fixed feedback term $F_{\mathrm{hip}}$. The late cortex then continues using the combined modulatory signal $F_{\mathrm{thal}}+F_{\mathrm{hip}}$. The final logits are
\begin{equation}
\mathrm{logits}
=
\mathrm{RMSNorm}\!\left(H^{(L)}\right)W_{\mathrm{vocab}}^\top,
\qquad
W_{\mathrm{vocab}}=E.
\end{equation}

Let $\mathcal{L}_{\mathrm{LM}}$ denote the masked next-token loss on labeled positions. The full objective is
\begin{equation}
\mathcal{L}
=
\mathcal{L}_{\mathrm{LM}}
+
\lambda_{\mathrm{router}}\mathcal{L}_{\mathrm{lb}}
+
\lambda_{\mathrm{td}}\mathcal{L}_{\mathrm{td}}
+
\lambda_{\mathrm{pred}}\mathcal{L}_{\mathrm{pred}}^{\mathrm{raw}}
+
\lambda_{\mathrm{rep}}\mathcal{L}_{\mathrm{rep}}.
\end{equation}
Memory writes are flushed after the backward pass and before the optimizer step. The slow predictor and value heads are then updated by exponential moving average:
\begin{align}
\bar{\theta} &\leftarrow \alpha \bar{\theta} + (1-\alpha)\theta, \\
\bar{\psi} &\leftarrow \alpha \bar{\psi} + (1-\alpha)\psi.
\end{align}
This yields a strictly causal read path, a delayed write path, and a separate replay path for long horizon consolidation.

\section{Experiments and Results}
\label{sec:experiments}

\subsection{Experimental setup}

We evaluate \textsc{TRC}$^2$ in a task-sequential language modeling setting with explicit task boundaries, which lets us measure both task boundary quality and cross task retention. All runs use a single node with 4 NVIDIA V100 GPUs (32GB per GPU), distributed data parallel training, and mixed precision.

\paragraph{Data, schedule, and baselines.}
The task stream is
\[
\text{C4} \rightarrow \text{WikiText-103} \rightarrow \text{GSM8K},
\]
using C4 \cite{raffel2020t5}, WikiText-103 \cite{merity2016pointer}, and GSM8K \cite{cobbe2021gsm8k}. All periodic evaluation, model selection, and forgetting measurements are computed on \texttt{validation} splits or a subset of \texttt{train} splits. Official \texttt{test} splits are used once at the end of training when available. The optimizer step budget is
\[
N_{\text{C4}}=10{,}000,\qquad
N_{\text{WikiText-103}}=10{,}000,\qquad
N_{\text{GSM8K}}=2{,}000,
\]
for a total of
\[
N_{\text{total}}=22{,}000
\]
steps. Each task also provides a small fixed control subset from its training split, used only to drive the replay controller.

We compare \textsc{TRC}$^2$ against decoder-only Transformer, Mamba, MoE, DeepSeekMoE~\cite{dai2024deepseekmoe}, Block Attention Residuals (BlkAttnRes)~\cite{kimiteam2026attentionresiduals}, FALCON~\cite{zhang2026falcon}, and LoraDRS~\cite{liu2025lora}, all trained under the same tokenizer, data pipeline, and optimization framework. Model comparisons use matched optimizer-step budgets, a shared tokenizer and data pipeline, and a shared optimization framework at comparable parameter scales.

\paragraph{Training and metrics.}
All models use the same trainer with AdamW, gradient accumulation, warmup, cosine decay, and gradient clipping, and are evaluated periodically on all tasks seen so far. After each task, we record the post-task validation score and use it as the reference point for later forgetting. For \textsc{TRC}$^2$, replay is controlled online by updating the replay loss weight, replay batch size, and long term replay fraction from measured degradation on earlier tasks. We report held-out loss and perplexity as the primary metrics. We additionally report secondary teacher-forced overlap metrics computed from token-wise $\arg\max$ predictions restricted to labeled positions in the packed language modeling targets: token accuracy, sequence exact match, BLEU, chrF, and ROUGE. For GSM8K, these are language modeling metrics on serialized solution text rather than answer-only solve accuracy, so we interpret them as next token modeling quality on the GSM8K distribution rather than as standard reasoning accuracy.

\subsection{Results}

\begin{figure}[t]
    \centering
    \includegraphics[width=0.9\textwidth]{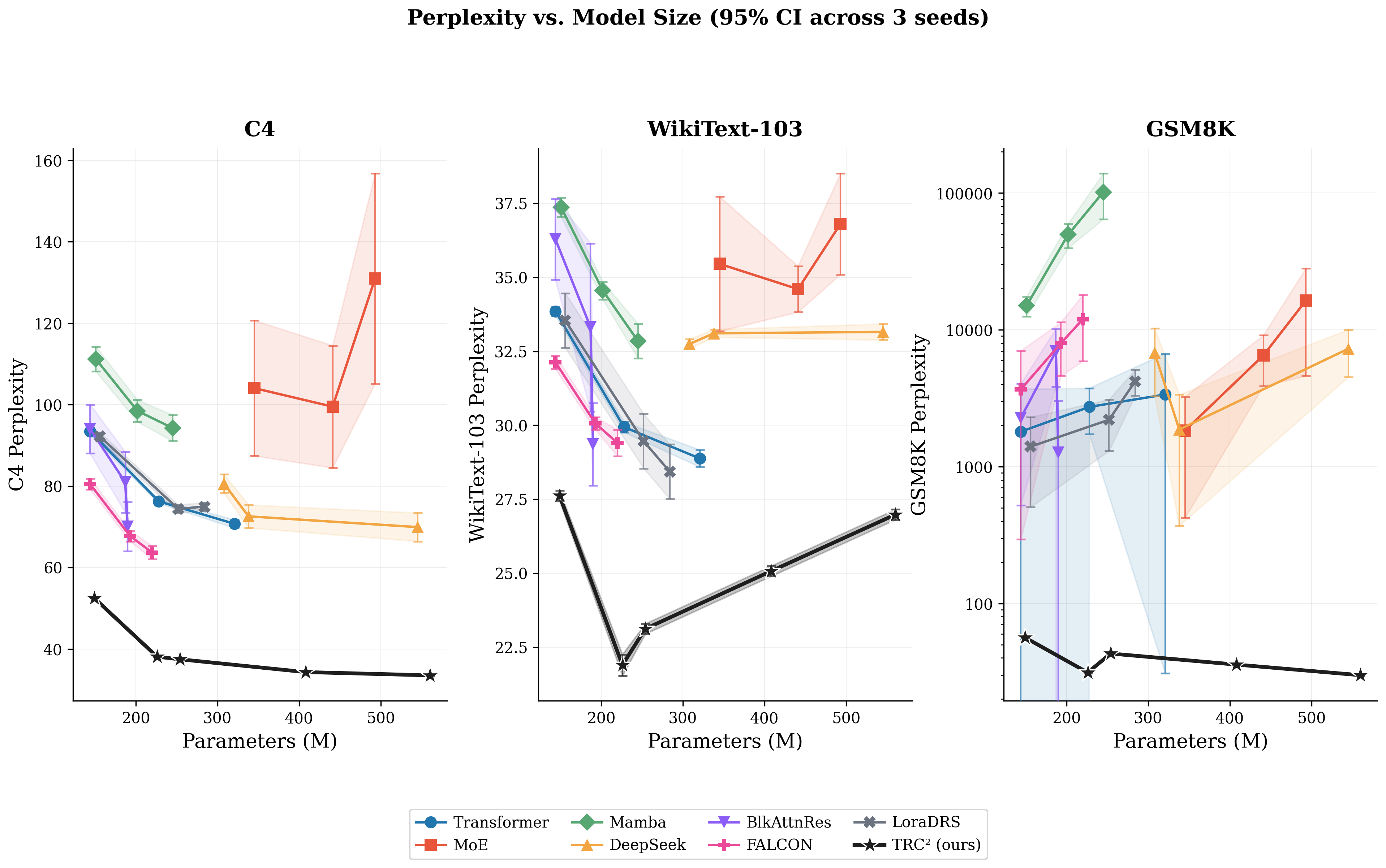}
    \caption{Task-boundary perplexity versus model size. Points show means across 3 seeds and shaded bands show 95\% confidence intervals. Lower values are better in all panels. C4 is reported at the 10k boundary, WikiText-103 at the 20k boundary, and GSM8K at the 22k boundary; the GSM8K axis is shown on a log scale.}
    \label{fig:ppl_vs_model_size}
\end{figure}

\begin{figure}[H]
    \centering
    \includegraphics[width=0.9\textwidth]{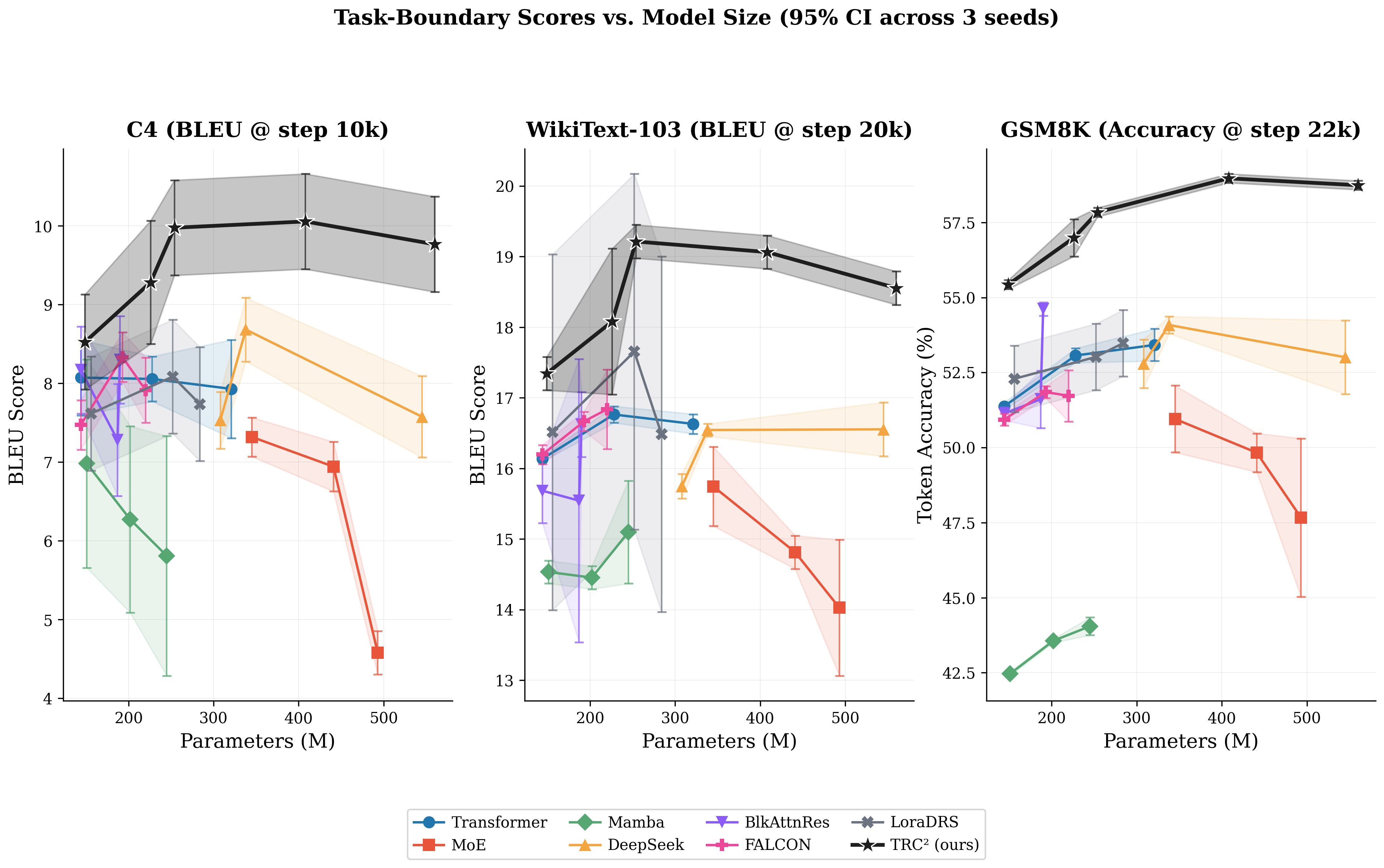}
    \caption{Secondary teacher-forced text metrics versus model size. Points show means across 3 seeds and shaded bands show 95\% confidence intervals. Higher values are better in all panels. For C4 and WikiText-103 we report BLEU at steps 10k and 20k, respectively; for GSM8K we report token accuracy at step 22k.}
    \label{fig:boundary_scores_vs_model_size}
\end{figure}

\begin{figure}[t]
    \centering
    \includegraphics[width=0.8\textwidth]{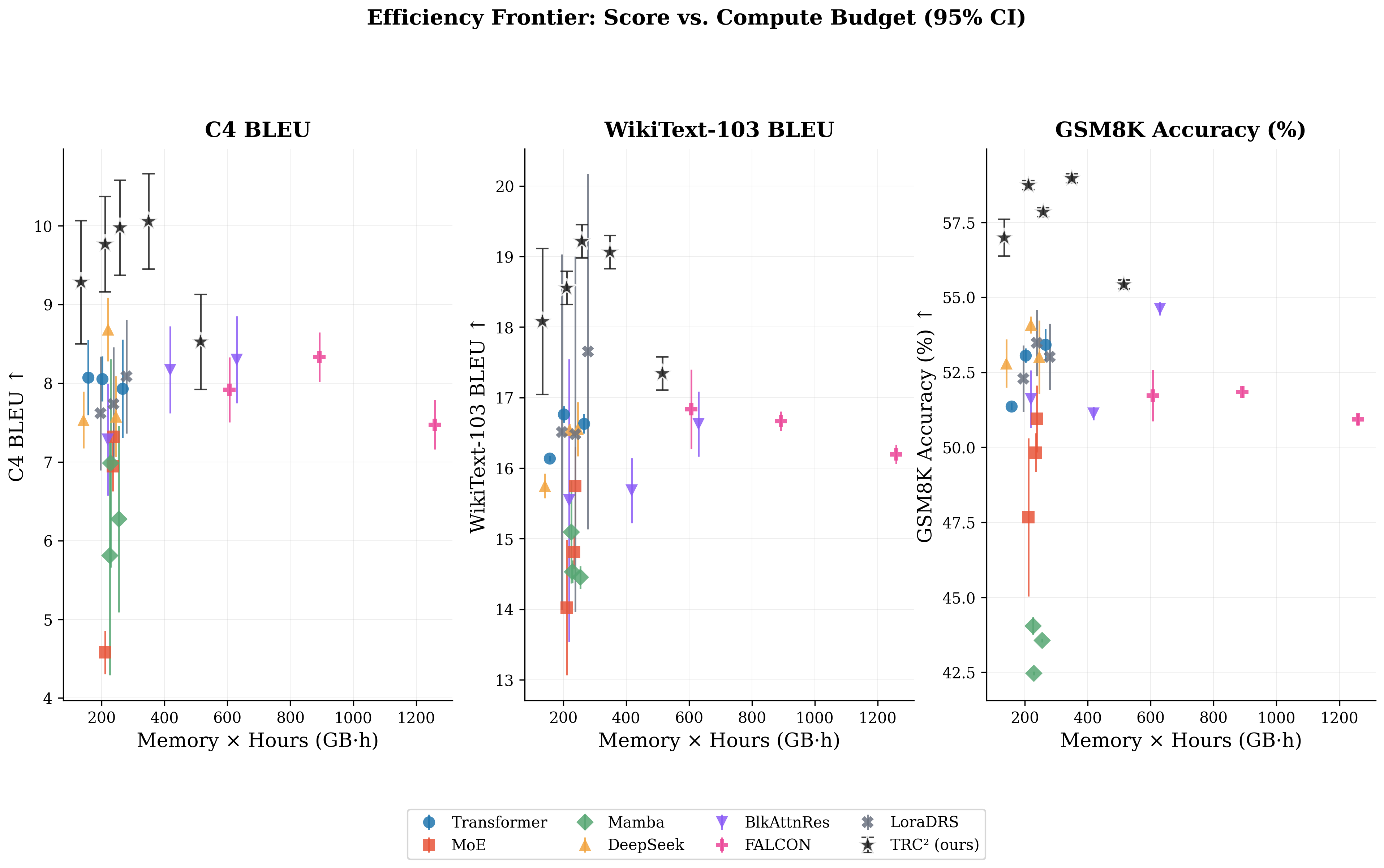}
    \caption{Efficiency frontier for task-boundary quality versus compute budget, using the Memory$\times$Hour/GPU statistic. Points show means across 3 seeds and error bars denote 95\% confidence intervals. Higher values are better on all three panels.}
    \label{fig:efficiency_frontier}
\end{figure}

\begin{figure}[t]
    \centering
    \includegraphics[width=0.6\textwidth]{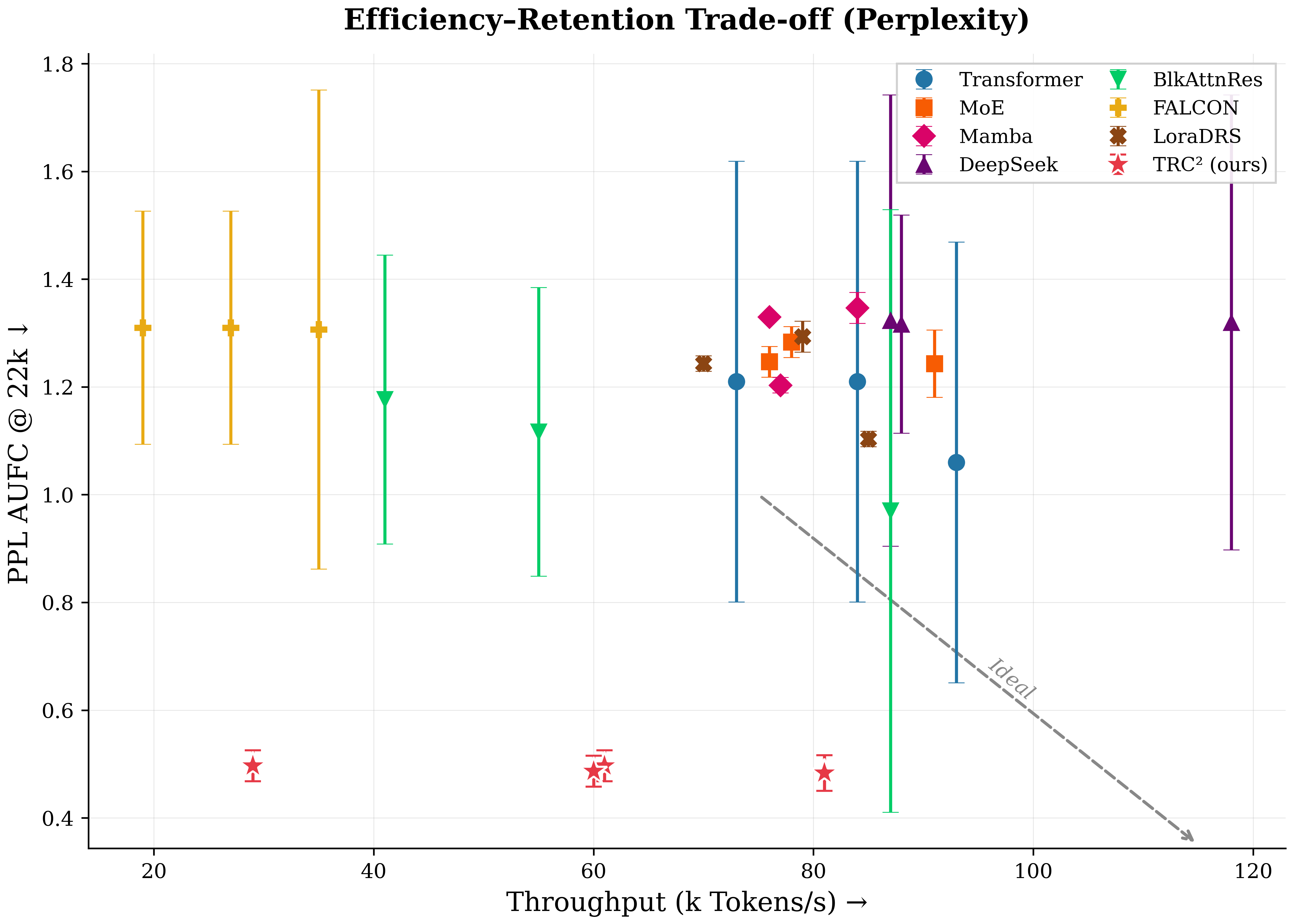}
    \caption{Efficiency retention tradeoff measured by throughput and perplexity forgetting. Points show model configurations, the horizontal axis reports throughput in thousands of tokens per second, and the vertical axis reports PPL AUFC at step 22k, where lower values indicate better retention. Error bars denote 95\% confidence intervals across seeds. The preferred region is the lower right corner, corresponding to higher throughput and lower cumulative forgetting.}
    \label{fig:efficiency_retention_ppl}
\end{figure}

We defer the full scalar boundary and retention results to the appendix and summarize the main empirical picture with Figures~\ref{fig:ppl_vs_model_size}--\ref{fig:efficiency_retention_ppl} and the ablation study in Table~\ref{tab:ablation_trc2_d768_l4}.

Figures~\ref{fig:ppl_vs_model_size} and~\ref{fig:boundary_scores_vs_model_size} show a consistent task boundary advantage for \textsc{TRC}$^2$ across model sizes. In perplexity, \textsc{TRC}$^2$ occupies the lowest region on all three tasks, with the largest separation appearing on GSM8K. The gap is substantial there: all \textsc{TRC}$^2$ variants remain far below the baseline families on the log-scale GSM8K panel. The same pattern appears in the secondary teacher-forced text metrics. On C4 and WikiText-103, \textsc{TRC}$^2$ remains in the strongest region of the BLEU frontier, and on GSM8K it achieves the highest token accuracy across the compared families. These gains are visible across multiple parameter scales, which suggests that the effect is architectural rather than tied to a narrow operating point.

The efficiency picture is more mixed, but still favorable. Figure~\ref{fig:efficiency_frontier} shows that larger \textsc{TRC}$^2$ variants pay a clear systems cost for thalamic routing, hippocampal retrieval, and replay, so the absolute fastest runs remain dense or hybrid baselines. Even so, several \textsc{TRC}$^2$ variants remain competitive despite this overhead. In particular, the d768-l4 configuration lies in a notably favorable region of the frontier, combining strong boundary quality with competitive compute cost. More broadly, the results suggest that the added routing and memory machinery buys a better quality retention tradeoff than simply scaling a standard backbone. Additional grouped by width and confidence interval views are deferred to Appendix~\ref{app:aggregate-task-boundary-viz}.

Figure~\ref{fig:efficiency_retention_ppl} makes the retention result especially clear. Relative to the baselines, \textsc{TRC}$^2$ shifts the throughput-retention frontier toward lower cumulative forgetting at competitive speed. All \textsc{TRC}$^2$ variants lie in the lowest band of PPL AUFC, while most baselines occupy a markedly higher forgetting region. The d768-l4 model is particularly strong in this tradeoff, combining one of the best throughputs among the \textsc{TRC}$^2$ variants with the lowest observed PPL AUFC. This pattern aligns with the full retention summaries deferred to the appendix and indicates that the retention gain is broad across scale rather than isolated to one model size.

The ablation study in Table~\ref{tab:ablation_trc2_d768_l4} clarifies the role of the two global modules. Removing thalamus and hippocampus reduces overhead and can improve some final step metrics, especially on the last task, but at a clear cost in cumulative retention. The full model gives the strongest AUFC profile, with the largest degradation appearing when hippocampal memory is removed. This separation between endpoint score and retention indicates that the thalamic and hippocampal pathways do more than add capacity: they change how plasticity is allocated across the stream.

Taken together, the results support the central claim of the paper. \textsc{TRC}$^2$ improves the stability-plasticity tradeoff in a practically meaningful way: it raises task boundary quality, lowers cumulative forgetting, and remains competitive in throughput and total training cost.

\begin{table}[t]
  \centering
  \scriptsize
  \caption{Ablation study on the TRC$^{2}$ d768-l4 configuration (top row in each panel). Rows are identified by the presence or absence of the thalamic and hippocampal modules. The upper panel reports the last logged held-out scores at step 22k for perplexity, BLEU, and GSM8K token accuracy. The lower panel reports area-under-forgetting-curve (AUFC) values at steps 20k and 22k.}
  \label{tab:ablation_trc2_d768_l4}
  
  \begin{tabular}{cc|cccccc|cc}
    \toprule
    \multicolumn{10}{c}{\textbf{Last-step scores at 22k}} \\
    \midrule
    \multirow{2}{*}{Thal.} & \multirow{2}{*}{Hippo.} & \multicolumn{3}{c}{PPL $\downarrow$} & \multicolumn{3}{c}{Text score $\uparrow$} & \multirow{2}{*}{Tokens/s $\uparrow$} & \multirow{2}{*}{$\frac{\mathrm{Mem}\times \mathrm{Hour}}{\mathrm{GPU}}\downarrow$} \\
    \cmidrule(lr){3-5}
    \cmidrule(lr){6-8}
    & & C4 & Wiki & GSM & C4 & Wiki & GSM (\%) & & \\
    \midrule
    \cmark & \cmark & \textbf{990.26} & 68.70 & 31.28 & \textbf{4.05} & 15.03 & 57.19 & $\sim$ 103000 & $134\,\mathrm{GB\cdot h}$ \\
    \xmark & \cmark & 1284.79 & \textbf{55.26} & 30.93 & 3.23 & \textbf{15.52} & 57.15 & $\sim$ 109000 & $131\,\mathrm{GB\cdot h}$ \\
    \cmark & \xmark & 2015.73 & 613.29 & 27.44 & 1.82 & 3.28 & 57.74 & $\sim$ 147000 & $95\,\mathrm{GB\cdot h}$ \\
    \xmark & \xmark & 2050.02 & 550.08 & \textbf{26.79} & 2.30 & 4.50 & \textbf{58.51} & \textbf{$\sim$ 154000} & \textbf{\boldmath$87\,\mathrm{GB\cdot h}$} \\
    \midrule
    \multicolumn{10}{c}{\textbf{Continual-learning retention (AUFC)}} \\
    \midrule
    \multirow{2}{*}{Thal.} & \multirow{2}{*}{Hippo.} & \multicolumn{2}{c}{PPL AUFC $\downarrow$} & \multicolumn{2}{c}{BLEU AUFC $\downarrow$} & \multicolumn{2}{c}{TokAcc AUFC $\downarrow$} & \multirow{2}{*}{Tokens/s $\uparrow$} & \multirow{2}{*}{$\frac{\mathrm{Mem}\times \mathrm{Hour}}{\mathrm{GPU}}\downarrow$} \\
    \cmidrule(lr){3-4}
    \cmidrule(lr){5-6}
    \cmidrule(lr){7-8}
    & & 20k & 22k & 20k & 22k & 20k & 22k & & \\
    \midrule
    \cmark & \cmark & \textbf{0.63} & \textbf{0.44} & \textbf{0.20} & \textbf{0.12} & 0.13 & \textbf{0.08} & $\sim$ 103000 & $134\,\mathrm{GB\cdot h}$ \\
    \xmark & \cmark & 0.86 & 0.56 & 0.20 & 0.13 & \textbf{0.12} & 0.08 & $\sim$ 109000 & $131\,\mathrm{GB\cdot h}$ \\
    \cmark & \xmark & 1.19 & 0.76 & 0.26 & 0.19 & 0.20 & 0.15 & $\sim$ 147000 & $95\,\mathrm{GB\cdot h}$ \\
    \xmark & \xmark & 1.23 & 0.78 & 0.26 & 0.20 & 0.21 & 0.16 & \textbf{$\sim$ 154000} & \textbf{\boldmath$87\,\mathrm{GB\cdot h}$} \\
    \bottomrule
  \end{tabular}
\end{table}

\section{Discussion}
\label{sec:discussion}

The main result is not only that \textsc{TRC}$^2$ reaches strong task boundary scores, but that it does so while forgetting less. Figures~\ref{fig:ppl_vs_model_size} and~\ref{fig:boundary_scores_vs_model_size} show that this advantage holds across scales and across both loss based and text based metrics. The gains are largest on the harder later tasks, especially WikiText-103 and GSM8K, which is where continual interference is most costly. 

The retention result is sharper still. Figure~\ref{fig:efficiency_retention_ppl} shows that \textsc{TRC}$^2$ shifts the throughput retention frontier toward lower cumulative forgetting at competitive speed. This matters because it supports the central architectural claim of the paper: separating fast adaptation from the main cortical pathway through thalamic modulation, episodic memory, and replay reduces destructive interference over the stream. In other words, the model is not simply learning new tasks well. It is learning them without erasing as much of what came before.

The ablation in Table~\ref{tab:ablation_trc2_d768_l4} helps explain why. Removing thalamus or hippocampus can improve some endpoint metrics and reduce overhead, but it weakens cumulative retention, with the largest drop appearing when hippocampal memory is removed. This shows that the global modules are not decorative additions. They alter how plasticity is distributed across training. The strongest final checkpoint on the most recent task is therefore not always the strongest continual-learning solution.

The systems tradeoff is real, but not prohibitive. Figure~\ref{fig:efficiency_frontier} shows that \textsc{TRC}$^2$ pays extra cost for routing, retrieval, and replay, yet several variants remain in a practical regime. The d768-l4 model is especially informative: it combines strong task boundary quality, the best perplexity retention profile, and competitive throughput. 

Beyond model quality, architectures reducing forgetting can make long-lived pipelines more efficient and easier to update without full retraining. However, stronger persistent adaptation may also preserve harmful or incorrect information more, which makes careful evaluation and monitoring important.

\section{Conclusion}
\label{sec:conclusion}

\textsc{TRC}$^2$ introduces a decoder architecture for continual language modeling built around thalamically modulated cortical columns, hippocampal episodic memory, and replay guided consolidation. Rather than letting adaptation diffuse through the full backbone, the model routes it through selective modulation, event-selective memory, and delayed consolidation.

Across the main experiments, \textsc{TRC}$^2$ improves task boundary performance and reduces cumulative forgetting relative to strong Transformer, MoE, Mamba, DeepSeekMoE, and BlkAttnRes baselines. The retention gain is strongest in perplexity based metrics, and the ablations show that it is driven by the thalamic and hippocampal pathways rather than scale alone.

The remaining limitation is mainly computational. Several configurations already achieve a favorable balance between quality, retention, and compute, but the sparse and memory-based pathways can still be executed more efficiently. That makes systems optimization the natural next step. Overall, the results support a simple conclusion: continual language modeling benefits from an explicit separation between stable computation and fast localized plasticity, and \textsc{TRC}$^2$ provides a concrete decoder design that realizes this principle.


\newpage
{\small
\bibliographystyle{plain}
\bibliography{references}
}

\newpage

\appendix
\renewcommand{\thefigure}{\thesection.\arabic{figure}}
\renewcommand{\thetable}{\thesection.\arabic{table}}
\renewcommand{\theequation}{\thesection.\arabic{equation}}
\setcounter{figure}{0}
\setcounter{table}{0}
\setcounter{equation}{0}

\section{Appendix Overview}
\label{app:overview}

This appendix is organized to read as a self-contained companion to the main paper. Section~\ref{app:full-arch} gives the full architectural view of \textsc{TRC}$^2$ and the detailed module definitions used in implementation. Section~\ref{app:protocol} collects the training setup, task schedule, evaluation protocol, and metric definitions. Section~\ref{app:full-main-results} provides the full result tables and the efficiency frontier figure. Section~\ref{app:ablation-components} expands the component level analysis. Section~\ref{app:causality} reports the causality and state semantics verification study. Section~\ref{app:complexity} summarizes computational complexity. Sections~\ref{app:aggregate-task-boundary-viz}--\ref{app:curves} provide additional aggregate visualizations and full training dynamics.

All configurations, scalar metrics, and system statistics are logged in Weights \& Biases.

\section{Full Architecture and Technical Details}
\label{app:full-arch}

This section collects the complete architectural specification of \textsc{TRC}$^2$. The model combines three interacting subsystems: a cortical backbone for causal sequence modeling, a thalamic router for state dependent inter-column modulation, and a hippocampal module for event-selective retrieval, delayed surprise based writing, and replay driven consolidation.

\begin{figure}[htbp]
    \centering
    \includegraphics[width=\textwidth]{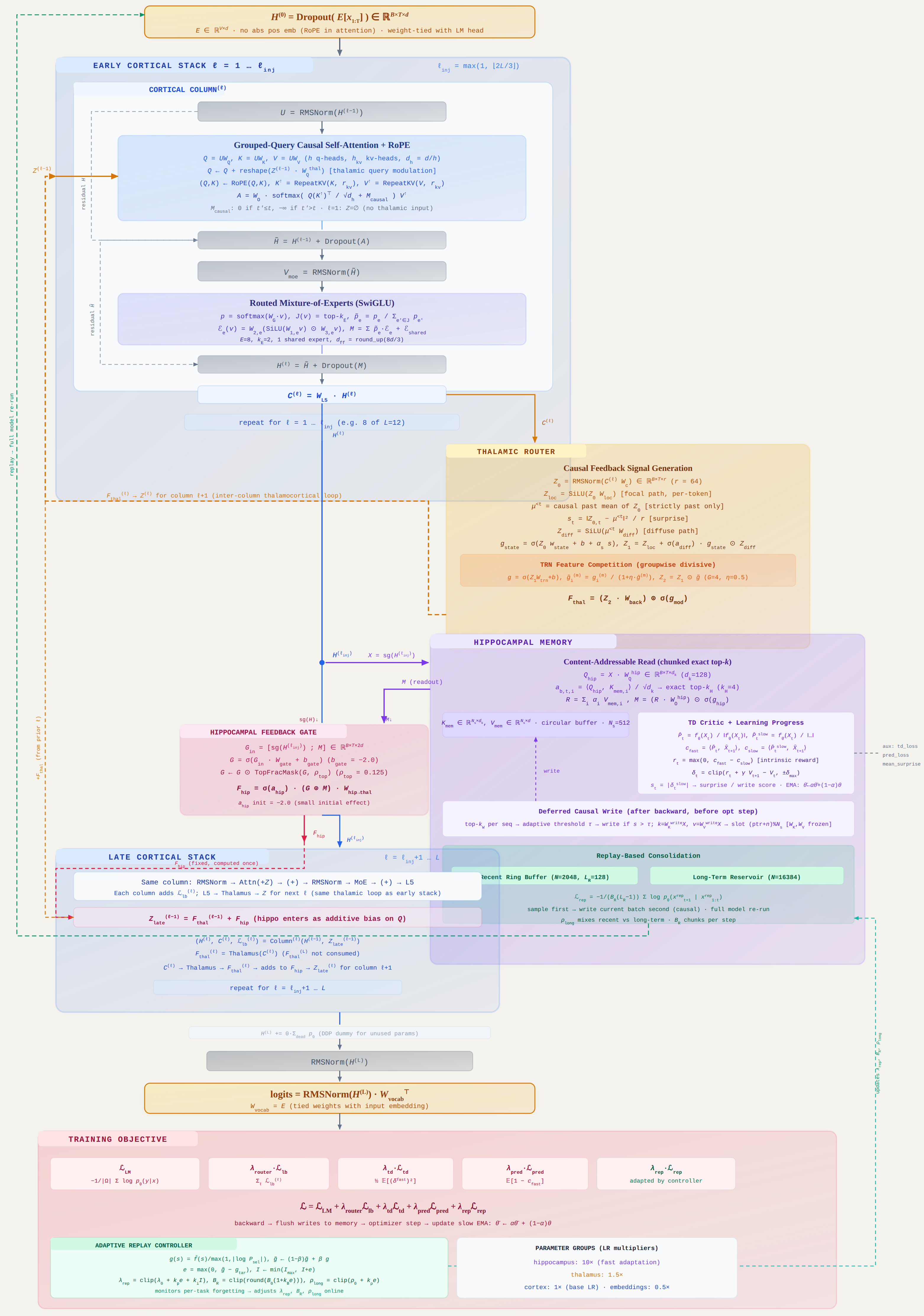}
    \caption{Full overview of the \textsc{TRC}$^2$ architecture. Tokens are processed through stacked cortical columns. Intermediate cortical state drives a thalamic modulatory pathway and a hippocampal pathway with event-selective retrieval, delayed surprise based writing, and replay based consolidation. The thalamic pathway shapes subsequent cortical processing, while the hippocampal pathway injects retrieved contextual information into later columns and supports continual adaptation across the task stream.}
    \label{fig:trc2_architecture}
\end{figure}

\subsection{Cortical column: attention and mixture-of-experts}
\label{app:cortex-details}

Each cortical column applies RMS normalization before attention and before the routed SwiGLU mixture-of-experts module. For input
\[
H\in\mathbb{R}^{B\times T\times d},
\]
let
\[
U=\mathrm{RMSNorm}(H).
\]
The attention projections are
\begin{align}
Q &= UW_Q \in \mathbb{R}^{B\times T\times h d_h},\\
K &= UW_K \in \mathbb{R}^{B\times T\times h_{\mathrm{kv}} d_h},\\
V &= UW_V \in \mathbb{R}^{B\times T\times h_{\mathrm{kv}} d_h}.
\end{align}

with $d_h=d/h$ and replication factor
\[
r_{\mathrm{kv}}=\frac{h}{h_{\mathrm{kv}}}.
\]
After reshaping into head form and applying rotary position encoding, the key and value heads are repeated to match the number of query heads:
\begin{equation}
K^{\uparrow}=\mathrm{RepeatKV}(K,r_{\mathrm{kv}}),\qquad
V^{\uparrow}=\mathrm{RepeatKV}(V,r_{\mathrm{kv}}).
\end{equation}
The thalamic signal $Z\in\mathbb{R}^{B\times T\times d}$ is applied only to the query stream:
\begin{equation}
Q \leftarrow Q + \mathrm{reshape}(ZW_Q^{(\mathrm{thal})}).
\end{equation}
The causal attention output is
\begin{equation}
Y_{\mathrm{attn}}
=
\mathrm{softmax}\!\left(
\frac{Q\left(K^{\uparrow}\right)^{\top}}{\sqrt{d_h}} + M_{\mathrm{causal}}
\right)V^{\uparrow},
\end{equation}
where
\begin{equation}
(M_{\mathrm{causal}})_{t,t'}=
\begin{cases}
0, & t' \le t,\\
-\infty, & t' > t.
\end{cases}
\end{equation}
The projected attention output is
\begin{equation}
A = W_O\,\mathrm{concat}_{\mathrm{heads}}(Y_{\mathrm{attn}}).
\end{equation}

The feed-forward half of the column is a routed mixture-of-experts module. For a token representation $u\in\mathbb{R}^{d}$, the gate computes
\begin{equation}
p=\mathrm{softmax}(W_Gu)\in\mathbb{R}^{E},
\end{equation}
and selects the top-$k_E$ experts. Let $\mathcal{J}(u)$ denote the selected set. The normalized expert weights are
\begin{equation}
\widetilde{p}_e
=
\frac{p_e}{\sum_{j\in\mathcal{J}(u)}p_j},
\qquad e\in\mathcal{J}(u).
\end{equation}
Each expert is a SwiGLU map
\begin{equation}
\mathcal{E}_e(u)
=
W_{2,e}\!\left(\mathrm{SiLU}(W_{1,e}u)\odot W_{3,e}u\right).
\end{equation}
The MoE output is
\begin{equation}
\mathrm{MoE}(u)
=
\sum_{e\in\mathcal{J}(u)} \widetilde{p}_e\,\mathcal{E}_e(u)
+
\mathcal{E}_{\mathrm{shared}}(u),
\end{equation}
where the shared expert term is omitted when no shared expert is used.

Let $N=BT$ be the number of tokens in the batch. The top-1 load and average gate importance are
\begin{align}
\mathrm{load}_e
&=
\frac{1}{N}\sum_{n=1}^{N}\mathbf{1}\!\left[e=\arg\max_j p_{n,j}\right],\\
\mathrm{imp}_e
&=
\frac{1}{N}\sum_{n=1}^{N} p_{n,e}.
\end{align}
The load balancing penalty contributed by one column is
\begin{equation}
A_{\mathrm{lb}}^{(\ell)}
=
E\sum_{e=1}^{E}\mathrm{load}_e\,\mathrm{imp}_e.
\end{equation}

The scaled MoE load-balancing regularizer is
\begin{equation}
\mathcal{L}_{\mathrm{lb}}
=
\lambda_{\mathrm{lb}}\sum_{\ell=1}^{L}A_{\mathrm{lb}}^{(\ell)}.
\end{equation}

\subsection{Thalamic router}
\label{app:thal-impl}

The thalamic router receives the layer-5 output
\[
C\in\mathbb{R}^{B\times T\times d}
\]
from one cortical column and produces a modulatory signal for the next column. The input is first compressed to thalamic width $r$:
\begin{equation}
Z_0 = \mathrm{RMSNorm}(CW_c)\in\mathbb{R}^{B\times T\times r}.
\end{equation}
A local pathway is always active:
\begin{equation}
Z_{\mathrm{loc}}=\mathrm{SiLU}(Z_0W_{\mathrm{loc}}).
\end{equation}

The diffuse pathway uses a strictly causal past average. For each token position $t$,
\begin{equation}
\mu^{<t}_{b,t}
=
\begin{cases}
0, & t=1,\\[4pt]
\dfrac{1}{t-1}\displaystyle\sum_{j=1}^{t-1} Z_{0,b,j,:}, & t>1.
\end{cases}
\end{equation}
The surprise statistic is
\begin{equation}
s_{b,t}
=
\frac{1}{r}\left\|Z_{0,b,t,:}-\mu^{<t}_{b,t}\right\|_2^2.
\end{equation}
The diffuse branch is
\begin{equation}
Z_{\mathrm{diff}}=\mathrm{SiLU}(\mu^{<t}W_{\mathrm{diff}}).
\end{equation}
A state gate controls how much of the diffuse context enters the final thalamic state:
\begin{equation}
g_{\mathrm{state}}
=
\sigma\!\left(Z_0w_{\mathrm{state}} + b_{\mathrm{state}} + \alpha_s s\right),
\end{equation}
where $g_{\mathrm{state}}\in\mathbb{R}^{B\times T\times 1}$. With a learned global gain
\[
\beta_{\mathrm{diff}}=\sigma(a_{\mathrm{diff}}),
\]
the combined signal is
\begin{equation}
Z_1 = Z_{\mathrm{loc}} + \beta_{\mathrm{diff}}\,g_{\mathrm{state}}\odot Z_{\mathrm{diff}}.
\end{equation}

The next stage introduces feature-wise competition. A gate vector is produced by
\begin{equation}
g=\sigma(Z_1W_{\mathrm{trn}}+b_{\mathrm{trn}})\in(0,1)^{B\times T\times r}.
\end{equation}
If the feature dimension is divided into $G$ equal groups, let $d_g=r/G$ and write $g^{(m)}\in\mathbb{R}^{B\times T\times d_g}$ for the $m$-th group. The group mean is
\begin{equation}
\mu^{(m)}_{b,t}
=
\frac{1}{d_g}\sum_{j=1}^{d_g} g^{(m)}_{b,t,j},
\end{equation}
and the divisively normalized gate is
\begin{equation}
\widetilde{g}^{(m)}_{b,t,i}
=
\frac{g^{(m)}_{b,t,i}}{1+\eta\,\mu^{(m)}_{b,t}}.
\end{equation}
If grouping is not used, the same normalization is applied across all $r$ features. The competed thalamic state is
\begin{equation}
Z_2 = Z_1 \odot \widetilde{g}.
\end{equation}
Finally,
\begin{equation}
F_{\mathrm{thal}}
=
\left(Z_2W_{\mathrm{back}}\right)\odot \sigma(g_{\mathrm{mod}}),
\end{equation}
where $g_{\mathrm{mod}}\in\mathbb{R}^{d}$ is a learned channel-wise gate.

\subsection{Hippocampal memory}
\label{app:hippo-impl}

\subsubsection{Recent-window content retrieval}

The hippocampal memory stores keys
\[
K_{\mathrm{mem}}\in\mathbb{R}^{N_s\times d_k}
\]
and values
\[
V_{\mathrm{mem}}\in\mathbb{R}^{N_s\times d},
\]
together with a circular write pointer. Given the early-cortex state
\[
H^{(\ell_{\mathrm{inj}})}\in\mathbb{R}^{B\times T\times d},
\]
queries are
\begin{equation}
Q_{\mathrm{hip}} = H^{(\ell_{\mathrm{inj}})}W_Q^{(\mathrm{hip})}.
\end{equation}

The memory may contain fewer than $N_s$ valid entries. Let $n_{\mathrm{tot}}$ be the current number of valid slots and let
\[
n_{\mathrm{read}}=\min(n_{\mathrm{tot}},S_{\max})
\]
be the maximum number of slots inspected during retrieval. The reader always uses the most recent $n_{\mathrm{read}}$ items. If the memory is not yet full,
\begin{equation}
\mathcal{I}_{\mathrm{read}}
=
\{n_{\mathrm{tot}}-n_{\mathrm{read}},\dots,n_{\mathrm{tot}}-1\}.
\end{equation}
If the memory is full and the circular pointer is $p$, the recent window is
\begin{equation}
\mathcal{I}_{\mathrm{read}}
=
\left\{(p-n_{\mathrm{read}}+j)\bmod N_s:\ j=0,\dots,n_{\mathrm{read}}-1\right\}.
\end{equation}
This choice biases retrieval toward recent episodic content.

The similarity score for slot $i\in\mathcal{I}_{\mathrm{read}}$ is
\begin{equation}
a_{b,t,i}
=
\frac{\langle Q_{\mathrm{hip},b,t,:},K_{\mathrm{mem},i,:}\rangle}{\sqrt{d_k}}.
\end{equation}
The reader scans this recent window in slot chunks, retains the exact top-$k_H$ scores across all chunks, and only then applies the softmax. Let $\mathcal{N}_{b,t}$ be the selected slot indices. The retrieval weights are
\begin{equation}
\alpha_{b,t,i}
=
\frac{\exp(a_{b,t,i})}{\sum_{j\in\mathcal{N}_{b,t}}\exp(a_{b,t,j})},
\qquad i\in\mathcal{N}_{b,t},
\end{equation}
and the readout is
\begin{equation}
R_{b,t,:}
=
\sum_{i\in\mathcal{N}_{b,t}}\alpha_{b,t,i}V_{\mathrm{mem},i,:}.
\end{equation}
The final hippocampal output is
\begin{equation}
M_{b,t,:}
=
\left(R_{b,t,:}W_O^{(\mathrm{hip})}\right)\odot \sigma(g_{\mathrm{hip}}).
\end{equation}

\subsubsection{Intrinsic reward, predictor loss, and TD loss}

The hippocampus carries a fast predictor $f_{\theta}$, a slow predictor $f_{\bar{\theta}}$, a fast value head $v_{\psi}$, and a slow value head $v_{\bar{\psi}}$. The slow networks are exponential-moving-average copies of the fast networks.

Let
\[
X=\mathrm{stopgrad}\!\left(H^{(\ell_{\mathrm{inj}})}\right).
\]
For $t=1,\dots,T-1$, define normalized future activity
\begin{equation}
\bar{X}_{t+1}
=
\frac{X_{t+1}}{\|X_{t+1}\|_2+\epsilon}.
\end{equation}
The predictor outputs are
\begin{align}
P_t &= f_{\theta}(X_t), &
\bar{P}_t &= \frac{P_t}{\|P_t\|_2+\epsilon},\\
\bar{P}^{\,\mathrm{slow}}_t
&= \frac{f_{\bar{\theta}}(X_t)}{\|f_{\bar{\theta}}(X_t)\|_2+\epsilon}.
\end{align}
The corresponding cosine similarities with the next cortical state are
\begin{align}
c_t^{\mathrm{fast}} &= \langle \bar{P}_t,\bar{X}_{t+1}\rangle,\\
c_t^{\mathrm{slow}} &= \langle \bar{P}^{\,\mathrm{slow}}_t,\bar{X}_{t+1}\rangle.
\end{align}
The intrinsic reward is the positive learning-progress signal
\begin{equation}
r_t = \max\!\left(0,c_t^{\mathrm{fast}}-c_t^{\mathrm{slow}}\right).
\end{equation}
The raw prediction loss is
\begin{equation}
\mathcal{L}_{\mathrm{pred}}^{\mathrm{raw}}
=
\frac{1}{B(T-1)}
\sum_{b=1}^{B}\sum_{t=1}^{T-1}
\left(1-c_{b,t}^{\mathrm{fast}}\right).
\end{equation}

The fast and slow value estimates are
\begin{equation}
V_t=v_{\psi}(X_t),\qquad
\bar{V}_t=v_{\bar{\psi}}(X_t).
\end{equation}
The fast temporal difference residual is
\begin{equation}
\delta_t^{\mathrm{fast}}
=
\mathrm{clip}\!\left(r_t+\gamma V_{t+1}-V_t,\,-\delta_{\max},\,\delta_{\max}\right),
\end{equation}
which yields the critic objective
\begin{equation}
\mathcal{L}_{\mathrm{td}}
=
\frac{1}{2B(T-1)}
\sum_{b=1}^{B}\sum_{t=1}^{T-1}
\left(\delta_{b,t}^{\mathrm{fast}}\right)^2.
\end{equation}

The surprise score used for writing is derived from the slow residual,
\begin{equation}
\delta_t^{\mathrm{slow}}
=
\mathrm{clip}\!\left(r_t+\gamma \bar{V}_{t+1}-\bar{V}_t,\,-\delta_{\max},\,\delta_{\max}\right),
\end{equation}
through
\begin{equation}
s_t
=
\begin{cases}
0, & t=1,\\[3pt]
\left|\delta_{t-1}^{\mathrm{slow}}\right|, & t>1.
\end{cases}
\end{equation}

\subsubsection{Deferred write and adaptive threshold}

Memory writes are deferred until after backpropagation. During the forward pass, the hippocampus stores pending pairs
\[
\left(X,s_{1:T}\right),
\]
and these pairs are flushed only at the optimizer-step boundary. This ensures that every forward pass reads from the memory state that existed at the start of that pass.

For each sequence we select the top-$k_W$ surprise values. Let
\[
\mathcal{T}_b=\mathrm{TopK}\!\left(\{s_{b,t}\}_{t=1}^{T},k_W\right)
\]
denote the selected positions and let $n_{\mathrm{target}}$ be the desired number of writes per sequence. The target keep fraction is
\begin{equation}
\rho_{\mathrm{keep}}
=
\min\!\left(1,\frac{n_{\mathrm{target}}}{k_W}\right).
\end{equation}
Across all selected values in the batch, the candidate threshold is the $(1-\rho_{\mathrm{keep}})$ quantile,
\begin{equation}
\tau_{\mathrm{batch}}
=
\mathrm{Quantile}\!\left(\{s_{b,t}: t\in\mathcal{T}_b\},\,1-\rho_{\mathrm{keep}}\right).
\end{equation}
Under distributed training, this quantile is computed from the union of the selected values across all ranks, so every worker uses the same write threshold. The threshold is updated by exponential smoothing,
\begin{equation}
\tau \leftarrow \beta_{\tau}\tau + (1-\beta_{\tau})\tau_{\mathrm{batch}}.
\end{equation}
Only states with $s_{b,t}>\tau$ are written.

For each selected state, the write key and value are
\begin{equation}
k_{b,t}=W_K^{(\mathrm{write})}X_{b,t,:},\qquad
v_{b,t}=W_V^{(\mathrm{write})}X_{b,t,:}.
\end{equation}
These are inserted into the circular memory by
\begin{equation}
\mathrm{slot}(n)=\left(\mathrm{ptr}+n\right)\bmod N_s.
\end{equation}
In the main model, both $W_K^{(\mathrm{write})}$ and $W_V^{(\mathrm{write})}$ are held fixed, so the memory remains a nonparametric store over cortical features.

\subsubsection{Slow-target update}

After each optimizer step, the slow predictor and slow value head are updated by exponential moving average:
\begin{align}
\bar{\theta} &\leftarrow \alpha \bar{\theta} + (1-\alpha)\theta,\\
\bar{\psi} &\leftarrow \alpha \bar{\psi} + (1-\alpha)\psi.
\end{align}
This update is performed once per optimizer step rather than once per micro-step.

\subsection{Hippocampal feedback into late cortex}
\label{app:feedback-impl}

After the early cortical stack, the memory readout $M$ is converted into a fixed feedback term that enters all late columns. The gating input is
\begin{equation}
G_{\mathrm{in}}
=
\left[\mathrm{stopgrad}\!\left(H^{(\ell_{\mathrm{inj}})}\right);\,M\right]
\in\mathbb{R}^{B\times T\times 2d},
\end{equation}
and the gate itself is
\begin{equation}
G=\sigma\!\left(G_{\mathrm{in}}W_{\mathrm{gate}}+b_{\mathrm{gate}}\right).
\end{equation}
If channel sparsification is enabled, only the largest fraction $\rho_{\mathrm{top}}$ of channels at each token is retained:
\begin{equation}
G \leftarrow G\odot \mathrm{TopFracMask}(G,\rho_{\mathrm{top}}).
\end{equation}
The final hippocampal feedback is
\begin{equation}
F_{\mathrm{hip}}
=
\sigma(a_{\mathrm{hip}})
\left((G\odot M)W_{\mathrm{hip}\rightarrow\mathrm{thal}}\right).
\end{equation}
For every late layer, the modulatory signal injected into attention is the sum of the thalamic signal from the previous cortical column and this fixed hippocampal term.

\subsection{Replay memory and consolidation}
\label{app:replay-impl}

The hippocampus maintains two replay stores over raw token chunks of fixed length $L_R$: a recent ring buffer and a long-term reservoir. Given a batch
\[
x\in\{1,\dots,V\}^{B\times T},
\]
we keep the first
\[
m=\left\lfloor \frac{T}{L_R}\right\rfloor
\]
non-overlapping chunks and reshape them into
\begin{equation}
\mathcal{C}(x)\in\{1,\dots,V\}^{(Bm)\times L_R}.
\end{equation}
These chunks are written into the recent ring buffer by circular insertion. The long term store uses reservoir sampling. If the reservoir capacity is $N_{\mathrm{long}}$ and it has already seen $n$ chunks, the next chunk replaces slot
\[
r \sim \mathrm{Unif}\{0,\dots,n\}
\]
whenever $r<N_{\mathrm{long}}$.

Replay sampling occurs before the current batch is written into either store. This guarantees that the replay loss at step $s$ depends only on chunks observed before step $s$. Let $B_R$ be the replay batch size and $\rho_{\mathrm{long}}$ the fraction drawn from the long-term store. The sample counts are
\begin{align}
B_{\mathrm{long}} &= \mathrm{round}(B_R\rho_{\mathrm{long}}),\\
B_{\mathrm{recent}} &= B_R - B_{\mathrm{long}}.
\end{align}
The sampled replay tokens are concatenated into
\[
x^{\mathrm{rep}}\in\{1,\dots,V\}^{B_R\times L_R}.
\]
These chunks are passed through the full autoregressive model and optimized with standard next-token supervision:
\begin{equation}
\mathcal{L}_{\mathrm{rep}}
=
-\frac{1}{B_R(L_R-1)}
\sum_{b=1}^{B_R}\sum_{t=1}^{L_R-1}
\log p_{\theta}\!\left(x^{\mathrm{rep}}_{b,t+1}\mid x^{\mathrm{rep}}_{b,1:t}\right).
\end{equation}

\subsection{Replay controller}
\label{app:replay-controller}

The replay controller monitors the degradation on previously learned tasks and adjusts the replay strength online. Let $P_k(t)$ denote the training perplexity of task $k$ at step $t$, and let $P_k^{\mathrm{post}}$ be the value recorded immediately after training task $k$. In transformed space,
\begin{equation}
u_k(t)=\log P_k(t),\qquad
u_k^{\mathrm{post}}=\log P_k^{\mathrm{post}}.
\end{equation}
The mean forgetting signal is
\begin{equation}
\bar{f}(t)
=
\frac{1}{|\mathcal{K}_{\mathrm{past}}|}
\sum_{k\in\mathcal{K}_{\mathrm{past}}}
\max\!\left(0,u_k(t)-u_k^{\mathrm{post}}\right).
\end{equation}
If $P_{\mathrm{sel}}(t)$ is the selected perplexity used for normalization, the relative forgetting gap is
\begin{equation}
g(t)
=
\frac{\bar{f}(t)}{\max\!\left(1,\left|\log P_{\mathrm{sel}}(t)\right|\right)}.
\end{equation}
An exponential moving average is maintained:
\begin{equation}
\widetilde{g}(t)
=
(1-\beta)\widetilde{g}(t-1)+\beta g(t).
\end{equation}
The controller error and integral state are
\begin{align}
e(t) &= \max\!\left(0,\widetilde{g}(t)-g_{\mathrm{tar}}\right),\\
I(t) &= \min\!\left(I_{\max}, I(t-1)+e(t)\right).
\end{align}
The replay coefficient, replay batch size, and long term replay fraction are updated by
\begin{align}
\lambda_{\mathrm{rep}}(t)
&=
\mathrm{clip}\!\left(
\lambda_0+k_p e(t)+k_i I(t),\,
\lambda_{\min},\,\lambda_{\max}
\right),\\
B_R(t)
&=
\mathrm{clip}\!\left(
\mathrm{round}\!\left(B_0(1+k_B e(t))\right),\,
B_{\min},\,B_{\max}
\right),\\
\rho_{\mathrm{long}}(t)
&=
\mathrm{clip}\!\left(
\rho_0+k_{\rho}e(t),\,
0,\,1
\right).
\end{align}
These values are then written directly into the model state used for replay.

\paragraph{Controller update frequency.}
The replay controller is not updated at every optimizer step. Instead, it is invoked every
\[
N_{\mathrm{ctrl}} = 240
\]
optimizer steps. At each controller call, the perplexity signal is estimated on a small fixed number of training batches per seen task. In our main configuration, we use
\[
N_{\mathrm{ctrl\mbox{-}batches}} = 5.
\]
This keeps the control signal inexpensive while still providing a stable estimate of forgetting.

\subsection{Training objective}
\label{app:trainer-objective}

Let $\Omega=\{(b,t): y_{b,t}\neq \texttt{ignore\_index}\}$ be the set of labeled positions. The language-model loss is
\begin{equation}
\mathcal{L}_{\mathrm{LM}}
=
-\frac{1}{|\Omega|}
\sum_{(b,t)\in\Omega}
\log p_{\theta}(y_{b,t}\mid x_b).
\end{equation}
The total MoE regularizer is
\begin{equation}
\mathcal{L}_{\mathrm{lb}}
=
\sum_{\ell=1}^{L}\mathcal{L}_{\mathrm{lb}}^{(\ell)}.
\end{equation}
The full training objective used for \textsc{TRC}$^2$ is
\begin{equation}
\mathcal{L}
=
\mathcal{L}_{\mathrm{LM}}
+
\lambda_{\mathrm{router}}\mathcal{L}_{\mathrm{lb}}
+
\lambda_{\mathrm{td}}\mathcal{L}_{\mathrm{td}}
+
\lambda_{\mathrm{pred}}\mathcal{L}_{\mathrm{pred}}^{\mathrm{raw}}
+
\lambda_{\mathrm{rep}}\mathcal{L}_{\mathrm{rep}},
\end{equation}

The $\lambda_{\mathrm{lb}}$ is the internal MoE load balancing scale and $\lambda_{\mathrm{router}}$ is the outer trainer weight applied to that auxiliary term; the effective coefficient is their product. The replay coefficient $\lambda_{\mathrm{rep}}$ is updated online by the replay controller.

Training uses distributed data parallelism with mixed precision and gradient accumulation. The optimizer step, slow-target update, and pending-write flush happen only on the last micro-step of each accumulation cycle. The learning rate is scheduled by optimizer step rather than by micro-step.

\section{Training Protocol, Task Schedule, and Metrics}
\label{app:protocol}

This section collects the experimental setup used throughout the paper.

\subsection{Task schedule}
\label{app:hyperparams}

The continual learning schedule consists of three tasks:
\[
\text{C4}\rightarrow \text{WikiText-103}\rightarrow \text{GSM8K}.
\]
C4 uses the English subset of \texttt{allenai/c4} with the \texttt{train} split for training and the \texttt{validation} split for validation and final reporting. WikiText-103 uses the standard \texttt{train}, \texttt{validation}, and \texttt{test} splits. GSM8K uses the \texttt{train} split for optimization and the \texttt{test} split for both validation and final reporting.

The optimizer step budgets are
\begin{equation}
N_{\text{C4}}=10{,}000,\qquad
N_{\text{WikiText-103}}=10{,}000,\qquad
N_{\text{GSM8K}}=2{,}000.
\end{equation}
Hence,
\begin{equation}
N_{\mathrm{total}}=22{,}000.
\end{equation}

\subsection{Optimization setup}

In the main training configuration, we use 4 NVIDIA V100 GPUs with 32GB memory each, fp16 mixed precision with gradient scaling, AdamW with learning rate $2\times 10^{-4}$, weight decay $0.1$, and
\[
(\beta_1,\beta_2)=(0.9,0.95).
\]
The learning rate follows linear warmup for 1{,}000 optimizer steps and cosine decay thereafter. Gradient clipping uses threshold $1.0$.

The batch configuration is
\[
\text{batch size per GPU}=8,\qquad
\text{gradient accumulation}=4,\qquad
\text{world size}=4,\qquad
T=1024.
\]
The effective global batch is therefore
\begin{equation}
8\times 4\times 4 = 128
\end{equation}
sequences per optimizer step, or
\begin{equation}
128\times 1024 = 131{,}072
\end{equation}
tokens per optimizer step. Over $22{,}000$ optimizer steps, the total token budget is
\begin{equation}
22{,}000\times 131{,}072 = 2{,}883{,}584{,}000.
\end{equation}

\subsection{Implementation details and default hyperparameters}

Examples are converted to plain text with a dataset specific formatting function, tokenized with the specified HuggingFace tokenizer.

Unless otherwise noted, \textsc{TRC}$^2$ uses $h=8$ query heads, $h_{\mathrm{kv}}=4$ key-value heads, thalamic rank $r=64$, TRN sparsity $0.5$, $N_s=512$ episodic slots, hippocampal key dimension $d_k=128$, retrieval top-$k_H=4$, writes-per-sequence $k_W=8$, read chunk size $2048$, maximum inspected memory slots $S_{\max}=8192$, gate top-fraction $\rho_{\mathrm{top}}=0.125$, and slow-target EMA factor $\alpha=0.9995$. The MoE uses $E=8$ experts, top-$k_E=2$ routing, one shared expert, and load-balancing coefficient $10^{-2}$. The feed-forward width is $d_{\mathrm{ff}}=\mathrm{round_up}(8d/3,256)$. Dropout is $0$ and the rotary base is $10^4$.

The replay system uses a recent ring buffer of 2048 token chunks, a long-term reservoir of 16384 chunks, and replay chunk length $L_R=128$. The replay controller uses coefficient range $[0,0.15]$, base replay coefficient $0.05$, target relative forgetting gap $0.001$, proportional and integral gains $(k_p,k_i)=(1.0,0.05)$, EMA factor $0.7$, replay batch range $[2,6]$ with base value $4$, and long-term replay base fraction $0.5$ with gain $4.0$. The controller probe evaluates 5 batches per seen task in the default setup. In all experiments, periodic evaluation is performed only at optimizer step boundaries after the pending write queue has been flushed and the optimizer and slow-target updates have completed. This ensures that evaluation does not discard queued writes mid-accumulation

Loss weights are $\lambda_{\mathrm{router}}=1$, $\lambda_{\mathrm{td}}=0.1$, $\lambda_{\mathrm{pred}}=0.1$. Exact per-run YAML settings, including task specific split strings and controller cadence, are released with the code.

\subsection{Evaluation metrics}

Validation is performed periodically on all tasks seen so far. For each task $k$, we record four quantities: the baseline score before the task is ever trained, the score immediately before training task $k$, the score immediately after training task $k$, and the score at later evaluation steps. For perplexity, the tracker operates in log space:
\begin{equation}
u_k(t)=\log \mathrm{PPL}_k(t).
\end{equation}
If $u_k^{\mathrm{post}}$ is the value recorded right after finishing task $k$, the forgetting value is
\begin{equation}
F_k(t)=\max\!\left(0,u_k(t)-u_k^{\mathrm{post}}\right).
\end{equation}
Backward transfer is measured by averaging
\begin{equation}
\mathrm{BWT}(t)
=
\frac{1}{|\mathcal{K}_{\mathrm{past}}|}
\sum_{k\in\mathcal{K}_{\mathrm{past}}}
\left(u_k^{\mathrm{post}}-u_k(t)\right),
\end{equation}
where $\mathcal{K}_{\mathrm{past}}$ excludes the task currently being trained. Forward transfer compares the score just before training a task with its baseline score. For minimization metrics,
\begin{equation}
\mathrm{FWT}
=
\frac{1}{K}\sum_{k=1}^{K}
\left(u_k^{\mathrm{base}}-u_k^{\mathrm{pre}}\right).
\end{equation}
The area under the forgetting curve is computed by trapezoidal integration over optimizer steps. For a transformed metric $u_k(t)$ with post-task reference $u_k^{\mathrm{post}}$, instantaneous forgetting is
\begin{equation}
F_k(t)=
\begin{cases}
\max\!\left(0,u_k(t)-u_k^{\mathrm{post}}\right), & \text{for minimization metrics},\\[4pt]
\max\!\left(0,u_k^{\mathrm{post}}-u_k(t)\right), & \text{for maximization metrics}.
\end{cases}
\end{equation}
For perplexity we use the transformed metric
\begin{equation}
u_k(t)=\log \mathrm{PPL}_k(t).
\end{equation}
The reported AUFC values are time-normalized running averages,
\begin{equation}
\mathrm{AUFC}_k(t)=\frac{1}{t-t_0}\int_{t_0}^{t} F_k(s)\,ds,
\end{equation}
where $t_0$ is the first evaluation step at which forgetting is tracked. For BLEU and token accuracy, the implementation additionally normalizes instantaneous forgetting by task gain when that denominator is available; otherwise it falls back to the raw gap. Reported AUFC summaries average these per-task values over previously seen tasks, excluding the task currently being trained. In addition to perplexity, the evaluation pipeline reports teacher forced token accuracy, exact match, BLEU, chrF, and ROUGE.

\section{Full Results}
\label{app:full-main-results}

\begin{figure}[H]
    \centering
    \includegraphics[width=\textwidth]{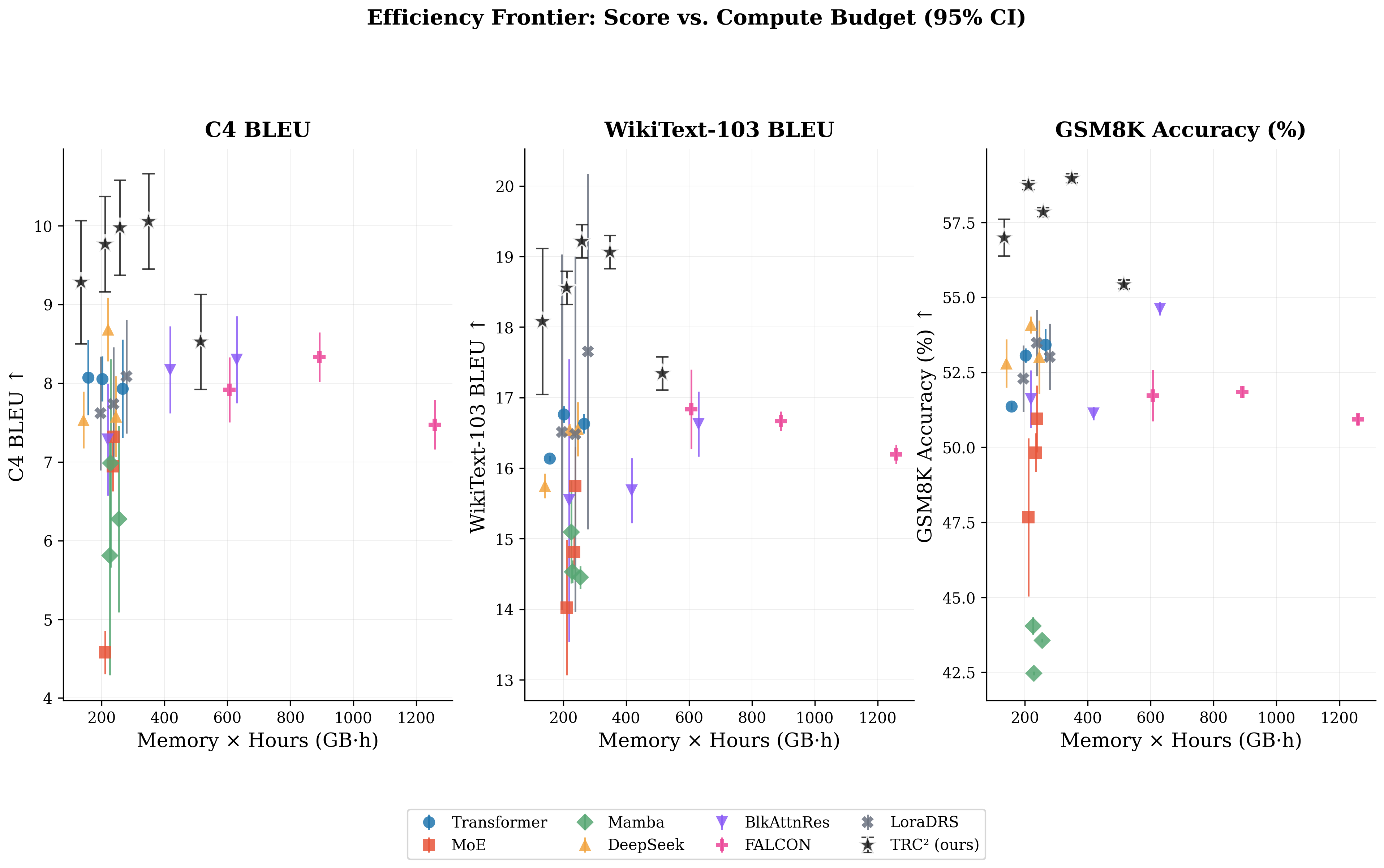}
    \caption{Efficiency frontier for task boundary quality versus compute budget, using the Memory$\times$Hour/GPU statistic reported in Table~\ref{tab:lm_boundary_scores}. Points show means across 3 seeds and error bars denote 95\% confidence intervals. Higher values are better on all three panels. The figure highlights the quality-efficiency tradeoff across architectures at the task boundaries.}
    \label{fig:efficiency_frontier2}
\end{figure}

This section reports the complete scalar results that are only summarized in the main text. Table~\ref{tab:lm_boundary_scores} gives the full task boundary comparison across all model families and all reported scales, including perplexity, teacher forced text scores, throughput, and compute cost. Table~\ref{tab:cl_aufc_20k_22k} gives the corresponding continual learning view through AUFC-based retention metrics at steps 20k and 22k. Figure~\ref{fig:efficiency_frontier2} reproduces the efficiency frontier for convenience and complements these tables by showing how task-boundary quality relates to compute budget.

Taken together, these components provide three aligned views of the same empirical picture. The task boundary table isolates endpoint modeling quality at each stage of the stream (Table~\ref{tab:lm_boundary_scores}); the retention table measures how much of that competence is preserved as training proceeds (Table~\ref{tab:cl_aufc_20k_22k}); and the efficiency frontier shows the systems cost at which those gains are obtained (Figure~\ref{fig:efficiency_frontier2}). Read jointly, they clarify that the advantage of \textsc{TRC}$^2$ is not confined to a single checkpoint or metric. It appears in boundary perplexity, in text based task scores, in cumulative forgetting, and in the resulting quality-compute tradeoff.
\begin{table*}[t]
  \centering
  \scriptsize
  \caption{Task boundary performance and efficiency compared across baselines. For C4 and WikiText-103, we report the last score at the corresponding task boundary, namely steps 10k and 20k. For GSM8K, we report the last score at the final task boundary, step 22k. Tokens/s is the last logged throughput value. We report BLEU at the C4 and WikiText-103 task boundaries, and token accuracy on GSM8K. Here $d$ denotes model width and $L$ the number of backbone blocks; for \textsc{TRC}$^2$ these blocks are cortical columns.}
  \label{tab:lm_boundary_scores}
  \begin{tabular}{lccccccccccc}
    \toprule
    \multirow{2}{*}{Model} & \multirow{2}{*}{Params} & \multirow{2}{*}{$d$} & \multirow{2}{*}{$L$} & \multicolumn{3}{c}{PPL $\downarrow$} & \multicolumn{3}{c}{Task-boundary text score $\uparrow$} & \multirow{2}{*}{Tokens/s $\uparrow$} & \multirow{2}{*}{$\frac{\mathrm{Mem}\times \mathrm{Hour}}{\mathrm{GPU}}\downarrow$} \\
    \cmidrule(lr){5-7}
    \cmidrule(lr){8-10}
    & & & & C4 & Wiki & GSM & C4 & Wiki & GSM (\%) & & \\
    \midrule
    Transformer & 144M & 512  & 28 & 93.17 & 33.83 & $1.7{\mathrm e}3$ & 7.90 & 16.15 & 51.42 & $\sim$ 93000  & $157\,\mathrm{GB\cdot h}$ \\
    Transformer & 228M & 768  & 20 & 76.27 & 29.94 & $2.8{\mathrm e}3$ & 8.06 & 16.78 & 53.09 & $\sim$ 84000  & $202\,\mathrm{GB\cdot h}$ \\
    Transformer & 321M & 1024 & 16 & 70.68 & 28.90 & $2.8{\mathrm e}3$ & 7.78 & 16.69 & 53.41 & $\sim$ 73000  & $266\,\mathrm{GB\cdot h}$ \\
    MoE         & 345M & 512  & 16 & 99.28 & 35.71 & $2.3{\mathrm e}3$ & 7.32 & 15.96 & 51.01 & $\sim$ 76000  & $238\,\mathrm{GB\cdot h}$ \\
    MoE         & 441M & 768  & 10 & 94.32 & 34.47 & $6.4{\mathrm e}3$ & 6.83 & 14.91 & 50.09 & $\sim$ 78000  & $235\,\mathrm{GB\cdot h}$ \\
    MoE         & 493M & 1024 & 6  & 136.04 & 36.75 & $1.8{\mathrm e}4$ & 4.70 & 13.69 & 47.46 & $\sim$ 91000  & $211\,\mathrm{GB\cdot h}$ \\
    Mamba       & 151M & 512  & 28 & 110.15 & 37.31 & $1.5{\mathrm e}4$ & 6.51 & 14.60 & 42.49 & $\sim$ 77000  & $229\,\mathrm{GB\cdot h}$ \\
    Mamba       & 202M & 768  & 18 & 97.53 & 34.50 & $4.9{\mathrm e}4$ & 5.85 & 14.52 & 43.59 & $\sim$ 76000  & $255\,\mathrm{GB\cdot h}$ \\
    Mamba       & 245M & 1024 & 12 & 95.25 & 33.01 & $9.2{\mathrm e}4$ & 5.92 & 15.41 & 44.01 & $\sim$ 84000  & $226\,\mathrm{GB\cdot h}$ \\
    DeepSeek    & 338M & 512  & 14 & 71.59 & 33.09 & $1.8{\mathrm e}3$ & 8.54 & 16.58 & 54.20 & $\sim$ 88000  & $220\,\mathrm{GB\cdot h}$ \\
    DeepSeek    & 308M & 768  & 6  & 80.84 & 32.75 & $6.9{\mathrm e}3$ & 7.53 & 15.75 & 52.80 & \textbf{$\sim$ 118000} & $142\,\mathrm{GB\cdot h}$ \\
    DeepSeek    & 545M & 1024 & 6  & 68.71 & 33.22 & $8.3{\mathrm e}3$ & 7.39 & 16.73 & 53.28 & $\sim$ 87000  & $246\,\mathrm{GB\cdot h}$ \\
    BlkAttnRes  & 144M & 512 & 28 & 91.94 & 36.05 & $2.2{\mathrm e}3$ & 7.97 & 15.87 & 51.22 & $\sim$ 55000  & $418\,\mathrm{GB\cdot h}$ \\
    BlkAttnRes  & 190M & 768 & 16 & 67.95 & 29.12 & $1.2{\mathrm e}3$ & 8.10 & 16.81 & 54.71 & $\sim$ 41000  & $630\,\mathrm{GB\cdot h}$ \\
    BlkAttnRes  & 187M & 1024 & 8 & 83.92 & 34.25 & $5.6{\mathrm e}3$ & 7.35 & 14.66 & 51.64 & $\sim$ 87000  & $219\,\mathrm{GB\cdot h}$ \\
    FALCON  & 144M & 512 & 32 & 80.04 & 32.09 & $3.5{\mathrm e}3$ & 7.36 & 16.25 & 51.01 & $\sim$ 19000  & $1259\,\mathrm{GB\cdot h}$ \\
    FALCON  & 193M & 768 & 20 & 67.27 & 30.02 & $7.8{\mathrm e}3$ & 8.22 & 16.72 & 51.93 & $\sim$ 27000  & $892\,\mathrm{GB\cdot h}$ \\
    FALCON  & 220M & 1024 & 12 & 64.06 & 29.30 & $12.9{\mathrm e}3$ & 7.74 & 16.92 & 51.54 & $\sim$ 35000  & $607\,\mathrm{GB\cdot h}$ \\
    LoraDRS  & 156M & 512 & 38 & 92.25 & 33.62 & $1.5{\mathrm e}3$ & 7.34 & 15.63 & 51.98 & $\sim$ 85000  & $196\,\mathrm{GB\cdot h}$ \\
    LoraDRS  & 252M & 768 & 30 & 74.47 & 29.54 & $2.3{\mathrm e}3$ & 7.81 & 16.77 & 52.71 & $\sim$ 70000  & $279\,\mathrm{GB\cdot h}$ \\
    LoraDRS  & 284M & 1024 & 18 & 75.03 & 28.51 & $4.3{\mathrm e}3$ & 7.46 & 15.60 & 53.17 & $\sim$ 79000  & $238\,\mathrm{GB\cdot h}$ \\
    \midrule
    TRC$^{2}$ (ours) & 149M & 256  & 24 & 52.48 & 27.59 & 56.36 & 8.31 & 17.44 & 55.49 & $\sim$ 29000  & $515\,\mathrm{GB\cdot h}$ \\
    TRC$^{2}$ (ours) & 254M & 512  & 10 & 37.43 & 23.09 & 42.99 & 9.76 & \textbf{19.31} & 57.90 & $\sim$ 61000  & $259\,\mathrm{GB\cdot h}$ \\
    TRC$^{2}$ (ours) & 226M & 768  & 4  & 38.06 & \textbf{21.73} & 31.28 & 9.28 & 18.11 & 57.19 & $\sim$ 103000 & \textbf{\boldmath$134\,\mathrm{GB\cdot h}$}\\
    TRC$^{2}$ (ours) & 408M & 768  & 8  & 34.29 & 25.04 & 35.70 & \textbf{9.84} & 19.16 & \textbf{59.03} & $\sim$ 59000  & $349\,\mathrm{GB\cdot h}$ \\
    TRC$^{2}$ (ours) & 560M & 1024 & 6  & \textbf{33.49} & 26.95 & \textbf{29.87} & 9.55 & 18.65 & 58.80 & $\sim$ 60000  & $211\,\mathrm{GB\cdot h}$ \\
    \bottomrule
  \end{tabular}
\end{table*}

\begin{table*}[t]
  \centering
  \scriptsize
  \caption{Continual learning performance at steps 20k and 22k. We report time-normalized area-under-forgetting-curve (AUFC) values. For perplexity, forgetting is measured in log-perplexity space; for BLEU and token accuracy, we additionally apply gain normalization when the required baseline denominator is available. Here $d$ denotes model width and $L$ the number of backbone blocks; for \textsc{TRC}$^2$ these blocks are cortical columns.}
  \label{tab:cl_aufc_20k_22k}
  \begin{tabular}{lccccccccccc}
    \toprule
    \multirow{2}{*}{Model} & \multirow{2}{*}{Params} & \multirow{2}{*}{$d$} & \multirow{2}{*}{$L$} & \multicolumn{2}{c}{PPL AUFC $\downarrow$} & \multicolumn{2}{c}{BLEU AUFC $\downarrow$} & \multicolumn{2}{c}{TokAcc AUFC $\downarrow$} & \multirow{2}{*}{Tokens/s $\uparrow$} & \multirow{2}{*}{$\frac{\mathrm{Mem}\times \mathrm{Hour}}{\mathrm{GPU}}\downarrow$} \\
    \cmidrule(lr){5-6}
    \cmidrule(lr){7-8}
    \cmidrule(lr){9-10}
    & & & & 20k & 22k & 20k & 22k & 20k & 22k & & \\
    \midrule
    Transformer & 144M & 512  & 28 & 1.16 & 1.15 & 2.21 & 1.33 & 0.21 & 0.15 & $\sim$ 93000  & $157\,\mathrm{GB\cdot h}$ \\
    Transformer & 228M & 768  & 20 & 1.23 & 1.30 & 2.16 & 1.30 & 0.20 & 0.15 & $\sim$ 84000  & $202\,\mathrm{GB\cdot h}$ \\
    Transformer & 321M & 1024 & 16 & 1.15 & 1.30 & 1.88 & 1.16 & 0.19 & 0.14 & $\sim$ 73000  & $266\,\mathrm{GB\cdot h}$ \\
    MoE         & 345M & 512  & 16 & 1.25 & 1.24 & 2.11 & 1.26 & 0.21 & 0.15 & $\sim$ 76000  & $238\,\mathrm{GB\cdot h}$ \\
    MoE         & 441M & 768  & 10 & 1.18 & 1.29 & 1.92 & 1.15 & 0.20 & 0.14 & $\sim$ 78000  & $235\,\mathrm{GB\cdot h}$ \\
    MoE         & 493M & 1024 & 6  & 1.11 & 1.22 & 1.22 & 0.74 & 0.20 & 0.14 & $\sim$ 91000  & $211\,\mathrm{GB\cdot h}$ \\
    Mamba       & 151M & 512  & 28 & 1.17 & 1.20 & 1.80 & 1.09 & 0.22 & 0.15 & $\sim$ 77000  & $229\,\mathrm{GB\cdot h}$ \\
    Mamba       & 202M & 768  & 18 & 1.12 & 1.33 & 1.51 & 0.94 & 0.21 & 0.15 & $\sim$ 76000  & $255\,\mathrm{GB\cdot h}$ \\
    Mamba       & 245M & 1024 & 12 & 1.12 & 1.34 & 1.50 & 0.93 & 0.21 & 0.15 & $\sim$ 84000  & $226\,\mathrm{GB\cdot h}$ \\
    DeepSeek    & 338M & 512  & 14 & 1.21 & 1.28 & 2.37 & 1.42 & 0.19 & 0.14 & $\sim$ 88000  & $220\,\mathrm{GB\cdot h}$ \\
    DeepSeek    & 308M & 768  & 6  & 1.22 & 1.32 & 2.08 & 1.26 & 0.20 & 0.14 & \textbf{$\sim$ 118000} & $142\,\mathrm{GB\cdot h}$ \\
    DeepSeek    & 545M & 1024 & 6  & 1.21 & 1.40 & 1.75 & 1.08 & 0.18 & 0.13 & $\sim$ 87000  & $246\,\mathrm{GB\cdot h}$ \\
    BlkAttnRes  & 144M & 512 & 28 & 1.19 & 1.07 & 0.21 & 0.15 & 0.28 & 0.20 & $\sim$ 55000  & $418\,\mathrm{GB\cdot h}$ \\
    BlkAttnRes  & 190M & 768 & 16 & 1.18 & 1.13 & 0.25 & 0.19 & 0.20 & 0.15 & $\sim$ 41000  & $630\,\mathrm{GB\cdot h}$ \\
    BlkAttnRes  & 187M & 1024 & 8 & 1.13 & 1.20 & 0.27 & 0.20 & 0.20 & 0.15 & $\sim$ 87000  & $219\,\mathrm{GB\cdot h}$ \\
    FALCON  & 144M & 512 & 32 & 1.14 & 1.27 & 1.47 & 0.94 & 0.19 & 0.14 & $\sim$ 19000  & $1259\,\mathrm{GB\cdot h}$ \\
    FALCON  & 193M & 768 & 20 & 1.15 & 1.37 & 2.09 & 1.29 & 0.19 & 0.14 & $\sim$ 27000  & $892\,\mathrm{GB\cdot h}$ \\
    FALCON  & 220M & 1024 & 12 & 1.13 & 1.41 & 1.79 & 1.13 & 0.18 & 0.14 & $\sim$ 35000  & $607\,\mathrm{GB\cdot h}$ \\
    LoraDRS  & 156M & 512 & 38 & 1.16 & 1.10 & 1.82 & 1.12 & 0.20 & 0.15 & $\sim$ 85000  & $196\,\mathrm{GB\cdot h}$ \\
    LoraDRS  & 252M & 768 & 30 & 1.17 & 1.24 & 1.72 & 1.09 & 0.19 & 0.14 & $\sim$ 70000  & $279\,\mathrm{GB\cdot h}$ \\
    LoraDRS  & 284M & 1024 & 18 & 1.13 & 1.28 & 1.73 & 1.06 & 0.19 & 0.14 & $\sim$ 79000  & $238\,\mathrm{GB\cdot h}$ \\
    \midrule
    TRC$^{2}$ (ours) & 149M & 256  & 24 & 0.71 & 0.49 & \textbf{0.18} & 0.12 & 0.13 & 0.08 & $\sim$ 29000  & $515\,\mathrm{GB\cdot h}$ \\
    TRC$^{2}$ (ours) & 254M & 512  & 10 & 0.69 & 0.49 & 0.21 & 0.13 & 0.13 & 0.09 & $\sim$ 61000  & $259\,\mathrm{GB\cdot h}$ \\
    TRC$^{2}$ (ours) & 226M & 768  & 4  & \textbf{0.63} & \textbf{0.44} & 0.20 & 0.12 & 0.13 & 0.08 & $\sim$ 103000 & \textbf{\boldmath$134\,\mathrm{GB\cdot h}$} \\
    TRC$^{2}$ (ours) & 408M & 768  & 8  & 0.71 & 0.50 & 0.19 & 0.12 & \textbf{0.13} & 0.08 & $\sim$ 59000  & $349\,\mathrm{GB\cdot h}$ \\
    TRC$^{2}$ (ours) & 560M & 1024 & 6  & 0.70 & 0.48 & 0.18 & \textbf{0.12} & 0.13 & \textbf{0.08} & $\sim$ 60000  & $211\,\mathrm{GB\cdot h}$ \\
    \bottomrule
  \end{tabular}
\end{table*}

\section{Additional Ablation and Component Analysis}
\label{app:ablation-components}

\begin{figure}[t]
    \centering

    \begin{subfigure}[t]{0.48\linewidth}
        \centering
        \includegraphics[width=\linewidth]{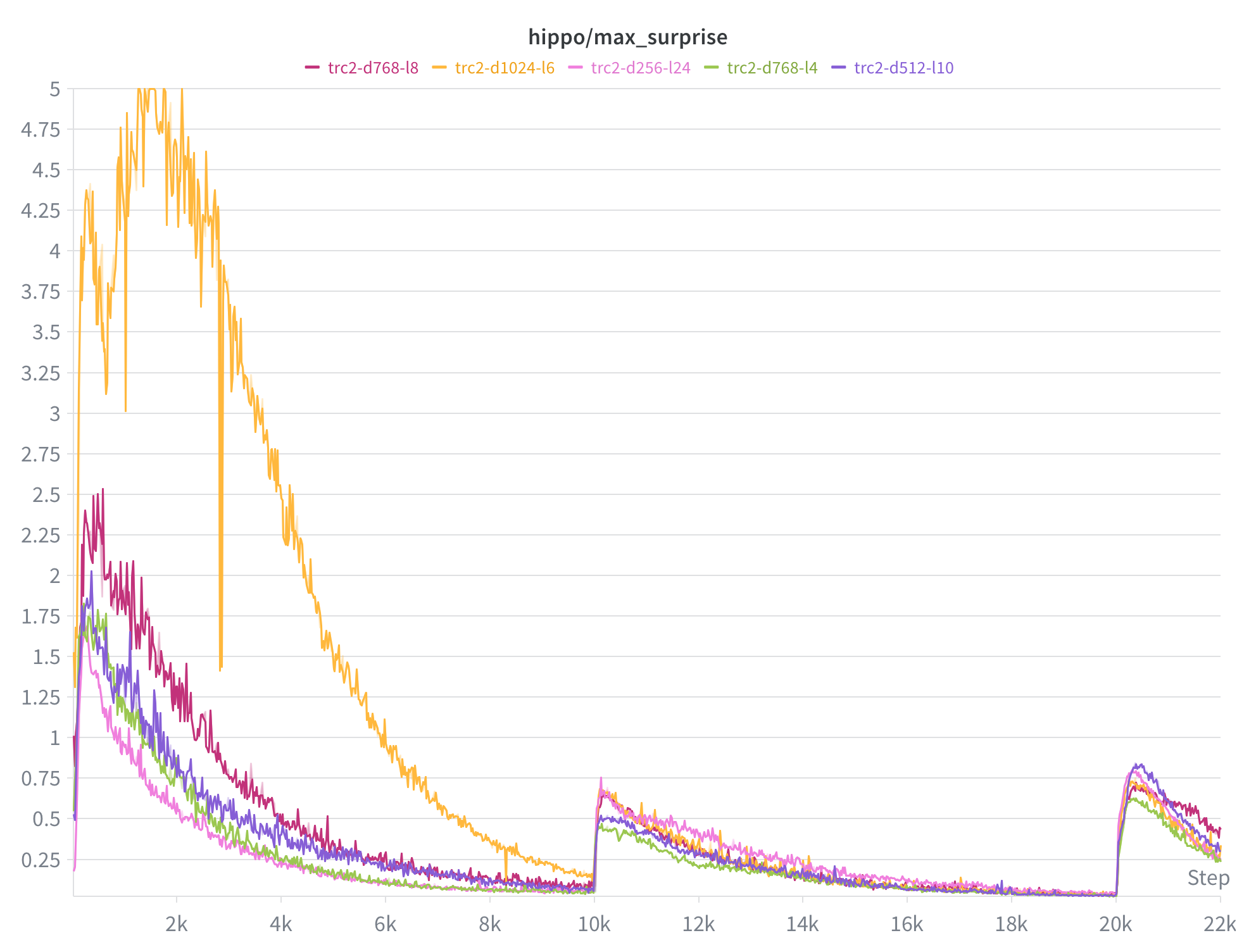}
    \end{subfigure}
    \hfill
    \begin{subfigure}[t]{0.48\linewidth}
        \centering
        \includegraphics[width=\linewidth]{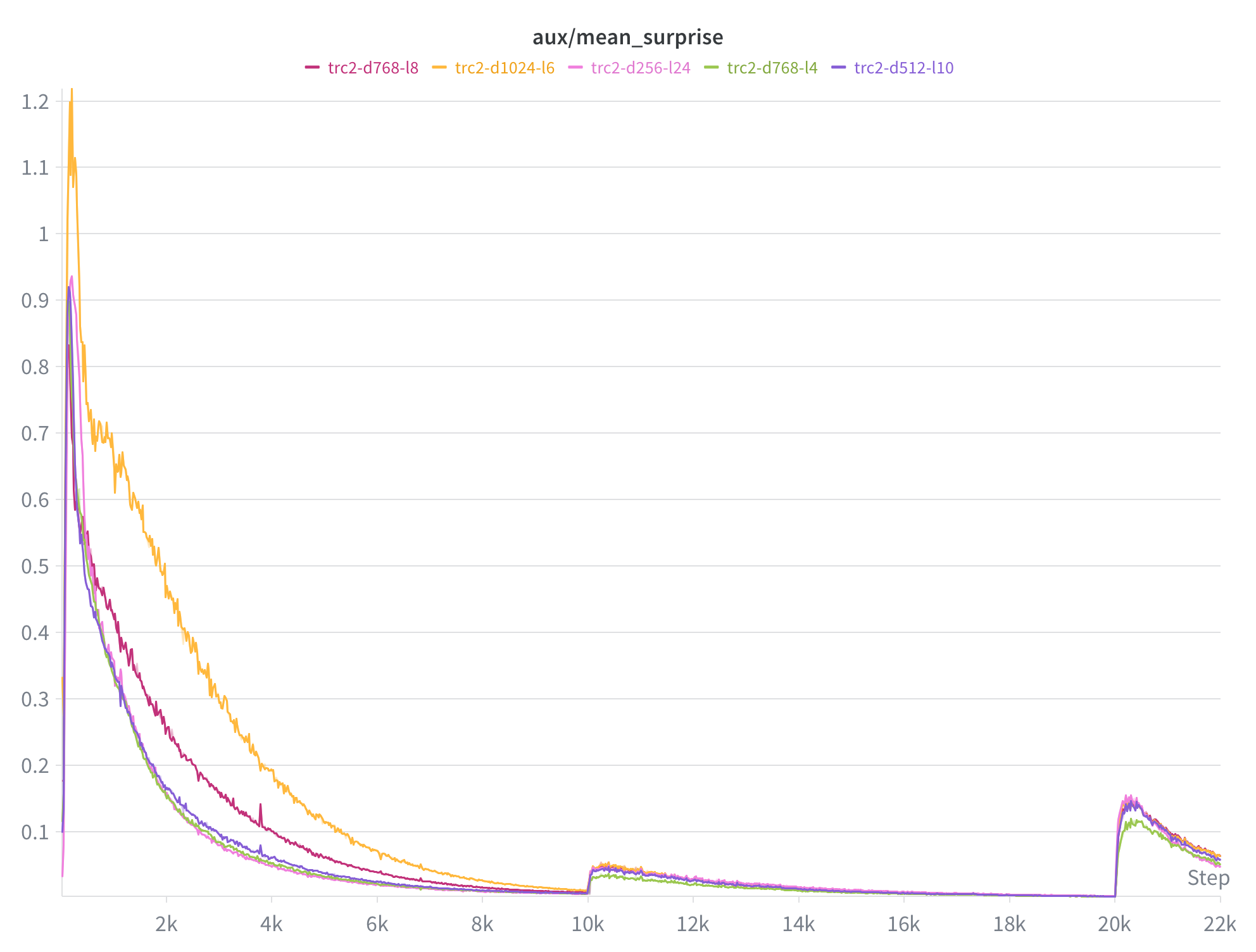}
    \end{subfigure}

    \vspace{0.8em}

    \begin{subfigure}[t]{0.48\linewidth}
        \centering
        \includegraphics[width=\linewidth]{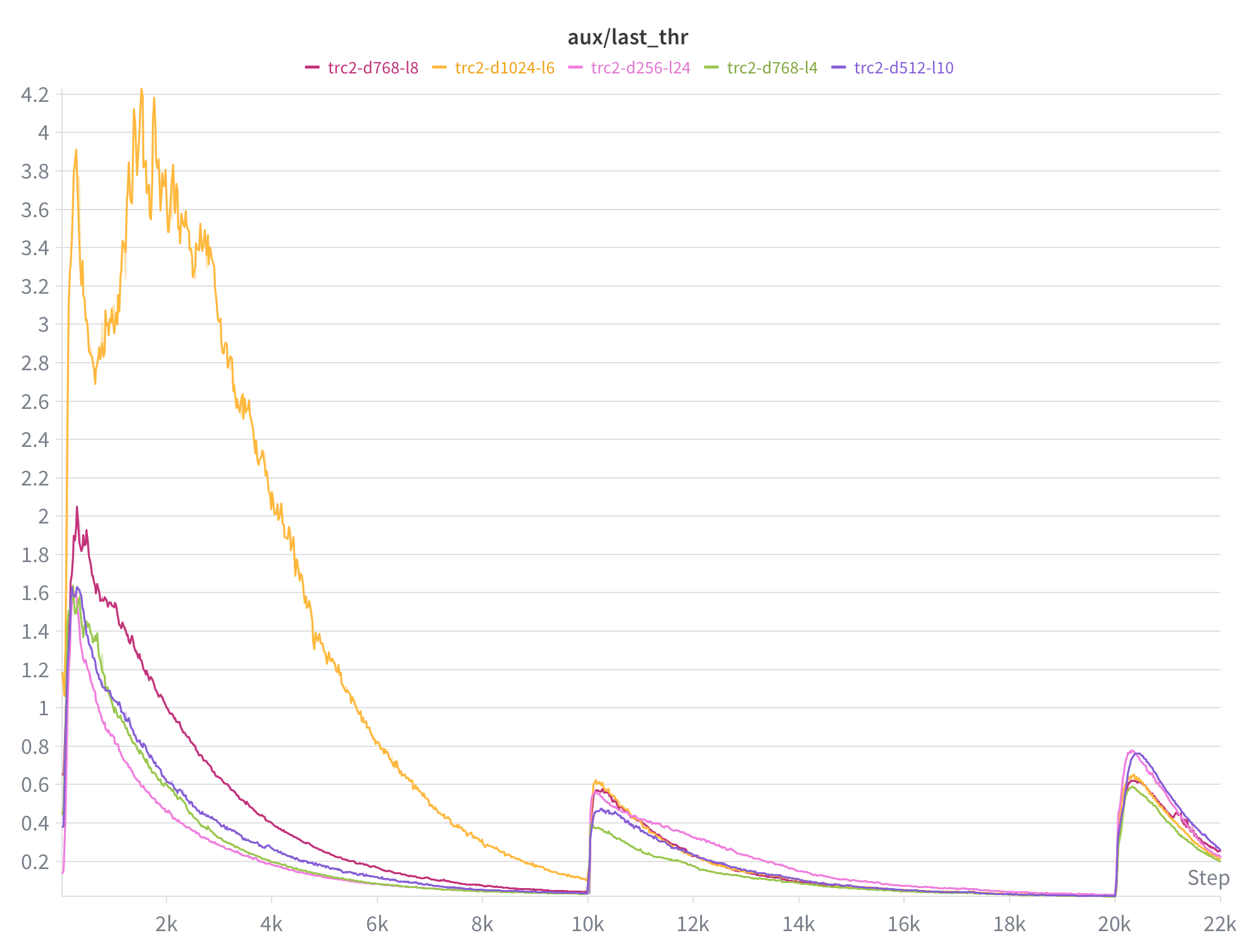}
    \end{subfigure}
    \hfill
    \begin{subfigure}[t]{0.48\linewidth}
        \centering
        \includegraphics[width=\linewidth]{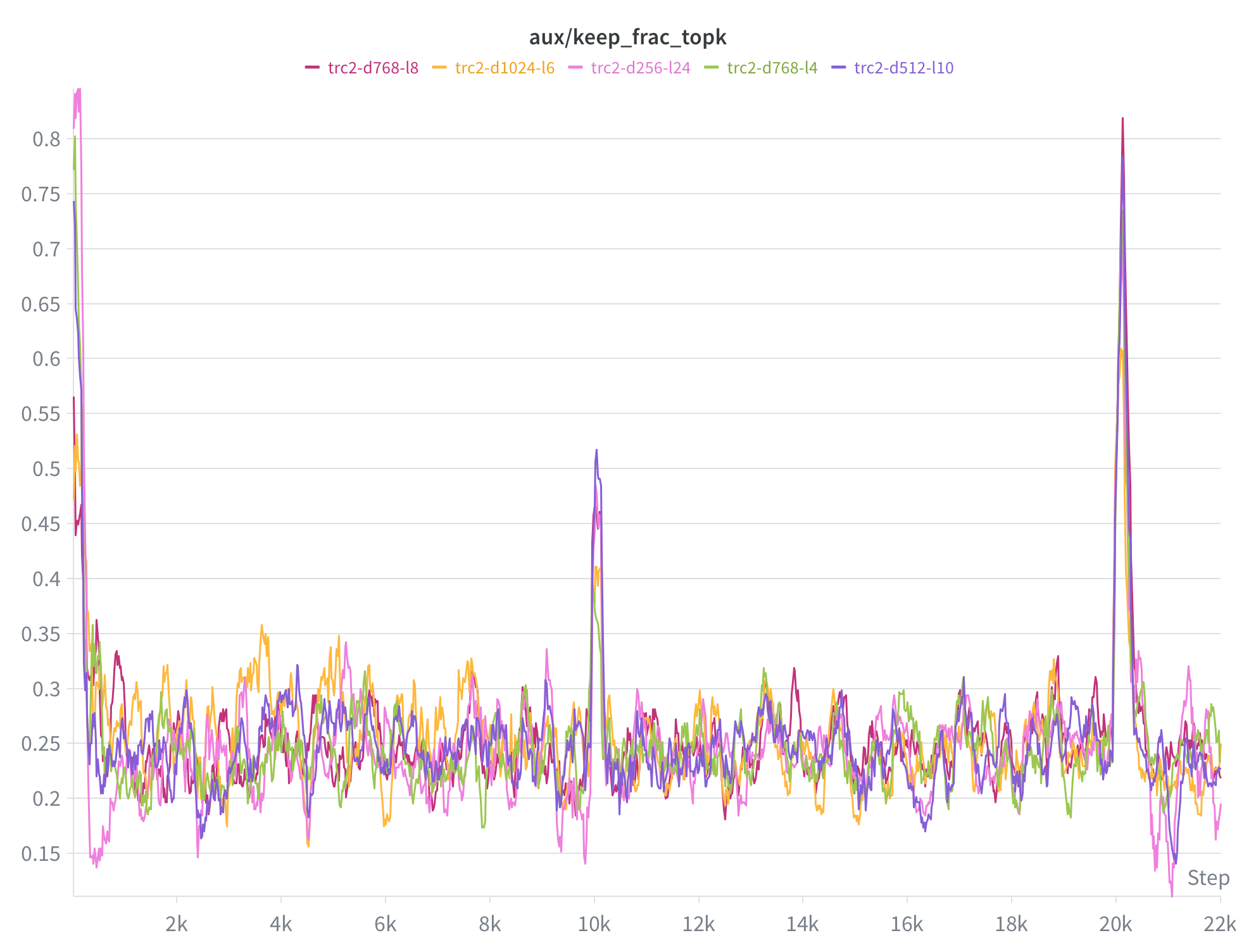}
    \end{subfigure}

    \vspace{0.8em}

    \begin{subfigure}[t]{0.48\linewidth}
        \centering
        \includegraphics[width=\linewidth]{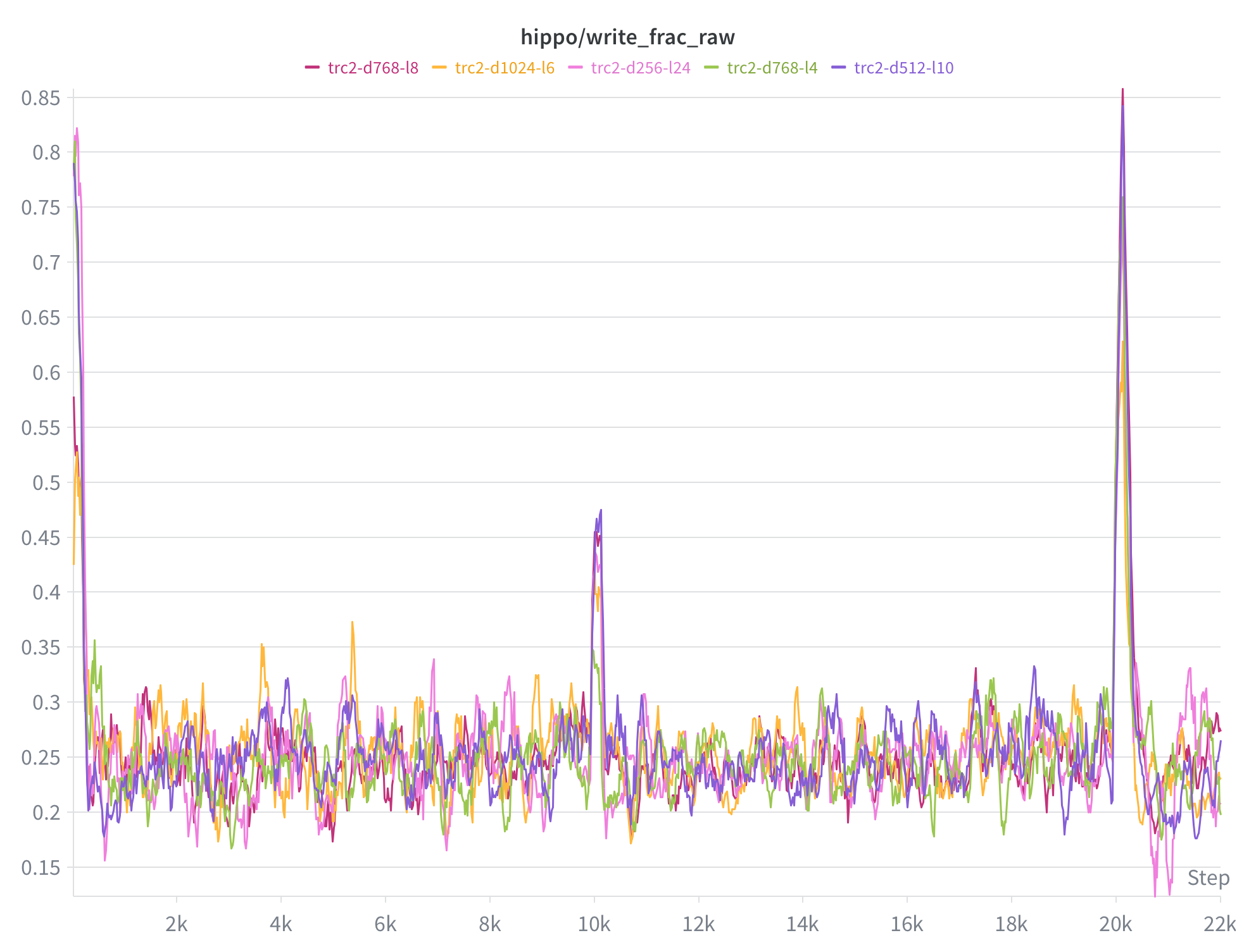}
    \end{subfigure}
    \hfill
    \begin{subfigure}[t]{0.48\linewidth}
        \centering
        \includegraphics[width=\linewidth]{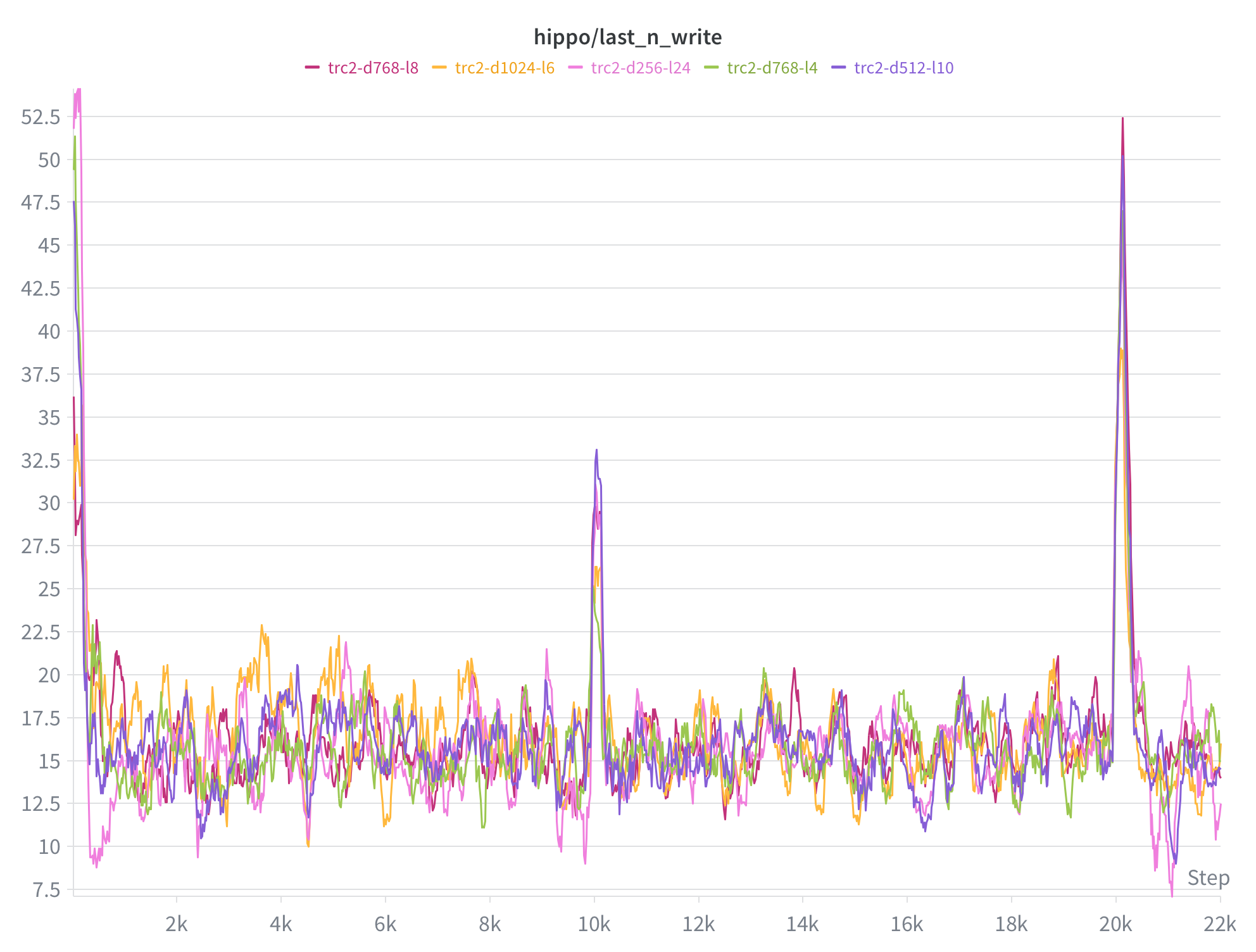}
    \end{subfigure}

    \caption{Hippocampal write dynamics across the task stream for \textsc{TRC}$^2$ variants. Vertical spikes in the traces mark task boundaries at $10$k and $20$k optimizer steps. Surprise statistics and the adaptive threshold $\tau$ rise after task changes and then relax within task, while the keep fraction and raw write fraction remain sparse for most of training.}
    \label{fig:app_trc2_hippo_dynamics}
\end{figure}

\begin{figure}[t]
    \centering

    \begin{subfigure}[t]{0.90\textwidth}
        \centering
        \includegraphics[width=\linewidth]{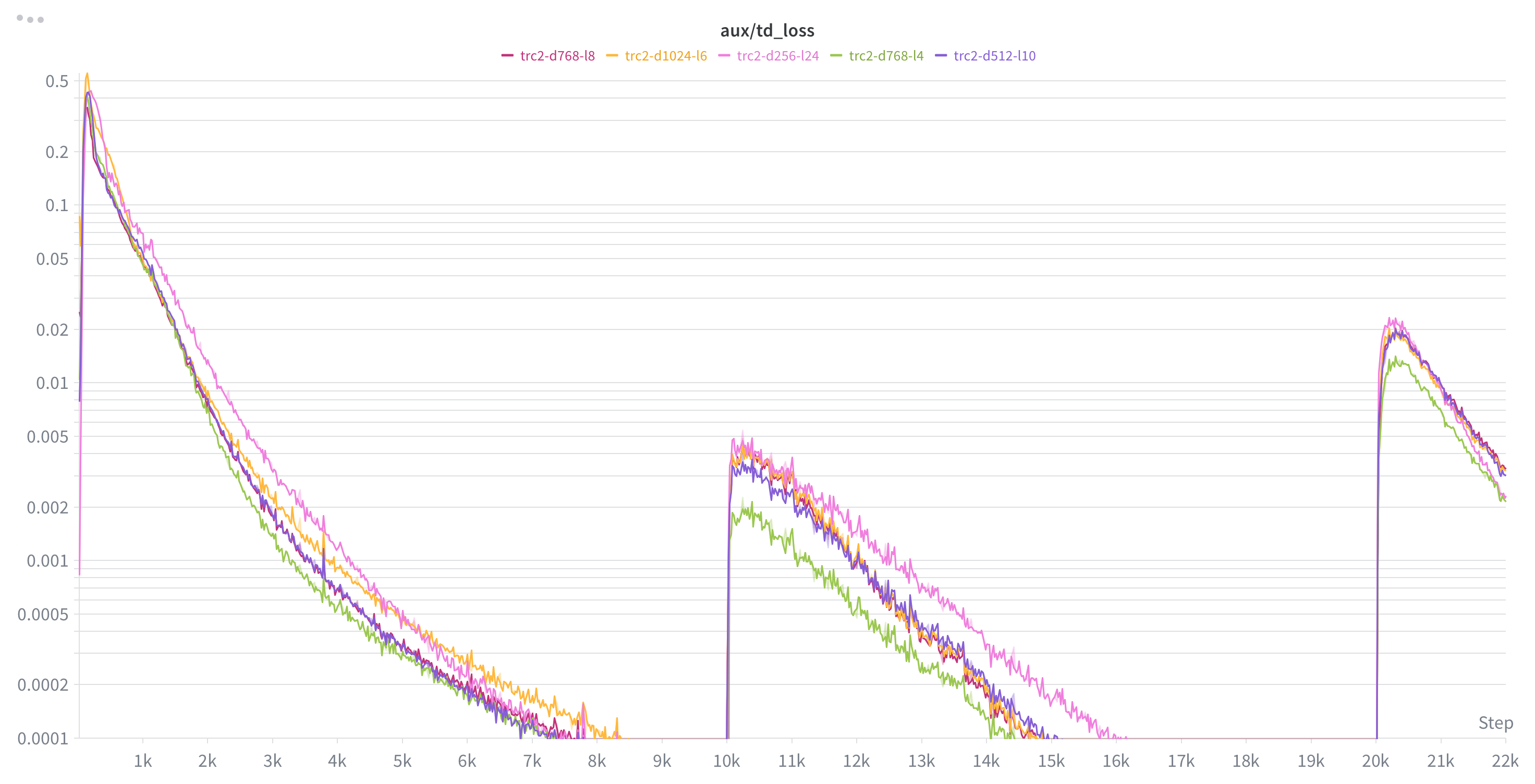}
        \caption{$\mathcal{L}_{\mathrm{td}}$.}
    \end{subfigure}

    \vspace{0.8em}

    \begin{subfigure}[t]{0.90\textwidth}
        \centering
        \includegraphics[width=\linewidth]{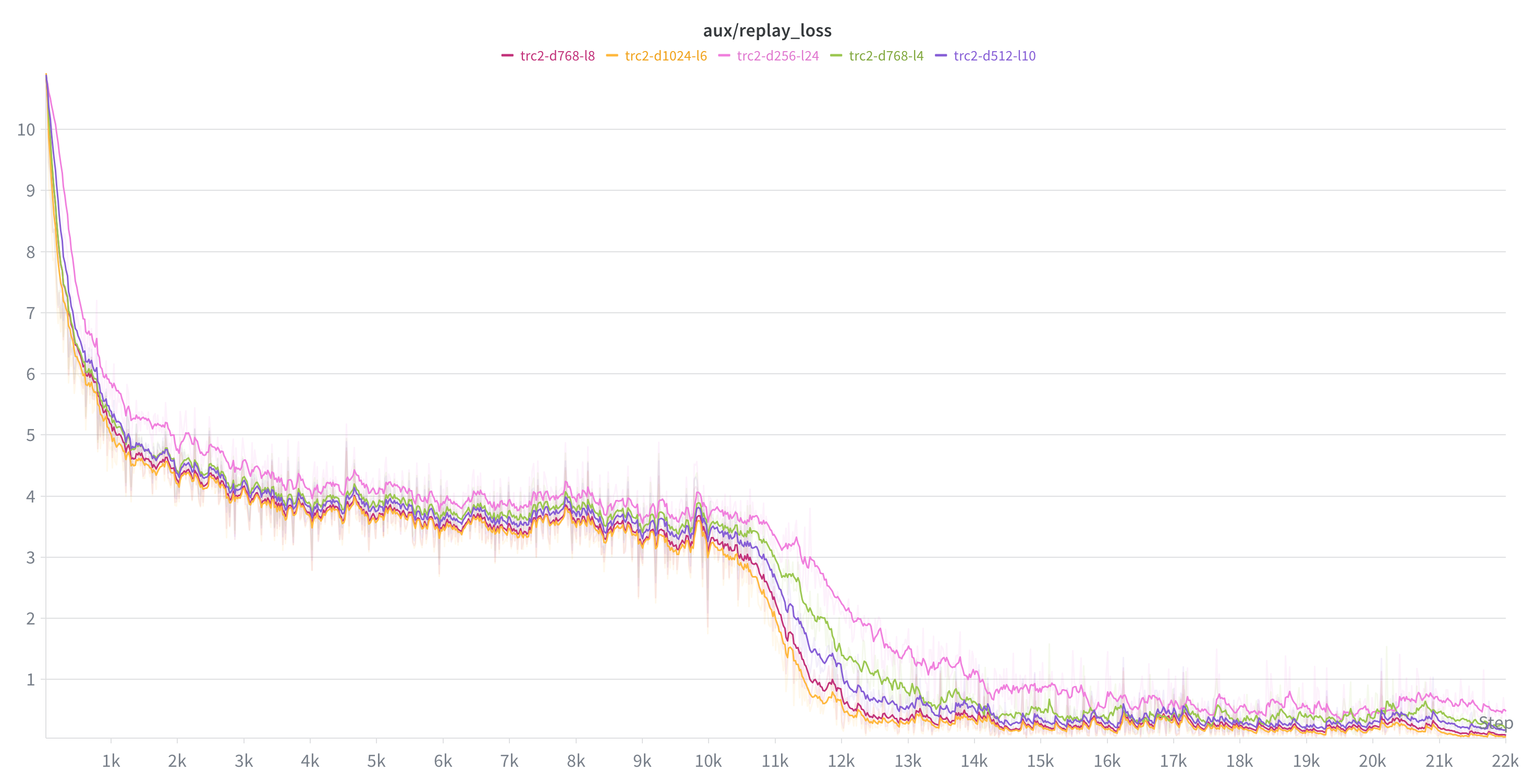}
        \caption{$\mathcal{L}_{\mathrm{rep}}$.}
    \end{subfigure}

    \vspace{0.8em}

    \begin{subfigure}[t]{0.90\textwidth}
        \centering
        \includegraphics[width=\linewidth]{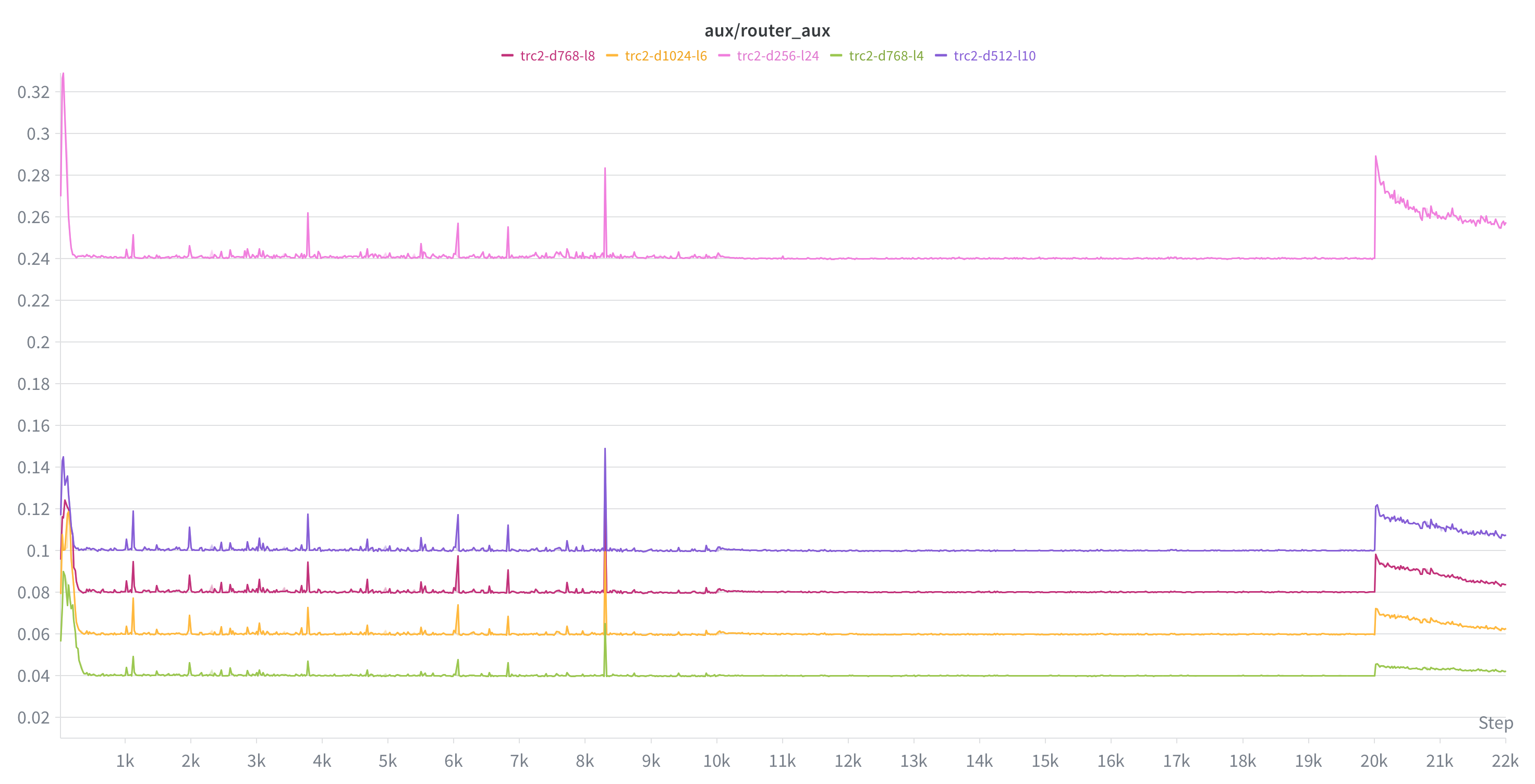}
        \caption{Router loss.}
    \end{subfigure}

    \caption{Additional auxiliary traces for \textsc{TRC}$^2$ variants. The logged router auxiliary is stable across the stream, while $\mathcal{L}_{\mathrm{td}}$ and $\mathcal{L}_{\mathrm{rep}}$ show transient increases but do not drift upward over time.}
    \label{fig:app_trc2_aux_dynamics}
\end{figure}

As a complement to the lesion study in Table~\ref{tab:ablation_trc2_d768_l4}, we inspected subsystem budgets and internal training dynamics. First, the lesion results are not driven by a large reallocation of parameters away from the cortical backbone: across all \textsc{TRC}$^2$ variants, the cortical columns contain between $97.94\%$ and $99.67\%$ of the logged parameters, while the hippocampal pathway contributes only $0.31\%$--$1.96\%$ and the thalamic pathway remains below $0.10\%$ (Table~\ref{tab:app_trc2_component_budget}). Second, the hippocampal controller remains sparse and well behaved during training: surprise scores and write thresholds rise briefly after distribution shifts, then relax within each task, while the effective write fraction stays far from saturation for almost all logged steps (Figures~\ref{fig:app_trc2_hippo_dynamics} and~\ref{fig:app_trc2_aux_dynamics}).

This accounting helps interpret the d768-l4 ablations in Table~\ref{tab:ablation_trc2_d768_l4}. For that configuration, the cortical columns account for $181.70$M of $185.51$M logged parameters ($97.94\%$), the hippocampus for $3.64$M ($1.96\%$), and the thalamus for only $0.176$M ($0.095\%$). In contrast, the logged component-memory split is less extreme: the same run allocates $0.679$ GB to the columns ($82.78\%$ of the logged component total), $0.141$ GB to the hippocampus ($17.14\%$), and $6.56\times 10^{-4}$ GB to the thalamus ($0.08\%$). This budget profile is consistent with the lesion results: removing the thalamus changes the parameter count only negligibly and yields a modest speedup, whereas removing the hippocampus eliminates a small parameter block that nonetheless carries a much larger memory footprint and an additional algorithmic path through retrieval, write selection, and replay.

\begin{table}[t]
  \centering
  \scriptsize
  \caption{Static subsystem budget for \textsc{TRC}$^2$ variants. The parameter columns report the number of trainable parameters assigned to the three logged subsystems: cortical columns, thalamic pathway, and hippocampal pathway. The memory columns report the corresponding static device footprint in GB, computed from trainable parameter storage together with persistent buffer storage, with shared tensors counted only once. The separate token embedding and output head group used in optimization is not included in this table. Percentages are computed relative to the sum of the three logged subsystems shown here.}
  \label{tab:app_trc2_component_budget}
  \begin{tabular}{lcccccc}
    \toprule
    \multirow{2}{*}{Variant} & \multicolumn{3}{c}{Parameters} & \multicolumn{3}{c}{Memory} \\
    \cmidrule(lr){2-4} \cmidrule(lr){5-7}
    & Col. (\%) & Hippo. (\%) & Thal. (\%) & Col. (\%) & Hippo. (\%) & Thal. (\%) \\
    \midrule
    TRC$^{2}$ d256-l24  & 99.67 & 0.31 & 0.01 & 88.52 & 11.46 & 0.01 \\
    TRC$^{2}$ d512-l10  & 99.24 & 0.72 & 0.03 & 89.22 & 10.75 & 0.03 \\
    TRC$^{2}$ d768-l4   & 97.94 & 1.96 & 0.09 & 82.78 & 17.14 & 0.08 \\
    TRC$^{2}$ d768-l8   & 98.96 & 0.99 & 0.05 & 90.58 & 9.37  & 0.04 \\
    TRC$^{2}$ d1024-l6  & 98.67 & 1.27 & 0.06 & 91.03 & 8.91  & 0.06 \\
    \bottomrule
  \end{tabular}
\end{table}

The internal traces further support the interpretation that the hippocampal pathway acts as a selective rather than dense plasticity channel. Figure~\ref{fig:app_trc2_hippo_dynamics} shows three consistent features across scales. First, the surprise statistics $s_t$ and the adaptive write threshold $\tau$ are highest immediately after entering a new task and then decay within task. The clearest reactivation occurs around the $20$k boundary, where the stream switches from WikiText-103 to GSM8K. Second, the effective write budget remains sparse. After the early warm-up transient, the logged keep fraction $\rho_{\mathrm{keep}}$ and the raw write fraction usually lie in the $0.1$--$0.4$ band rather than near $1$. For the d768-l4 model, the final logged values are $\rho_{\mathrm{keep}}=0.28125$, raw write fraction $=0.171875$, and $18$ writes at the last step; the maximum of $64$ writes is reached on only $5$ of the $1100$ logged checkpoints. Third, the scale trend is intuitive: the larger d1024-l6 model carries the largest early surprise and threshold values, but all variants converge toward substantially smaller within-task operating points after the initial transient.

The auxiliary losses tell a similar story. Figure~\ref{fig:app_trc2_aux_dynamics} shows that the critic loss $\mathcal{L}_{\mathrm{td}}$ decreases by orders of magnitude within each task and rises only transiently at task onsets, which is consistent with the temporal difference surprise signal stabilizing once the new regime has been absorbed. The replay loss $\mathcal{L}_{\mathrm{rep}}$ is largest during the earliest part of C4 and is much smaller over the later parts of the stream, suggesting that replay becomes easier to fit as the cortical backbone and episodic store mature. Finally, the logged router auxiliary remains bounded and nearly piecewise stationary within tasks. For the d768-l4 model, it stays in the $[0.0397, 0.0901]$ range over the full run and ends at $0.0421$, which indicates that the thalamic route does not require an increasingly aggressive auxiliary penalty in order to stay active.

Putting all above together, these traces sharpen the interpretation of the main ablation table. The retention gains of the full model do not come from moving a large fraction of parameters out of the cortical backbone. Instead, they arise from a small set of global pathways whose main effect is to change \emph{where} fast plasticity is expressed. The thalamic pathway adds very little parameter or memory mass but changes inter-column communication, whereas the hippocampal pathway adds only a small fraction of parameters yet a nontrivial memory and systems footprint. These appendix traces therefore support the main conclusion of the paper: the retention advantage of \textsc{TRC}$^2$ is architectural and dynamical rather than a simple consequence of scale.

\section{Causality and State Semantics Verification}
\label{app:causality}

We carried out an exact verification study of \textsc{TRC}$^2$ at the backbone level. The purpose of this study was to validate four properties central to the model design: strict autoregressive causality of the main logits path, causal assignment of hippocampal write scores, delayed write semantics under gradient accumulation, and persistence of episodic memory across evaluation calls.

Training calls use the signature
\[
(\mathrm{logits}, \mathrm{aux}) = \mathrm{CortexNet}(x,\mathrm{targets}=y),
\]
where hippocampal pending writes are created inside the forward pass in training mode, but are committed only when
\[
\mathrm{flush\_pending\_write}()
\]
is invoked at the optimizer step boundary. Switching the model to evaluation mode does not reset hippocampal memory. By contrast, an evaluation forward clears any pending write queue that may still exist, without committing those writes to memory. Replay buffers are written during training forward passes when targets are present, and they are not updated during evaluation. To isolate each claim, every check clones a fresh model with identical parameters and buffer state before applying a perturbation.

\paragraph{Test configuration.}
All tests reported below use a compact \textsc{TRC}$^2$ instance with
\[
d=128,\quad L=4,\quad h=8,\quad h_{\mathrm{kv}}=4,\quad E=4,
\]
thalamic bottleneck rank $16$, and $1024$ hippocampal slots. The model contains $8{,}520{,}244$ parameters. Stochasticity is disabled by setting dropout to zero.

\subsection{Verified properties}
\label{app:causality-properties}

\paragraph{A. Causality of the main logits path.}
We evaluate four complementary checks.

\textbf{Forward causality in evaluation mode.}
Given an input sequence $x_{1:T}$ and a position $t$, we form a perturbed sequence $\widetilde{x}_{1:T}$ that matches the original prefix up to position $t$ and replaces all tokens after $t$ by random values. We then compare the logits at position $t$:
\begin{equation}
\Delta_{\mathrm{eval}}(t)
=
\left\|
f_{\theta}(x_{1:T})_{:,t,:}
-
f_{\theta}(\widetilde{x}_{1:T})_{:,t,:}
\right\|_{\infty}.
\end{equation}
A causal model must satisfy $\Delta_{\mathrm{eval}}(t)\approx 0$ up to numerical precision.

\textbf{Gradient causality in evaluation mode.}
Let
\[
z_t=\sum_{v=1}^{V}\mathrm{logits}_{t,v}.
\]
We compute the gradient of $z_t$ with respect to the token embeddings and inspect the suffix positions:
\begin{equation}
\Gamma_{\mathrm{eval}}(t)
=
\max_{j>t}
\left\|
\frac{\partial z_t}{\partial e_j}
\right\|_{\infty}.
\end{equation}
Causality requires $\Gamma_{\mathrm{eval}}(t)=0$.

\textbf{Prefix consistency.}
We compare the logits at position $t$ from a full forward pass with the logits obtained by evaluating only the prefix $x_{1:t}$:
\begin{equation}
\Delta_{\mathrm{prefix}}(t)
=
\left\|
f_{\theta}(x_{1:T})_{:,t,:}
-
f_{\theta}(x_{1:t})_{:,-1,:}
\right\|_{\infty}.
\end{equation}
In a causal decoder these two quantities must match.

\textbf{Forward causality in training mode before flush.}
To verify that train mode itself does not introduce information leakage, we repeat the forward perturbation test with the model in training mode, but without flushing the pending write queue:
\begin{equation}
\Delta_{\mathrm{train,noflush}}(t)
=
\left\|
f_{\theta}^{\mathrm{train}}(x_{1:T})_{:,t,:}
-
f_{\theta}^{\mathrm{train}}(\widetilde{x}_{1:T})_{:,t,:}
\right\|_{\infty}.
\end{equation}

\paragraph{A1. Causality of the hippocampal write score.}
The hippocampus stores a surprise tensor
\[
s_{1:T}
\]
inside the pending-write queue. The score assigned to position $t$ should not depend on token $t+1$. We therefore compare the queued surprise tensors for two sequences that differ only at token $t+1$:
\begin{align}
\Delta_{\mathrm{score,prefix}}(t)
&=
\left\|
s(x)_{:,1:t}
-
s(\widetilde{x})_{:,1:t}
\right\|_{\infty},\\
\Delta_{\mathrm{score},t}
&=
\left\|
s(x)_{:,t}
-
s(\widetilde{x})_{:,t}
\right\|_{\infty},\\
\Delta_{\mathrm{score},t+1}
&=
\left\|
s(x)_{:,t+1}
-
s(\widetilde{x})_{:,t+1}
\right\|_{\infty}.
\end{align}
Causal write-score assignment requires
\[
\Delta_{\mathrm{score,prefix}}(t)\approx 0
\quad\text{and}\quad
\Delta_{\mathrm{score},t}\approx 0,
\]
while $\Delta_{\mathrm{score},t+1}$ may be nonzero.

\paragraph{B. Pending-write and evaluation semantics.}
We test the exact interaction between the pending write queue, evaluation calls, and gradient accumulation.

\textbf{No pre-flush effect.}
A training micro-step without optimizer update should create pending writes but should not alter the committed hippocampal memory. Let
\[
m_{\mathrm{pre}}
\]
be a model that has executed one training forward and backward pass without flush, and let
\[
m_{\mathrm{fresh}}
\]
be a fresh clone. For a probe sequence $x$, we measure
\begin{equation}
\Delta_{\mathrm{preflush}}(t)
=
\left\|
f_{\theta}^{m_{\mathrm{pre}}}(x)_{:,t,:}
-
f_{\theta}^{m_{\mathrm{fresh}}}(x)_{:,t,:}
\right\|_{\infty}.
\end{equation}
This difference should remain zero while the committed memory count stays unchanged.

\textbf{Evaluation clears pending writes without committing them.}
If the model enters evaluation mode while pending writes still exist, an evaluation forward should empty the pending queue but leave the memory store unchanged. We therefore verify
\begin{equation}
|\mathcal{Q}_{\mathrm{pending}}^{\mathrm{before}}| > 0,\qquad
|\mathcal{Q}_{\mathrm{pending}}^{\mathrm{after}}| = 0,
\end{equation}
together with
\begin{equation}
N_{\mathrm{mem}}^{\mathrm{before}} = N_{\mathrm{mem}}^{\mathrm{after}} = 0,
\end{equation}
and unchanged probe logits.

\textbf{Flush order under gradient accumulation.}
We mirror the two-micro-step training sequence used by the trainer. After the first micro-step, the model should contain pending writes but no committed hippocampal update:
\begin{equation}
N_{\mathrm{mem}}^{(1)} = 0.
\end{equation}
After the second micro-step, the trainer flushes the queue, performs the optimizer step, and updates the slow targets. At this point the pending queue must be empty and the memory count may increase:
\begin{equation}
|\mathcal{Q}_{\mathrm{pending}}^{(2)}| = 0,\qquad
N_{\mathrm{mem}}^{(2)} \ge 0.
\end{equation}

\paragraph{C. Persistence of hippocampal memory.}
We repeatedly run exact training steps until at least one episodic write is committed. We then switch the model to evaluation mode and verify that the memory count is unchanged:
\begin{equation}
N_{\mathrm{mem}}^{\mathrm{eval}} = N_{\mathrm{mem}}^{\mathrm{train}} > 0.
\end{equation}
We also compare the hippocampal readout at the real injection point against a fresh model:
\begin{equation}
\Delta_{\mathrm{read}}
=
\left\|
R_{\mathrm{written}} - R_{\mathrm{fresh}}
\right\|_{\infty},
\end{equation}
and compare the final logits:
\begin{equation}
\Delta_{\mathrm{logit,persist}}
=
\left\|
f_{\theta}^{\mathrm{written}}(x)-f_{\theta}^{\mathrm{fresh}}(x)
\right\|_{\infty}.
\end{equation}
A persistent memory should give $\Delta_{\mathrm{read}} > 0$, while the logits may differ only slightly at initialization.

\paragraph{D. Replay buffer semantics.}
The replay stores should be updated only during a training forward pass with targets present. We therefore record the recent and long term replay counts before and after a train forward:
\begin{equation}
N_{\mathrm{rep}}^{\mathrm{after}} \ge N_{\mathrm{rep}}^{\mathrm{before}},
\qquad
N_{\mathrm{long}}^{\mathrm{after}} \ge N_{\mathrm{long}}^{\mathrm{before}},
\end{equation}
and verify that evaluation leaves both counts unchanged:
\begin{equation}
N_{\mathrm{rep}}^{\mathrm{eval,after}} = N_{\mathrm{rep}}^{\mathrm{eval,before}},
\qquad
N_{\mathrm{long}}^{\mathrm{eval,after}} = N_{\mathrm{long}}^{\mathrm{eval,before}}.
\end{equation}

\paragraph{E. Target layout and graph coverage.}
Two additional checks are useful for exact reproduction.

First, the current forward path expects contiguous target tensors. Non-contiguous targets produced by slicing a larger tensor may trigger a reshape error. In practice, targets should be materialized as contiguous tensors before the forward call.

Second, we test whether every trainable parameter in the bare \textsc{TRC}$^2$ backbone receives a gradient in the exact autograd graph induced by the training objective. Let
\[
\mathcal{P}_{\mathrm{train}}
\]
be the set of trainable parameters and
\[
\mathcal{P}_{\mathrm{grad}}
\]
the subset whose gradients are not \texttt{None} after backpropagation. Perfect graph coverage requires
\begin{equation}
\mathcal{P}_{\mathrm{train}}=\mathcal{P}_{\mathrm{grad}}.
\end{equation}

\subsection{Observed results}
\label{app:causality-results}

\begin{figure}[t]
    \centering
    \includegraphics[width=0.85\linewidth,keepaspectratio]{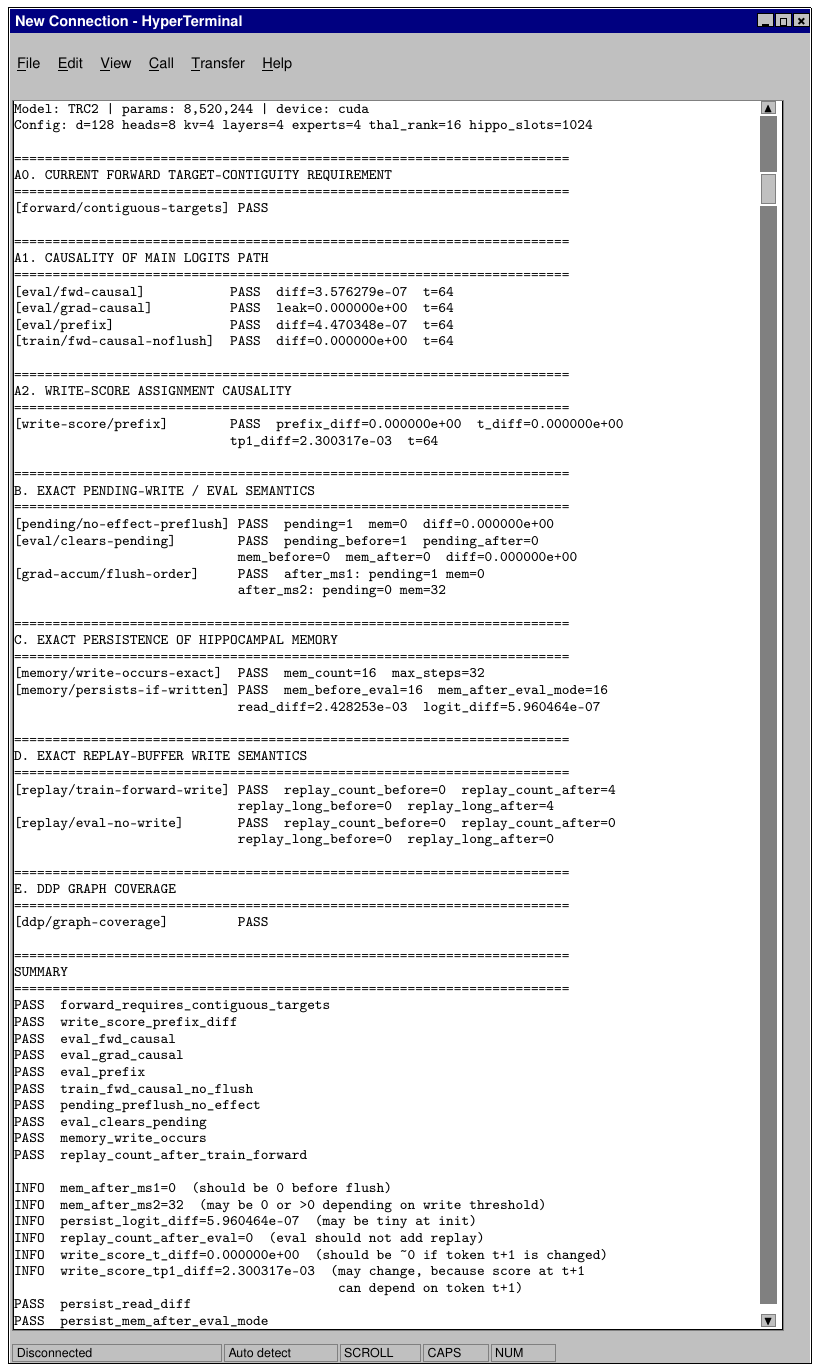}
    \caption{Verification summary for causality and state semantics in \textsc{TRC}$^2$. The checks cover forward causality, gradient causality, prefix consistency, delayed write behavior, memory persistence, and replay buffer update semantics.}
    \label{fig:app_causality_test}
\end{figure}

All causality checks on the main logits path pass at numerical precision (Figure \ref{fig:app_causality_test}). The write-score test confirms that changing token $t+1$ does not alter the score assigned to position $t$, while the score at position $t+1$ can change, which is the expected behavior. The state semantic tests confirm that pending writes do not affect the next forward pass before flush, that an evaluation forward clears the pending queue without committing memory, and that committed hippocampal memory persists across evaluation calls. Replay buffers are written during training forward passes with targets and remain unchanged during evaluation.

\paragraph{Interpretation.}
The results establish that the observable output path of \textsc{TRC}$^2$ is causally correct in both evaluation and training modes, provided that queued hippocampal writes have not yet been flushed. They also confirm that the hippocampal subsystem follows a strict delayed write discipline: reads occur during the forward pass, writes are committed only at the optimizer step boundary, and committed memory remains available during later evaluation calls. The replay store follows the intended training only update rule. 

\section{Complexity}
\label{app:complexity}

Let $B$ be the batch size, $T$ the sequence length, $d$ the model width, $L$ the number of cortical columns, $h$ the number of query heads, $h_{\mathrm{kv}}$ the number of key-value heads, $d_h=d/h$ the head width, $E$ the number of experts, $k_E$ the number of selected experts per token, $d_{\mathrm{ff}}$ the expert hidden width, $r$ the thalamic bottleneck width, $N_s$ the number of hippocampal memory slots, $d_k$ the hippocampal key width, $k_H$ the number of retrieved memory slots, and $S_{\max}$ the maximum number of recent memory slots inspected by the reader.

\paragraph{Cortical attention.}
For one cortical column, the attention projections contribute
\begin{equation}
O(BTd^2).
\end{equation}
The causal attention matrix products contribute
\begin{equation}
O(BT^2d),
\end{equation}
and the output projection adds another
\begin{equation}
O(BTd^2).
\end{equation}
Hence, one attention block costs
\begin{equation}
O(BT^2d + BTd^2).
\end{equation}

\paragraph{Mixture-of-experts.}
The gate computation contributes
\begin{equation}
O(BTdE).
\end{equation}
Since only $k_E$ experts are executed per token, the routed expert cost is
\begin{equation}
O(BT\,k_E\,d\,d_{\mathrm{ff}}).
\end{equation}
If a shared expert is present, it adds
\begin{equation}
O(BT\,d\,d_{\mathrm{ff}}^{(\mathrm{shared})}).
\end{equation}
The dominant MoE cost per column is therefore
\begin{equation}
O(BTdE + BT\,k_E\,d\,d_{\mathrm{ff}}).
\end{equation}

\paragraph{Thalamic router.}
For one inter-column thalamic call, the compression and decompression maps cost
\begin{equation}
O(BTdr).
\end{equation}
The local projection, diffuse projection, and TRN gate each contribute quadratic terms in the thalamic width,
\begin{equation}
O(BTr^2).
\end{equation}
The causal past-mean computation is linear in the sequence length and width,
\begin{equation}
O(BTr).
\end{equation}
The overall thalamic cost is therefore
\begin{equation}
O(BT(dr+r^2)).
\end{equation}

\paragraph{Hippocampal read.}
The query projection contributes
\begin{equation}
O(BTd\,d_k).
\end{equation}
If the reader inspects at most $S_{\max}$ recent memory slots, the similarity computation contributes
\begin{equation}
O(BT\,S_{\max}\,d_k).
\end{equation}
Selecting the top-$k_H$ entries and forming the weighted sum contribute
\begin{equation}
O(BT\,S_{\max}) + O(BT\,k_H\,d).
\end{equation}
The output projection contributes
\begin{equation}
O(BTd^2).
\end{equation}
Hence the dominant read cost is
\begin{equation}
O(BTd\,d_k + BT\,S_{\max}\,d_k + BTd^2).
\end{equation}
The chunked scan used by the reader does not change the arithmetic complexity, but it reduces peak memory by avoiding a full $(BT)\times S_{\max}$ score tensor.

\paragraph{Predictor and value heads.}
The fast predictor uses two width-$d$ linear maps, giving
\begin{equation}
O(BTd^2),
\end{equation}
and the slow predictor has the same forward cost but no gradient path. The value heads contribute only
\begin{equation}
O(BTd).
\end{equation}
Their cost is typically smaller than the main cortical stack.

\paragraph{Deferred write.}
If $W$ token states survive the adaptive write threshold during one optimizer step, the write projections cost
\begin{equation}
O(W\,d\,d_k) + O(W\,d^2).
\end{equation}
Since $W$ is controlled by the adaptive threshold and the target writes-per-sequence parameter, the write path scales with the number of selected events rather than with the full token count.

\paragraph{Replay consolidation.}
Replay adds an extra autoregressive forward and loss evaluation on a replay batch of shape $B_R\times L_R$. Its leading cost is that of another pass through the cortical stack:
\begin{equation}
O\!\left(
L\left(B_RL_R^2d + B_RL_Rd^2 + B_RL_Rk_E d d_{\mathrm{ff}}\right)
\right).
\end{equation}
Thus replay overhead grows linearly with the replay batch size and chunk length and is directly controlled by the replay controller.

\paragraph{Total forward cost.}
Ignoring lower-order terms, a forward pass of \textsc{TRC}$^2$ is dominated by the cortical stack,
\begin{equation}
O\!\left(
L\left(BT^2d + BTd^2 + BTk_E d d_{\mathrm{ff}}\right)
\right),
\end{equation}
with additional costs from thalamic routing,
\begin{equation}
O(L\,BT(dr+r^2)),
\end{equation}
and a single hippocampal read,
\begin{equation}
O(BTd\,d_k + BT\,S_{\max}\,d_k + BTd^2).
\end{equation}
Replay contributes an additional forward term on the sampled replay chunks whenever replay is active.

\paragraph{Memory footprint.}
The static episodic store requires
\begin{equation}
O(N_s(d_k+d))
\end{equation}
parameters in buffer form for keys and values. The replay buffers require
\begin{equation}
O\!\left((N_{\mathrm{recent}}+N_{\mathrm{long}})L_R\right)
\end{equation}
integer storage for token chunks. The remaining memory is dominated by standard transformer activations and optimizer states.

\section{Additional Aggregate Visualizations of Task Boundary Results}
\label{app:aggregate-task-boundary-viz}

\begin{figure*}[t]
    \centering
    \includegraphics[width=\textwidth]{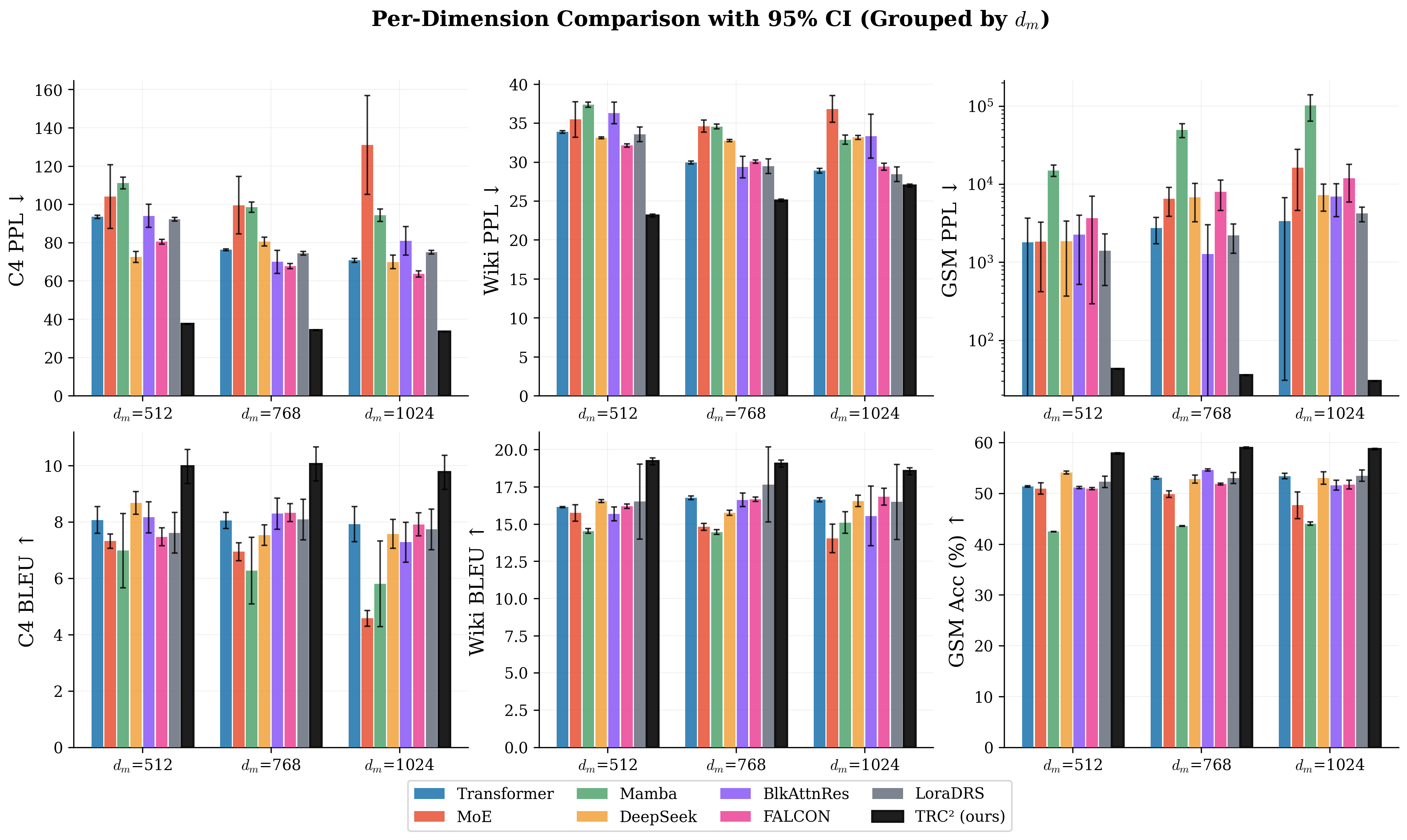}
    \caption{Per-dimension comparison of task boundary results grouped by model width $d_m$. Bars show means across 3 seeds and error bars show 95\% confidence intervals. The top row reports perplexity, lower is better, and the GSM8K panel is shown on a log scale. The bottom row reports task boundary text metrics, higher is better.}
    \label{fig:grouped_by_dm}
\end{figure*}

\begin{figure}[p]
    \centering
    \includegraphics[width=\textwidth]{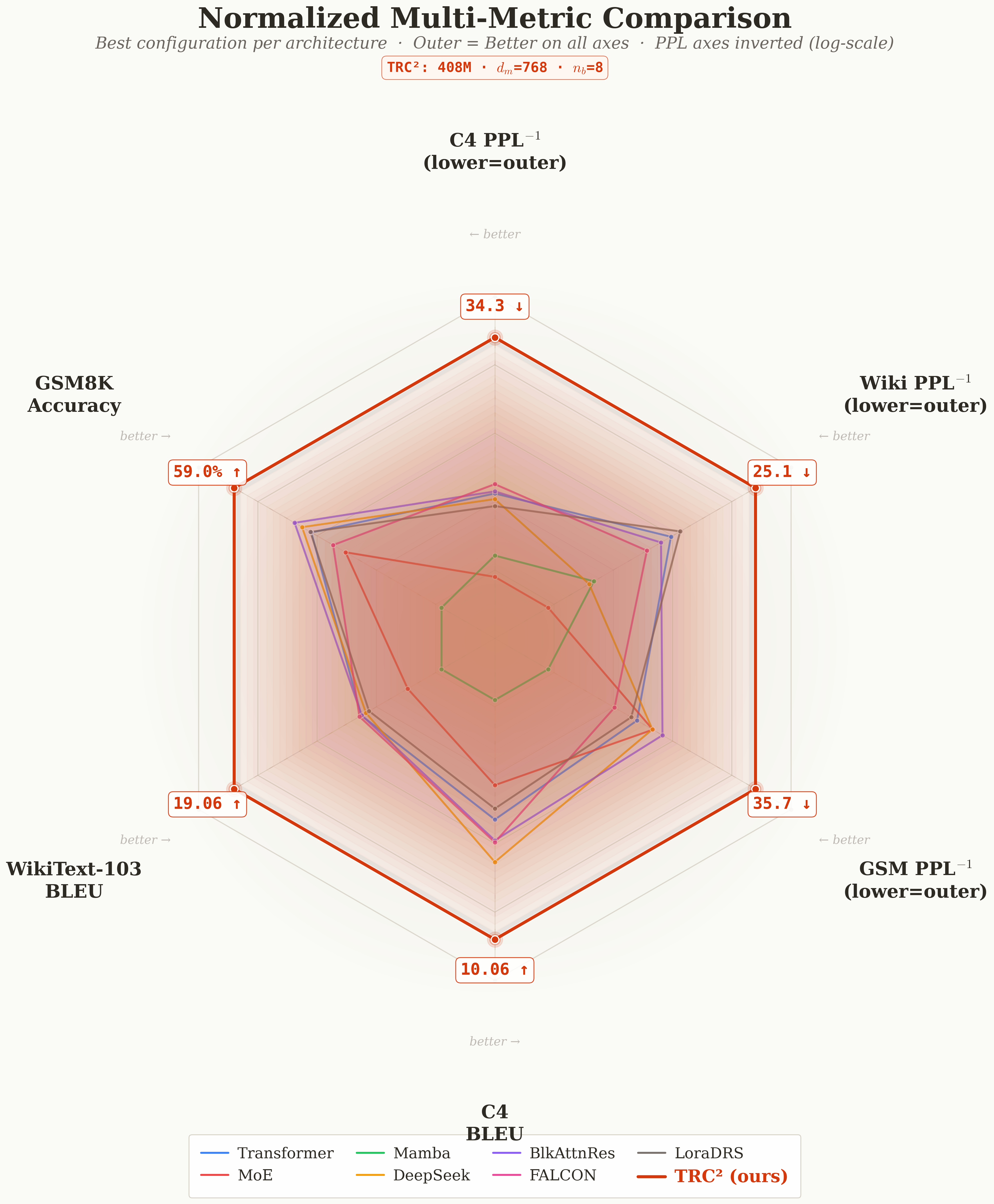}
    \caption{Normalized multi-metric comparison using the best selected configuration per architecture. Outer radius is better on all axes, with perplexity axes inverted after normalization so that larger values always indicate stronger performance. The highlighted polygon corresponds to the TRC$^{2}$ 408M configuration with $d_m=768$ and $n_b=8$. This figure is intended as a qualitative aggregate summary, since the normalization compresses absolute metric differences.}
    \label{fig:normalized_radar}
\end{figure}

\begin{sidewaysfigure*}
    \centering
    \includegraphics[width=\textheight,height=\textwidth,keepaspectratio]{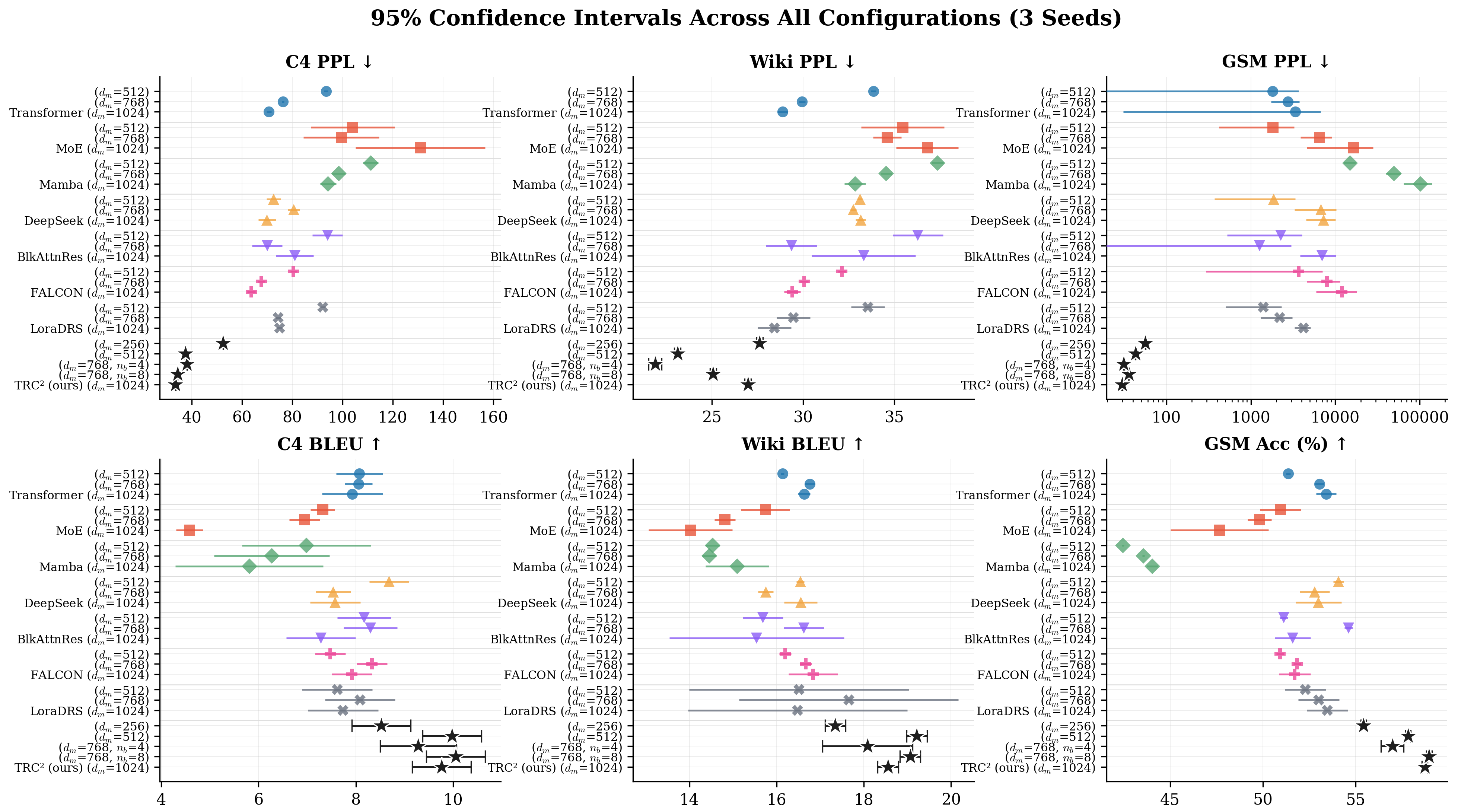}
    \caption{Configuration level 95\% confidence intervals across all task boundary results. Each point shows the mean over 3 seeds and horizontal bars denote 95\% confidence intervals. Lower values are better for perplexity, whereas higher values are better for BLEU and GSM8K token accuracy. This view exposes uncertainty at the individual configuration level and complements the aggregated scale based comparisons in the main text.}
    \label{fig:ci_forest_all_configs}
\end{sidewaysfigure*}

To complement Table~\ref{tab:lm_boundary_scores}, we provide three additional aggregate views of the same task boundary measurements. Figure~\ref{fig:grouped_by_dm} groups results by model width $d_m$, which helps separate architecture effects from width effects. Figure~\ref{fig:normalized_radar} summarizes the best configuration per architecture after metric normalization, and Figure~\ref{fig:ci_forest_all_configs} reports configuration level 95\% confidence intervals across all runs.

For Figure~\ref{fig:normalized_radar}, we select one configuration per architecture by computing the mean rank across the six task boundary metrics in Table~\ref{tab:lm_boundary_scores} (C4/WikiText-103/GSM8K perplexity, C4/WikiText-103 BLEU, and GSM8K token accuracy), with lower rank indicating better overall performance. The radar plot then visualizes that single selected configuration for each architecture after axis-wise normalization.

\section{Additional Aggregate Visualizations of Retention Results}
\label{app:aggregate-retention-viz}

\begin{figure}[t]
    \centering
    \includegraphics[width=\textwidth]{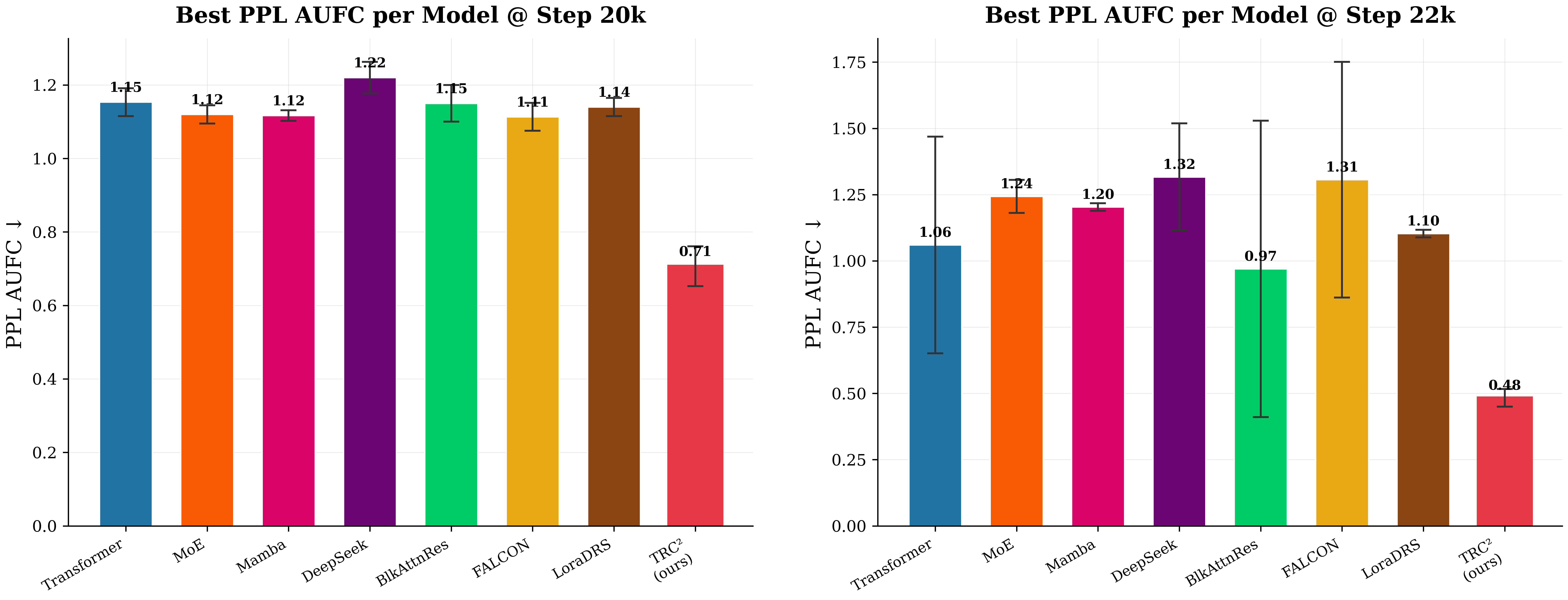}
    \caption{Best PPL AUFC per architecture at steps 20k and 22k. Each bar reports the best performing configuration for that model family, with lower values indicating better retention. Error bars denote 95\% confidence intervals across seeds. The summary highlights that \textsc{TRC}$^2$ attains the strongest retention profile at both checkpoints.}
    \label{fig:app_best_ppl_aufc_bar}
\end{figure}

Figure~\ref{fig:app_best_ppl_aufc_bar} summarizes the best perplexity AUFC achieved by each architecture at steps 20k and 22k. This per architecture view complements Table~\ref{tab:cl_aufc_20k_22k} by collapsing configuration level results into a single best setting summary.

\section{Order Robustness Under Task-Sequence Shuffling}
\label{app:order-robustness}

The main paper studies the continual learning stream
\[
\text{C4} \rightarrow \text{WikiText-103} \rightarrow \text{GSM8K},
\]
with per-task step budgets fixed at $10$k, $10$k, and $2$k, respectively. To test whether the retention advantage of \textsc{TRC}$^2$ depends on this particular ordering, we additionally evaluate the remaining five permutations of the same three-task stream while keeping the architecture, optimizer, tokenizer, evaluation protocol, and task-specific step budgets unchanged. Throughout this section, we use the shorthand
\[
C=\text{C4},\qquad W=\text{WikiText-103},\qquad G=\text{GSM8K},
\]
with the same budgets as in the main experiments:
\[
N_C = 10{,}000,\qquad N_W = 10{,}000,\qquad N_G = 2{,}000.
\]

These shuffled-order experiments provide a direct test of whether the proposed architecture improves retention only in the canonical stream or whether it continues to do so when the point of interference is moved. This is a meaningful stress test because the three tasks differ substantially in corpus style, scale, and difficulty, and changing the order changes which task is revisited late, which task induces the strongest transition, and which task becomes most vulnerable to cumulative forgetting.

\begin{sidewaysfigure*}[p]
    \centering
    \includegraphics[width=\textheight,height=\textwidth,keepaspectratio]{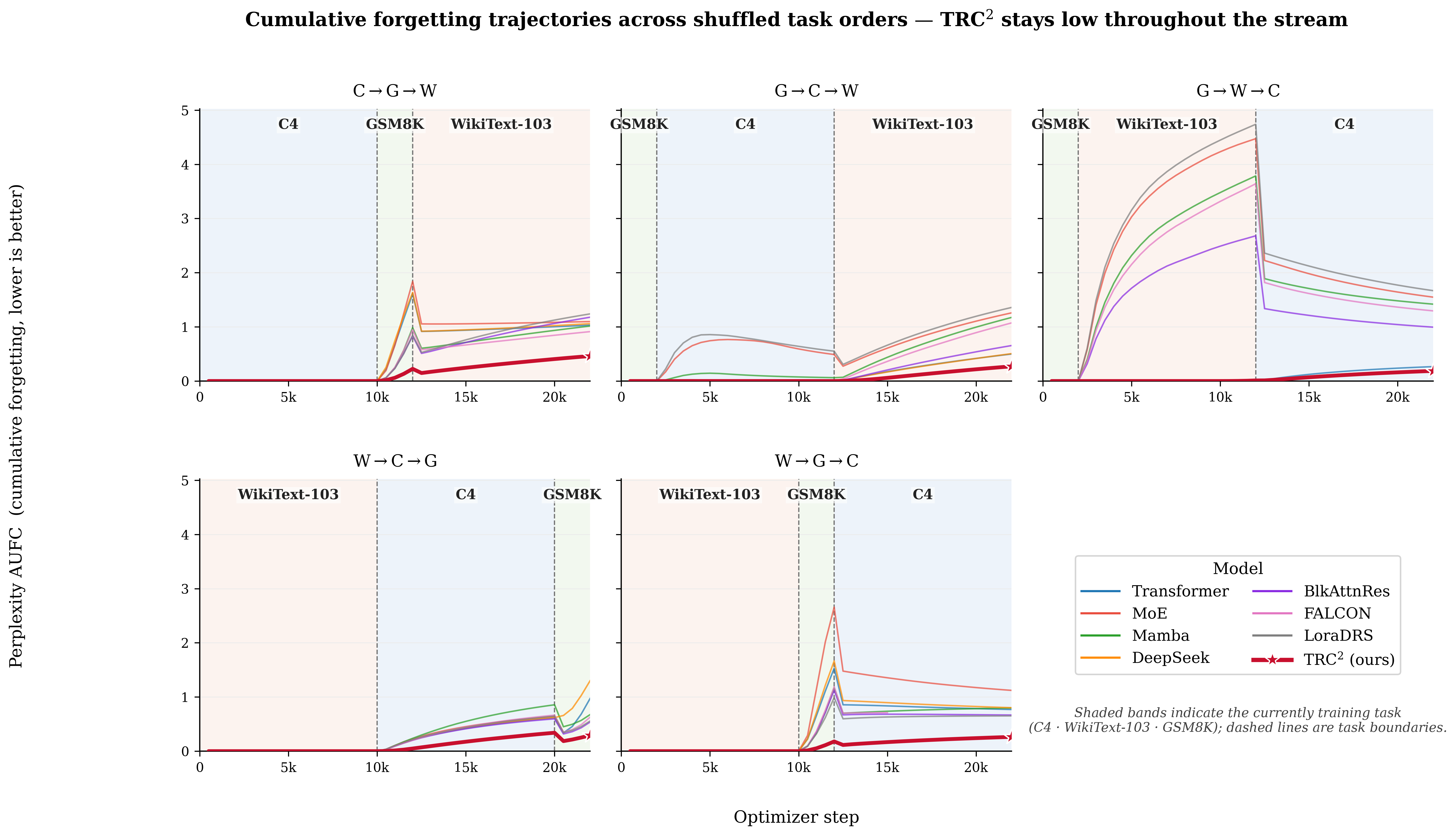}
    \caption{Cumulative forgetting trajectories across the five additional task-order shuffles. Each panel corresponds to one permutation of the original three-task stream, with shaded background bands indicating the task currently being trained and dashed vertical lines marking task boundaries. The vertical axis reports perplexity AUFC, where lower values indicate better retention. Across all five shuffled streams, \textsc{TRC}$^2$ remains in the lowest forgetting band throughout training, while most baselines accumulate substantially larger forgetting after task transitions and especially when a task is revisited late in the stream.}
    \label{fig:app_shuffle_aufc_traj}
\end{sidewaysfigure*}

\begin{sidewaysfigure*}[p]
    \centering
    \includegraphics[width=\textheight,height=\textwidth,keepaspectratio]{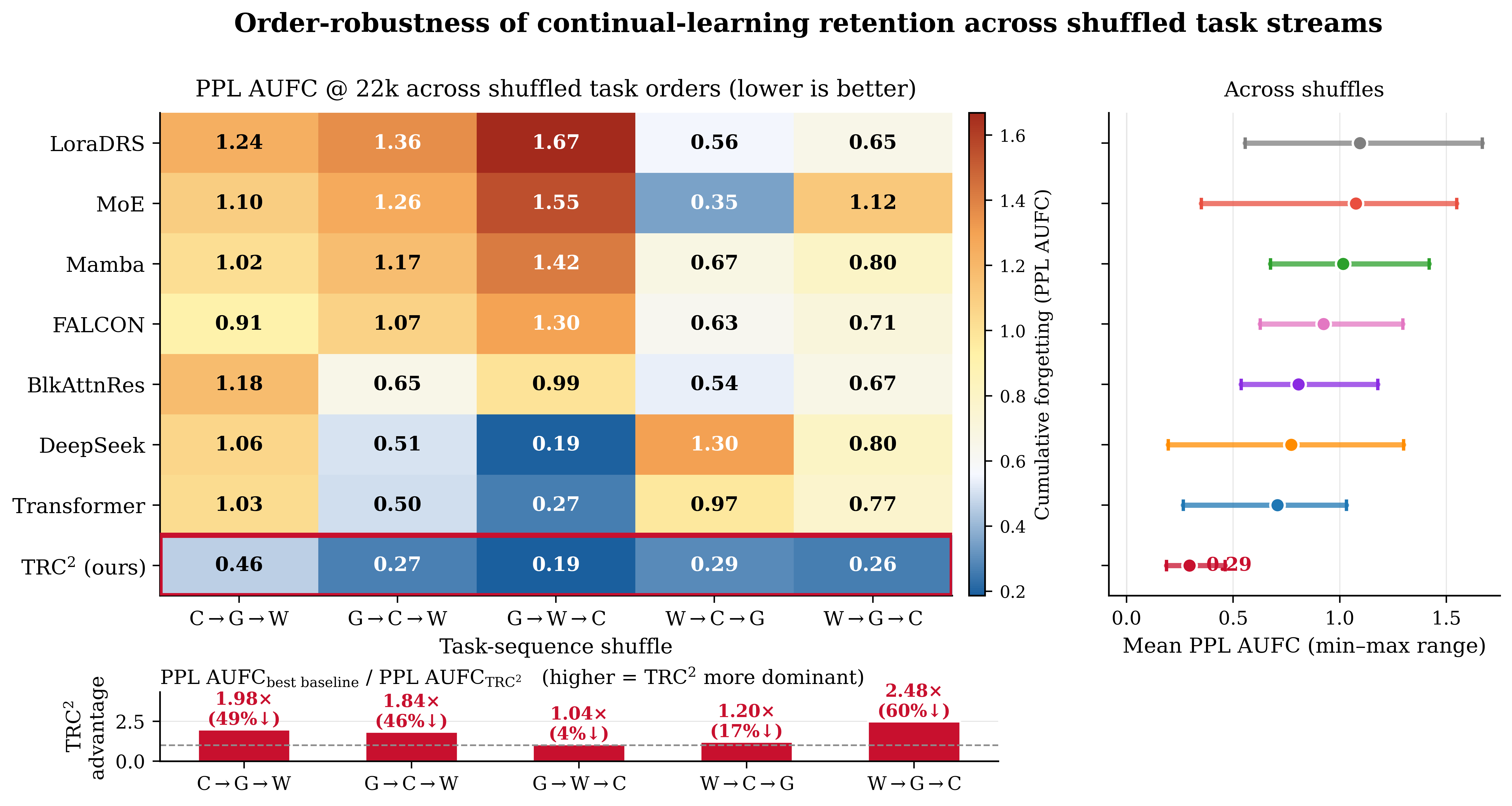}
    \caption{Order robustness of perplexity-based retention across shuffled task streams. \textbf{Left:} heatmap of PPL AUFC at step $22$k for each model family and each of the five additional shuffles, with lower values indicating better retention. \textbf{Top right:} mean PPL AUFC across shuffles together with the min--max range for each model family. \textbf{Bottom:} ratio between the best baseline PPL AUFC and the corresponding \textsc{TRC}$^2$ PPL AUFC for each shuffle, where values above $1$ indicate an advantage for \textsc{TRC}$^2$. The figure shows that \textsc{TRC}$^2$ is not only best on average but also the most order-robust family under this stress test.}
    \label{fig:app_shuffle_order_robustness}
\end{sidewaysfigure*}

\begin{sidewaysfigure*}[p]
    \centering
    \includegraphics[width=\textheight,height=\textwidth,keepaspectratio]{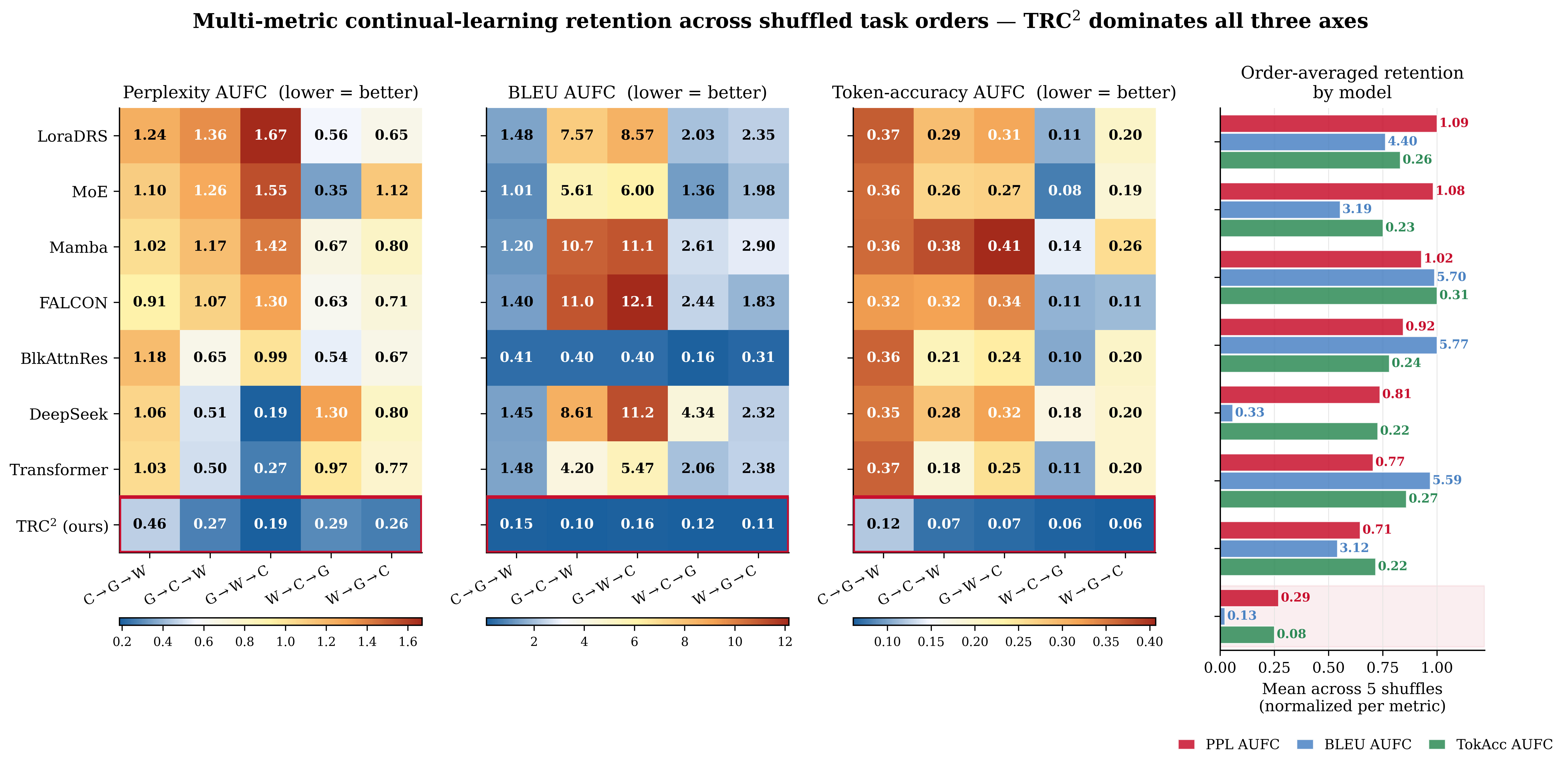}
    \caption{Multi-metric retention across shuffled task orders. The three heatmaps report PPL AUFC, BLEU AUFC, and token-accuracy AUFC across the five additional shuffles, with lower values better in every panel. The bar plot on the right summarizes the order-averaged retention profile of each model family for the three metrics. \textsc{TRC}$^2$ remains strongest across all three axes, indicating that its order robustness is not specific to one retention statistic but extends to both loss-based and text-based continual-learning summaries.}
    \label{fig:app_shuffle_multimetric}
\end{sidewaysfigure*}

\begin{sidewaysfigure*}[p]
    \centering
    \includegraphics[width=\textheight,height=\textwidth,keepaspectratio]{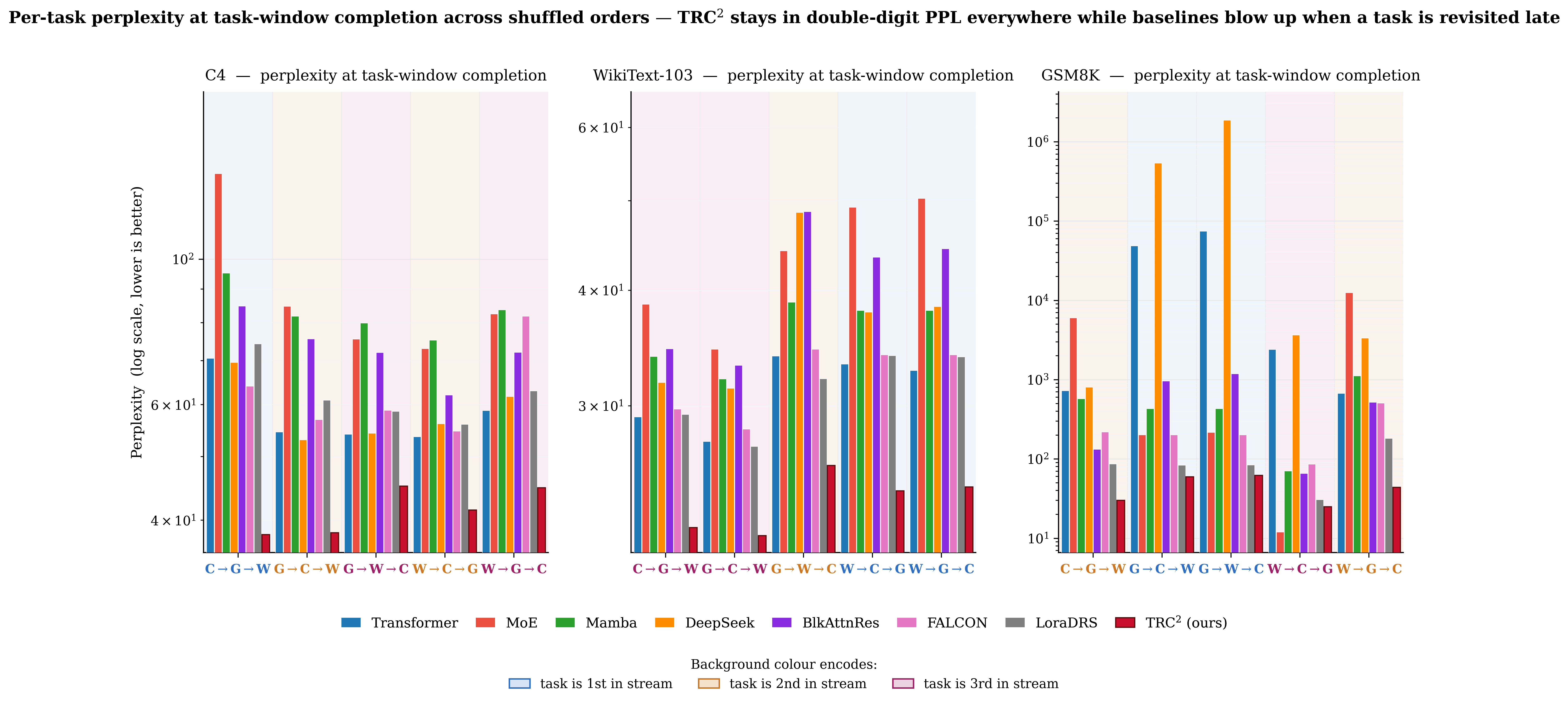}
    \caption{Per-task perplexity at task-window completion across the five additional shuffled streams. Each panel corresponds to one task, and each grouped set of bars corresponds to one shuffled order; the background color behind each group indicates whether that task appeared first, second, or third in the stream. The vertical axis is on a log scale. The main pattern is that several baselines deteriorate sharply when a task is encountered late or revisited after strong distribution shift, whereas \textsc{TRC}$^2$ remains in a comparatively low perplexity regime across all orders and all three tasks.}
    \label{fig:app_shuffle_boundary_ppl}
\end{sidewaysfigure*}

Across all five additional shuffles, the same qualitative conclusion holds: \textsc{TRC}$^2$ remains in the strongest retention region and shows substantially lower cumulative forgetting than the baseline families. The detailed evidence appears in Figures~\ref{fig:app_shuffle_aufc_traj}--\ref{fig:app_shuffle_boundary_ppl}. Figure~\ref{fig:app_shuffle_aufc_traj} shows the full AUFC trajectories over optimizer steps for each shuffled stream. Figure~\ref{fig:app_shuffle_order_robustness} summarizes the resulting PPL AUFC values at step $22$k, together with the mean and min--max range across shuffles. Figure~\ref{fig:app_shuffle_multimetric} extends the comparison to BLEU AUFC and token-accuracy AUFC, showing that the order robustness is not confined to perplexity alone. Finally, Figure~\ref{fig:app_shuffle_boundary_ppl} reports the per-task perplexity measured at the completion of each task window, which clarifies how strongly late revisits degrade the baseline models relative to \textsc{TRC}$^2$.

Two patterns are especially clear. First, the absolute amount of forgetting varies substantially with task order for the baseline families, particularly when GSM8K or WikiText-103 appears earlier and C4 is revisited late. Second, the shuffled-order variability of \textsc{TRC}$^2$ is much smaller: it stays in a low-forgetting regime throughout the stream, and its order-averaged AUFC remains below all baselines. This supports the interpretation that the retention gain in the main paper is architectural rather than an artifact of a favorable task sequence.

\section{Learning Dynamics Across the Task Stream}
\label{app:curves}

\begin{figure}[p]
    \centering

    \begin{subfigure}[t]{0.94\textwidth}
        \centering
        \includegraphics[width=\linewidth,height=0.28\textheight,keepaspectratio]{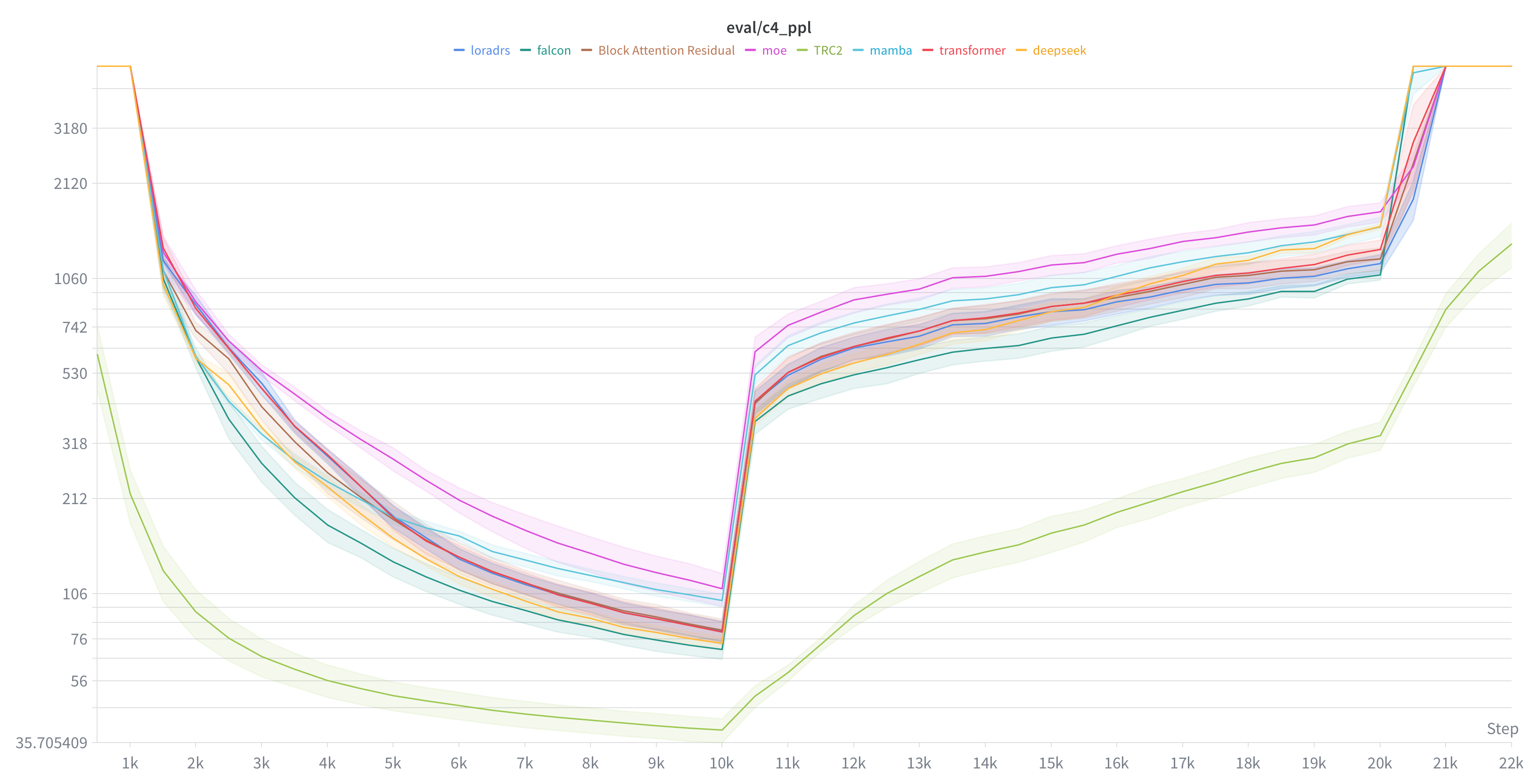}
        \caption{C4 perplexity.}
    \end{subfigure}

    \vfill

    \begin{subfigure}[t]{0.94\textwidth}
        \centering
        \includegraphics[width=\linewidth,height=0.28\textheight,keepaspectratio]{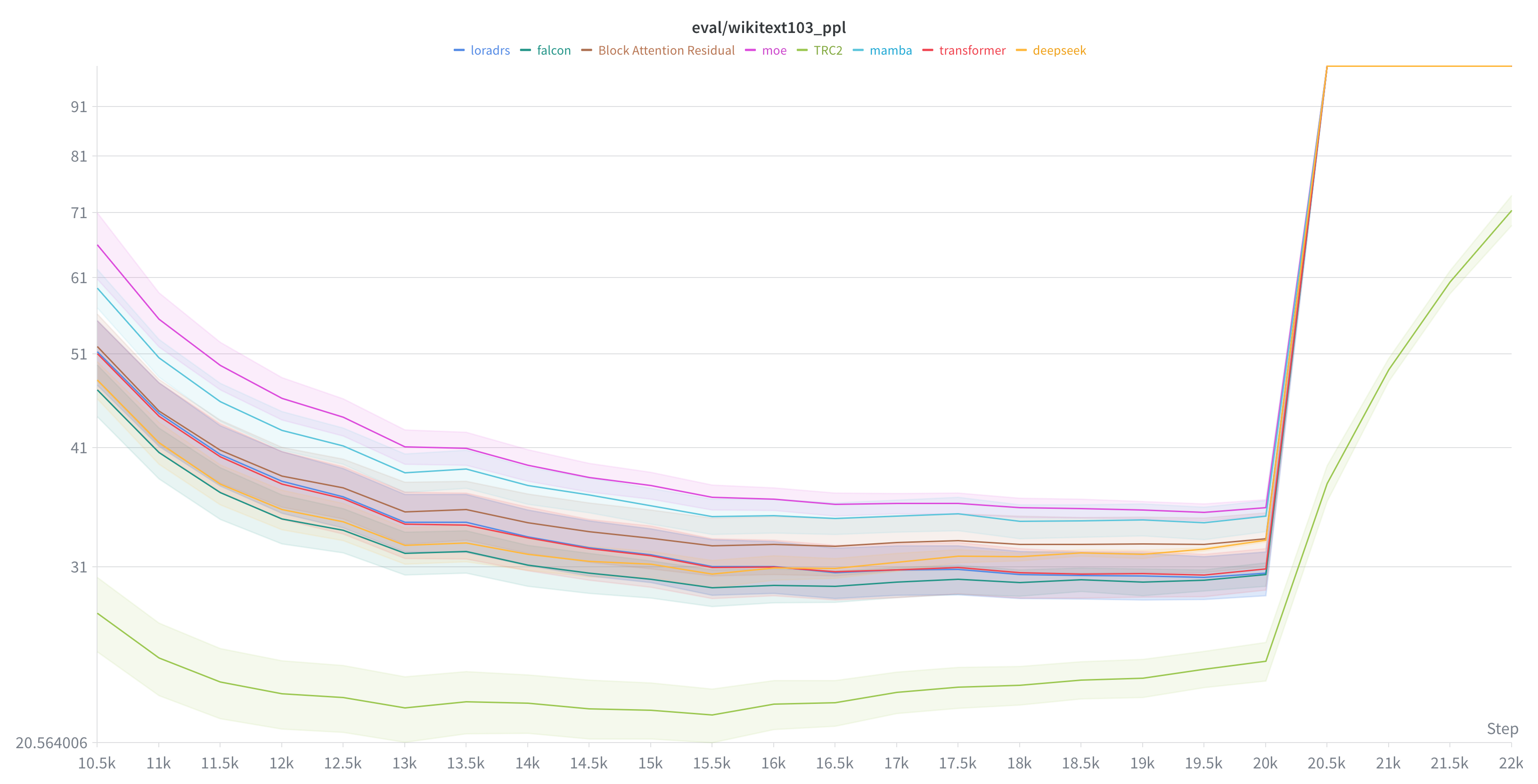}
        \caption{WikiText-103 perplexity.}
    \end{subfigure}

    \vfill

    \begin{subfigure}[t]{0.94\textwidth}
        \centering
        \includegraphics[width=\linewidth,height=0.28\textheight,keepaspectratio]{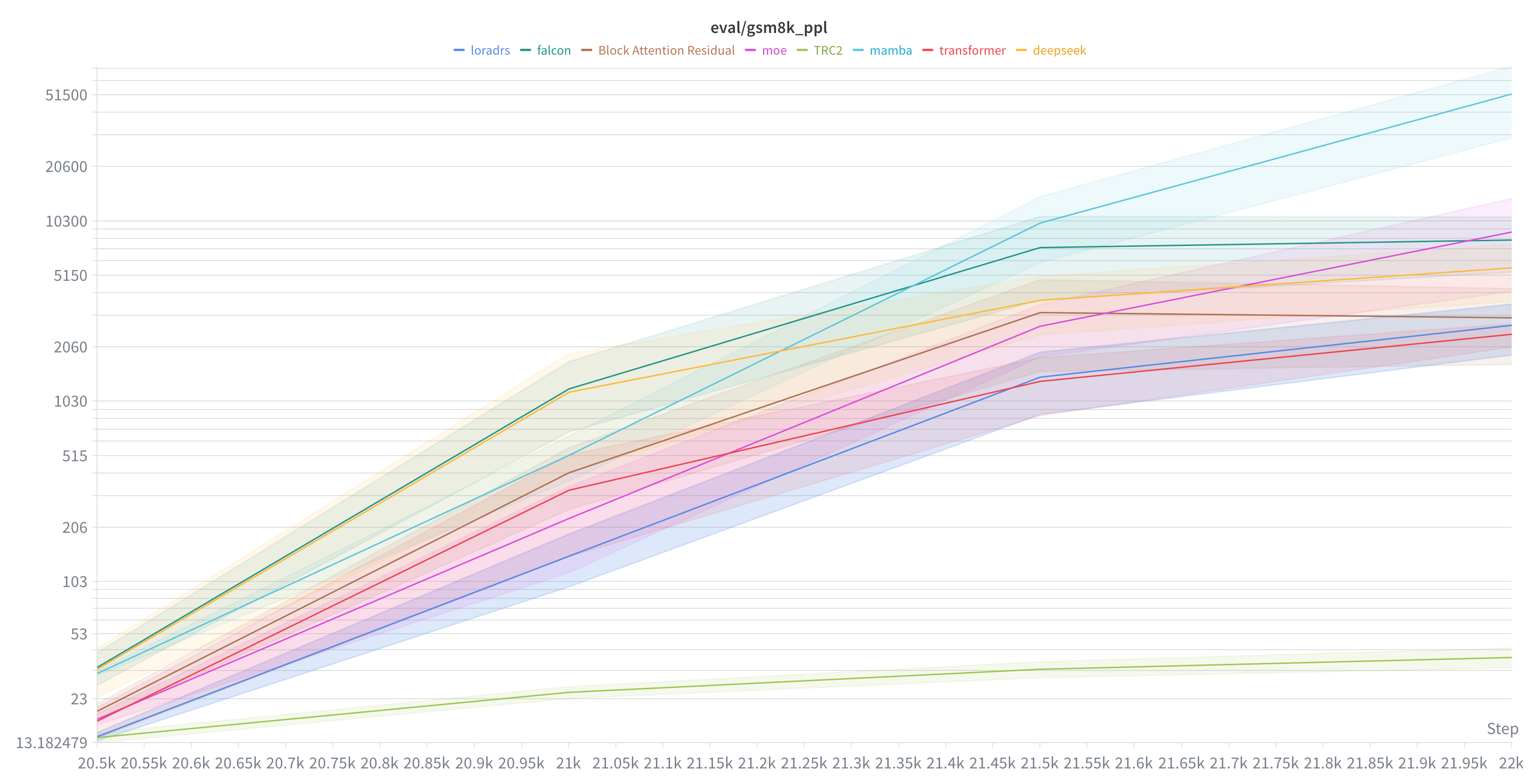}
        \caption{GSM8K perplexity.}
    \end{subfigure}

    \caption{Task-boundary perplexity trajectories across training. Solid lines show the mean across runs, and shaded bands indicate the standard error.}
    \label{fig:appendix_ppl_curves}
\end{figure}

\begin{figure}[p]
    \centering

    \begin{subfigure}[t]{0.94\textwidth}
        \centering
        \includegraphics[width=\linewidth,height=0.28\textheight,keepaspectratio]{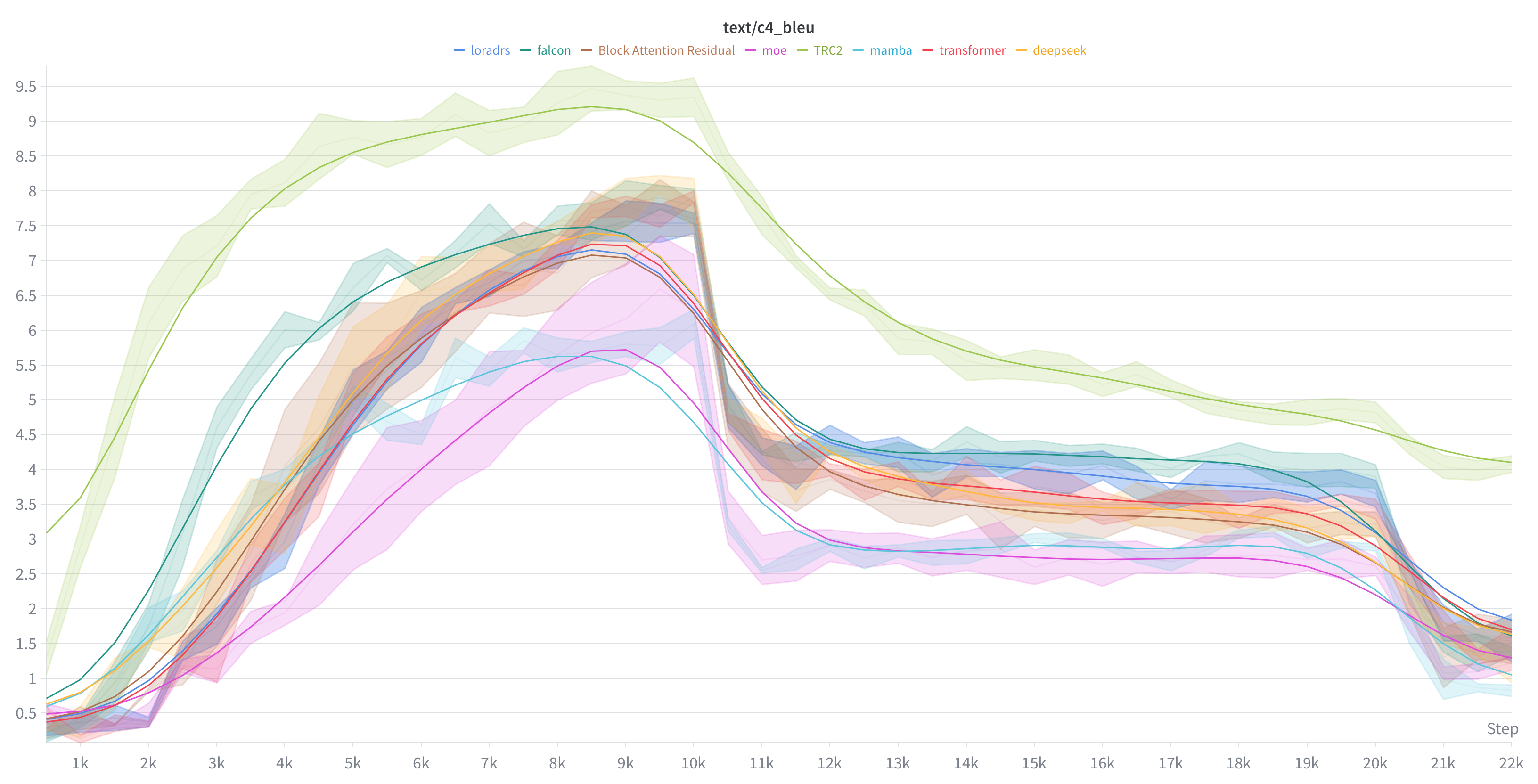}
        \caption{C4 BLEU.}
    \end{subfigure}

    \vfill

    \begin{subfigure}[t]{0.94\textwidth}
        \centering
        \includegraphics[width=\linewidth,height=0.28\textheight,keepaspectratio]{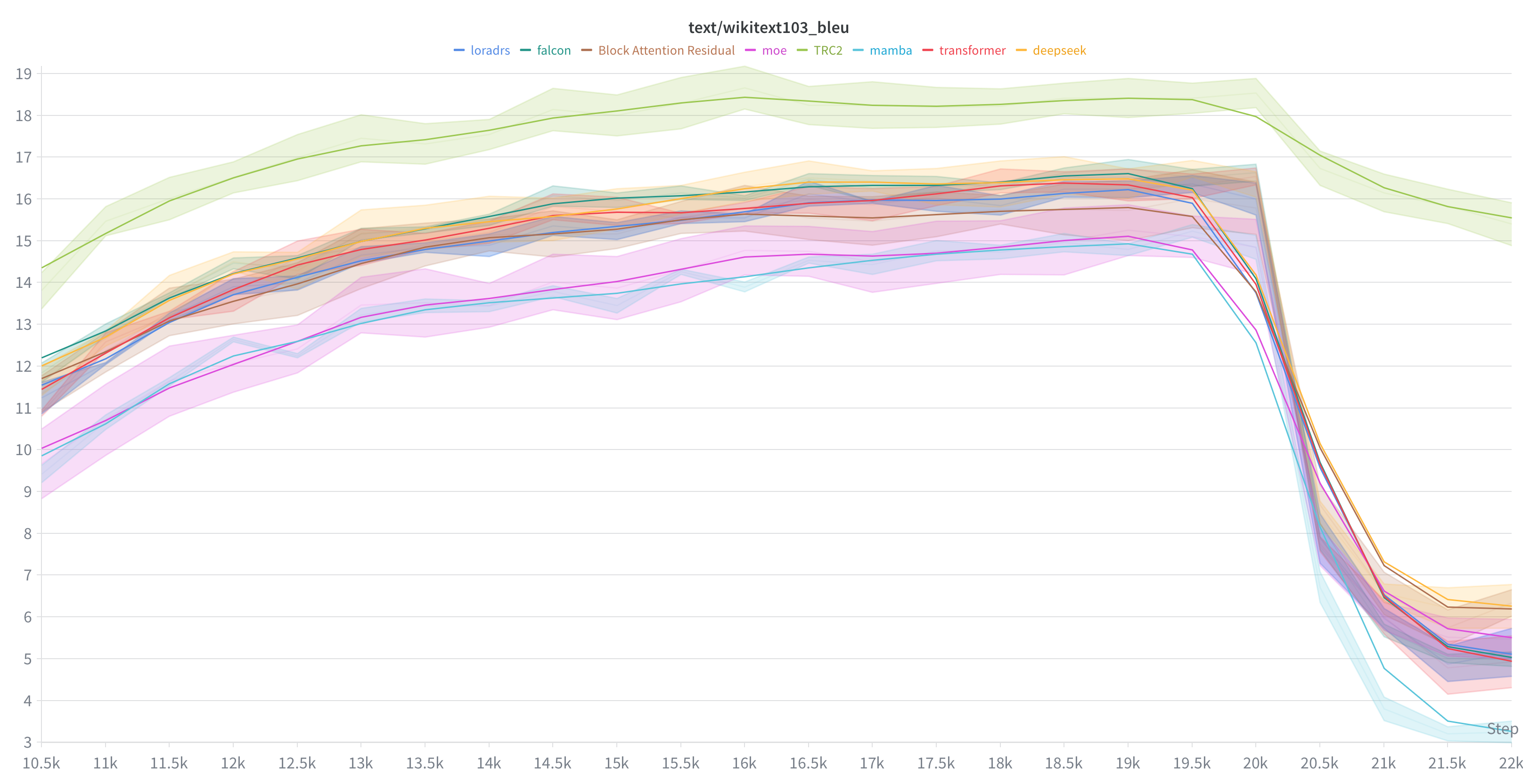}
        \caption{WikiText-103 BLEU.}
    \end{subfigure}

    \vfill

    \begin{subfigure}[t]{0.94\textwidth}
        \centering
        \includegraphics[width=\linewidth,height=0.28\textheight,keepaspectratio]{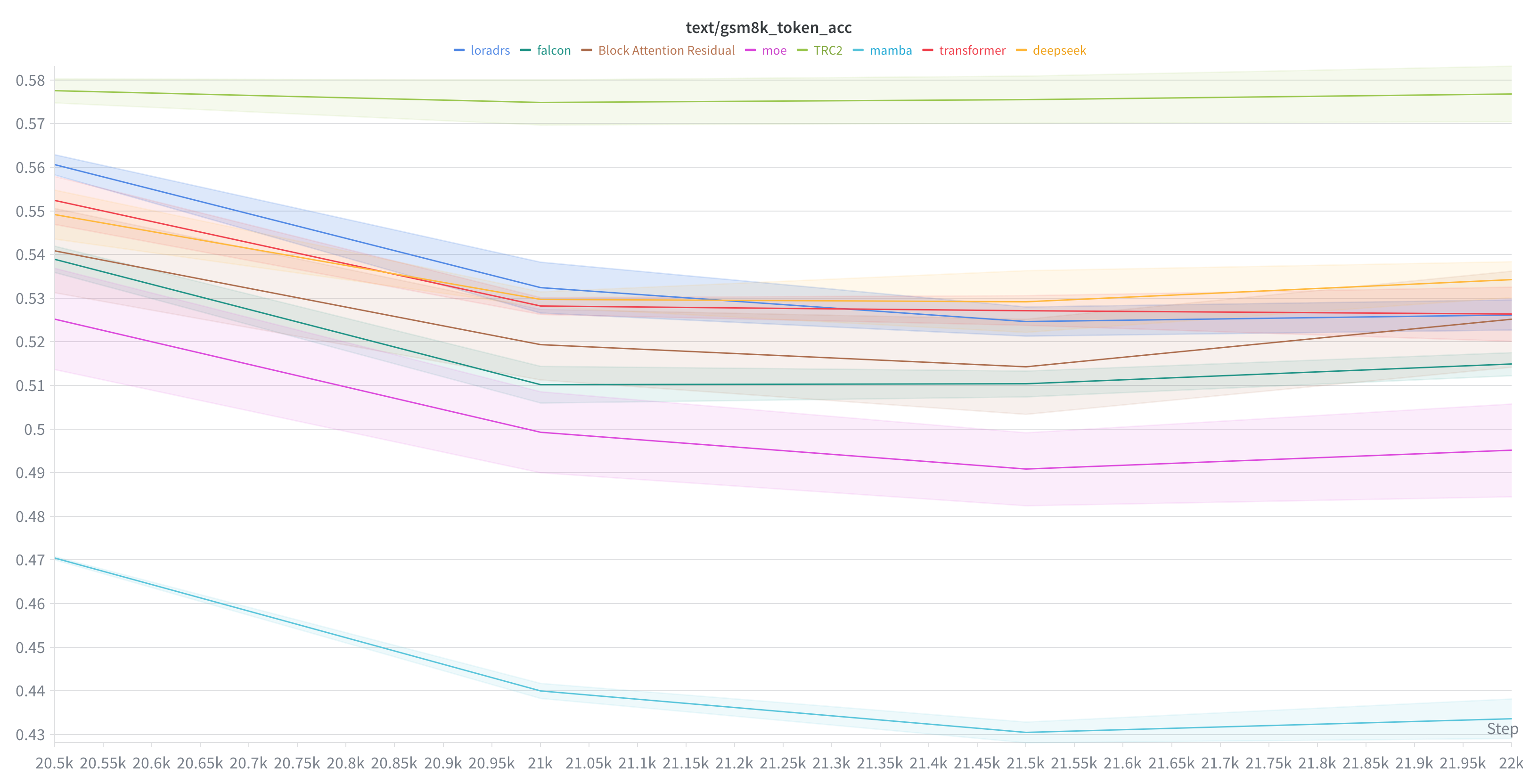}
        \caption{GSM8K token accuracy.}
    \end{subfigure}

    \caption{Task boundary text metrics across training. Solid lines for C4 and WikiText-103 show Gaussian smoothed means across runs, and shaded bands indicate the standard error.}
    \label{fig:appendix_text_curves}
\end{figure}

\begin{figure}[p]
    \centering

    \begin{subfigure}[t]{0.94\textwidth}
        \centering
        \includegraphics[width=\linewidth,height=0.28\textheight,keepaspectratio]{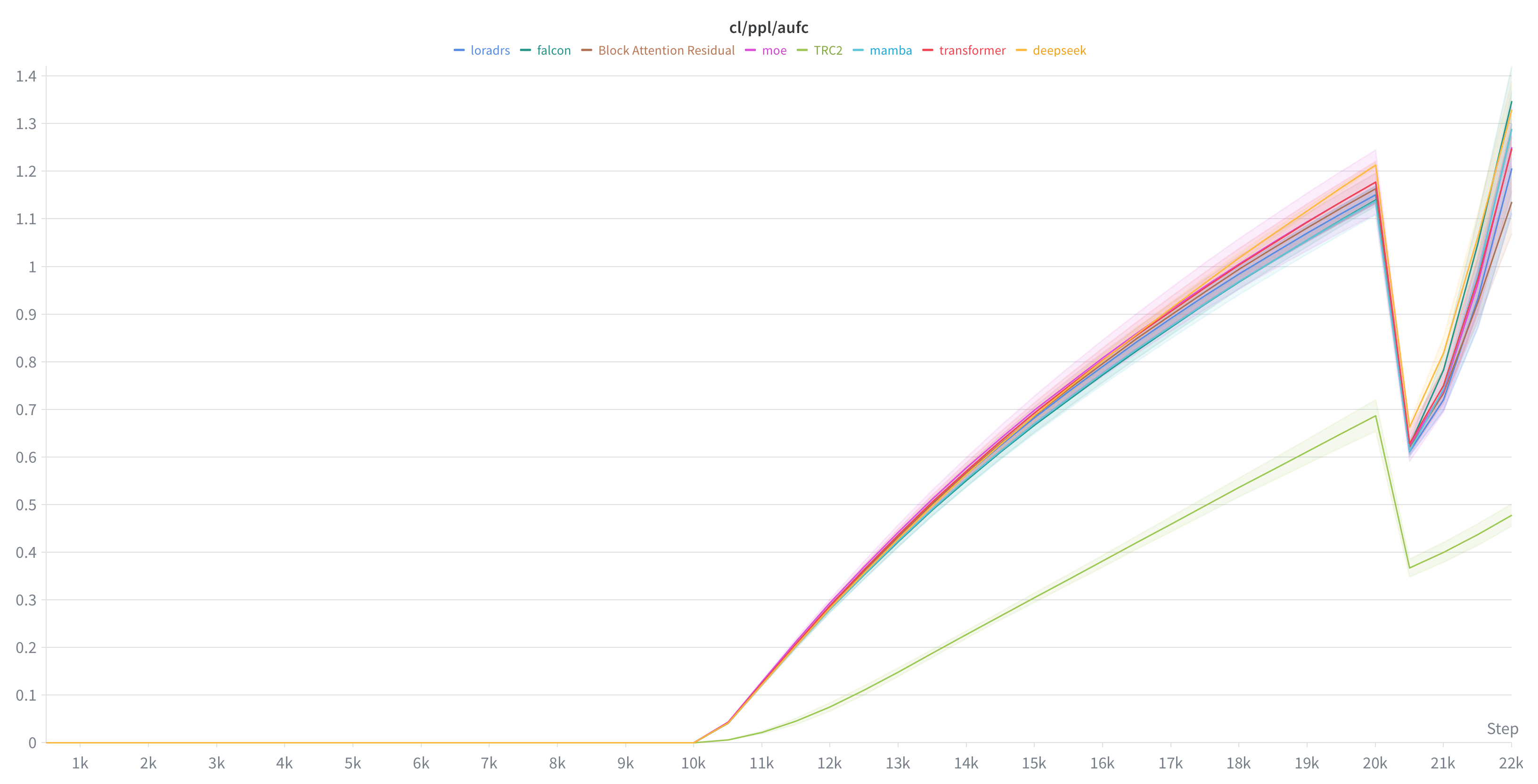}
        \caption{PPL AUFC.}
    \end{subfigure}

    \vfill

    \begin{subfigure}[t]{0.94\textwidth}
        \centering
        \includegraphics[width=\linewidth,height=0.28\textheight,keepaspectratio]{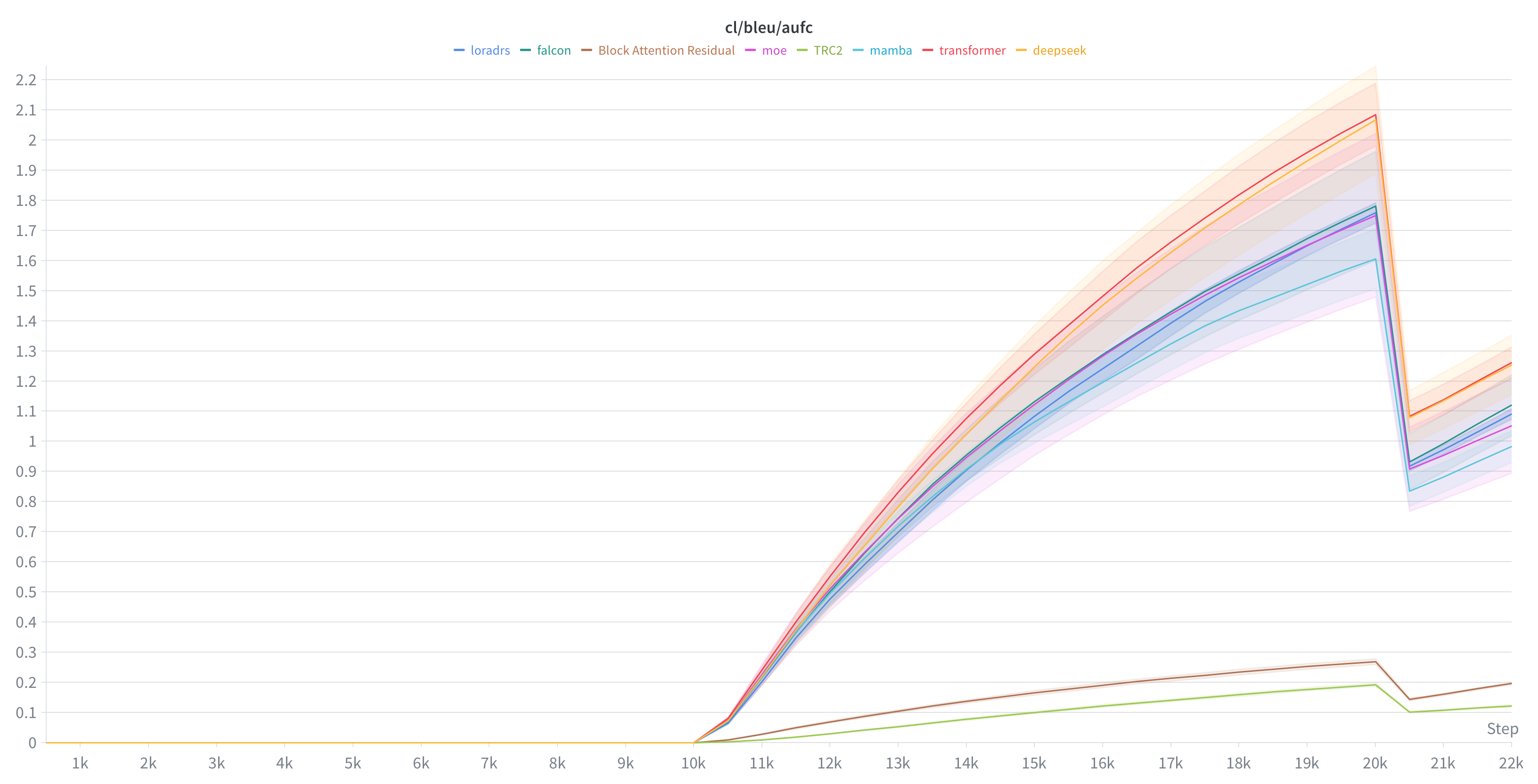}
        \caption{BLEU AUFC.}
    \end{subfigure}

    \vfill

    \begin{subfigure}[t]{0.94\textwidth}
        \centering
        \includegraphics[width=\linewidth,height=0.28\textheight,keepaspectratio]{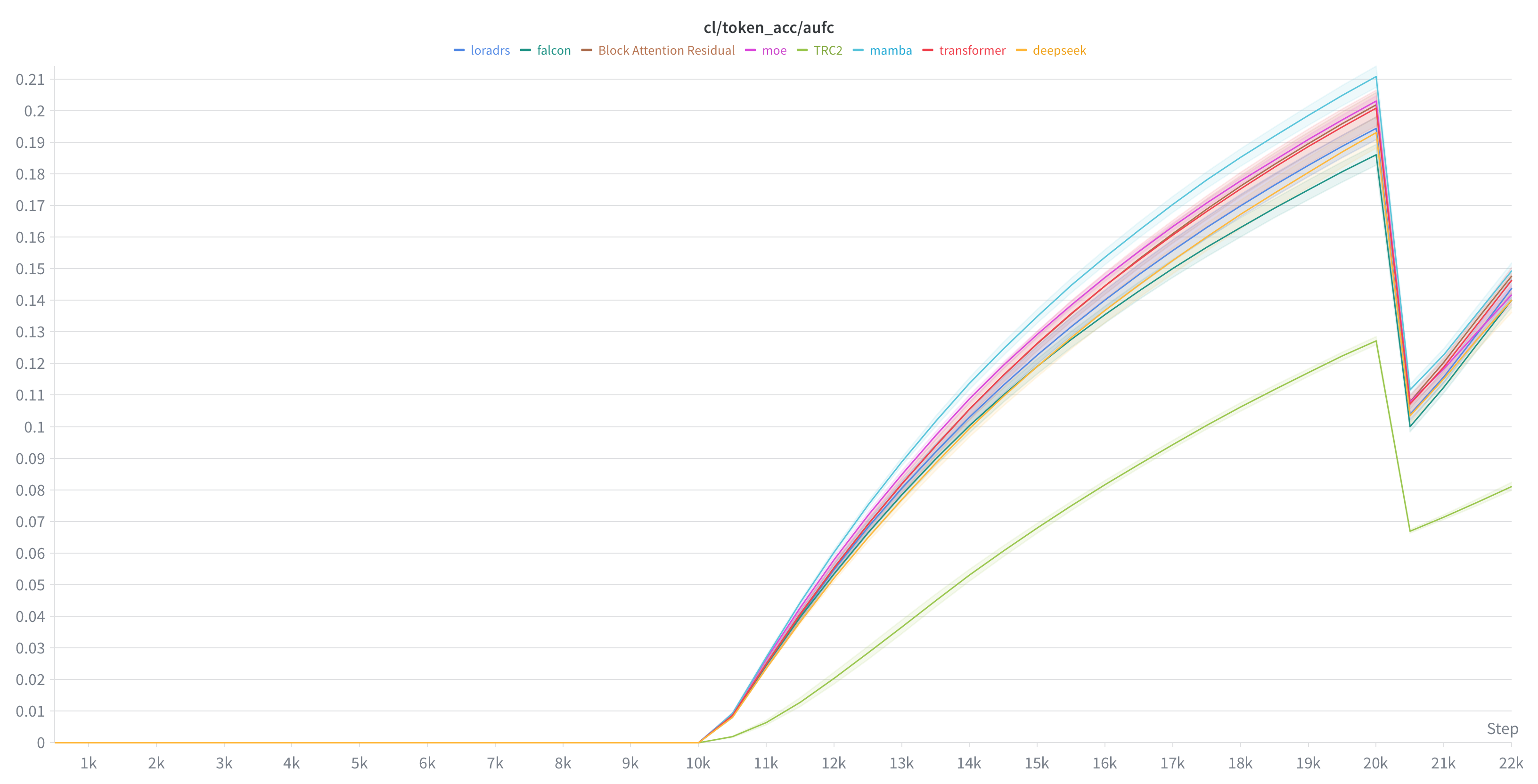}
        \caption{Token-accuracy AUFC.}
    \end{subfigure}

    \caption{Continual learning retention measured by area under the forgetting curve. Solid lines show the mean across runs, and shaded bands indicate one standard deviation. Lower values are better.}
    \label{fig:appendix_aufc_curves}
\end{figure}

Figures~\ref{fig:appendix_ppl_curves}--\ref{fig:appendix_aufc_curves} show the full evaluation trajectories. Two patterns are clear. First, the proposed \textsc{TRC}$^2$ model remains in the strongest region of the frontier on the task boundary quality curves, with the separation becoming especially visible after task switches. On C4 and WikiText-103, it reaches lower perplexity while retaining competitive or stronger teacher-forced text scores. On GSM8K, the gap is larger: several baselines deteriorate sharply after the 20k transition, whereas \textsc{TRC}$^2$ stays in a substantially lower perplexity band and maintains the strongest token-accuracy profile.

The retention plots sharpen the same conclusion. The AUFC trajectories for \textsc{TRC}$^2$ grow more slowly and remain below the baselines throughout the stream, indicating that the gains are not confined to a single checkpoint but reflect lower cumulative forgetting. This is most pronounced in perplexity AUFC, while BLEU and token accuracy AUFC show the same qualitative tendency. Taken together, these trajectory views support the main interpretation of the paper: the thalamic and hippocampal pathways improve the stability of learning over time rather than only boosting the final score on the last task.

\clearpage

\clearpage

\clearpage


\end{document}